\documentclass[twoside,10pt,dvipdfmx]{article}

\usepackage{jagi-mod}
\usepackage{times}

\usepackage{amsmath}  
\usepackage{at}
\usepackage{multirow}
\usepackage{arydshln} 
\usepackage{url}
\usepackage{tikz}
\usetikzlibrary{positioning,intersections,calc,arrows.meta}
\usepackage{graphicx}
\usepackage{color}
\usepackage[utf8]{inputenc} 

\renewcommand{\cite}[1]{\citep{#1}}

\newcommand{\notto}{\nrightarrow}
\newcommand{\To}{\Rightarrow}

\newcommand{\isInstanceOf}{{\ \circ\hspace{-4pt}\to\ }}

\newcommand{\term}[1]{\mathit{#1}}

\mathchardef\mhyp="2D 
\newcommand{\mh}{\mhyp}

\newcommand{\xxparatran}[4]{

{\color{black}#4}} 

\ShortHeadings{}{}
\firstpageno{1}

\begin{document}

\title{Non-Axiomatic Term Logic:\\
A Computational Theory of Cognitive Symbolic Reasoning}

\author{\name Kotaro Funakoshi \email funakoshi(at)lr.pi.titech.ac.jp \\
       \addr Tokyo Institute of Technology 
}

\maketitle

\begin{abstract}
This paper\footnote{This manuscript is an author-translation (to be proofread) of the original paper in Japanese, which has been submitted to ``Transactions of the Japanese Society for Artificial Intelligence.''}
presents Non-Axiomatic Term Logic (NATL) as a theoretical computational framework of humanlike symbolic reasoning in artificial intelligence.
NATL unites a discrete syntactic system inspired from Aristotle's term logic and a continuous semantic system based on the modern idea of distributed representations, or embeddings.
This paper positions the proposed approach in the phylogeny and the literature of logic, and explains the framework. 
As it is yet no more than a theory and it requires much further elaboration to implement it, no quantitative evaluation is presented. 
Instead, qualitative analyses of arguments using NATL, some applications to possible cognitive science/robotics-related research, and remaining issues towards a machinery implementation are discussed. 
\end{abstract}

\begin{keywords}
argumentation, creativity, neuro-symbolic reasoning, artificial general intelligence
\end{keywords}

\section{Introduction}\label{sec:introduction}

\xxparatran{論理学と論理は英語においてはどちらもlogicと表現される．では論理学と論理が同義であるかといえばそうではないと考えるのが自然だろう．
まず論理学の定義については，「内容によらず論証の妥当性を明らかにする方法を体系的に研究する学問」\cite{Todayama00}でよいと思われる．
一方で，論理が何をさすのかはより曖昧であるが，
Toulminは議論に関するその著作のなかで，論理の捉え方に対して，少なくとも4つの立場があるとしている\cite{Toulmin58}．
すなわち，
}{%
Both logic and logic are expressed as logic in English. Then, it is natural to think that they are synonymous with each other.
First of all, regarding the definition of logic, it seems that "a study that systematically studies the method of clarifying the validity of the argument regardless of the content" is sufficient.
On the other hand, what logic means is more ambiguous,
In his book on disputation, Toulmin states that there are at least four positions in the way of thinking about logic.
}{%
Both the study of logic and the conception of logic are expressed as simply `logic' in English. 
Regarding the definition of the study of logic, it seems that ``a study that systematically studies the method of clarifying the validity of the argument regardless of the content''~\cite{Todayama00} is sufficient.
On the other hand, what the conception of logic means is more ambiguous.
In his book on argument, \citet{Toulmin58} states that there are at least four positions in the way of thinking about logic. That is,
}{
In his book on argument, \citet{Toulmin58} states that there are at least four positions in the way of thinking about logic. That is,
}

\xxparatran{\begin{enumerate}
\item 心理的立場：論理とは個人の内の思考の法則（laws of thought）に関するものである． 
\item 社会的立場：論理とは個人と世代を超えて親や教師から伝達される推論の習慣（habits of inference）に関するものである．
\item 技術的立場：論理とは堅実な論証を行うための技巧（art of argumentation）に関するものである．
\item 数理的立場：論理とは議論の内容によらずその形式（form of argument）の正しさに関するものである．
\end{enumerate}}
{\begin{enumerate}
\item Psychological position: logic is about the law of thought within individuals.
\item Social position: Logic is about habits of inference transmitted by parents and teachers across individuals and generations.
\item Technical position: Logic is about the art of argumentation for making a solid argument.
\item Mathematical position: Logic is about the correctness of the form of argument regardless of the content of the argument.
\end{enumerate}}
{\begin{enumerate}
\item Psychological position: logic is the laws of thought within individual minds． 
\item Sociological position: Logic is the habits and practices of inference developed in the course of social evolution and passed on by parents and teachers from one generation to another.
\item Technical position: Logic is a technology, the rules of a craft, to argue soundly.
\item Mathematical position: Logic is about the validity of the form of argument rather than the content of the argument．
  \end{enumerate}
}
{
\begin{enumerate}
\item[(1)] Psychological position: Logic is the laws of thought within individual minds.
\item[(2)] Sociological position: Logic is the habits and practices of inference developed in the course of social evolution and passed on by parents and teachers from one generation to another.
\item[(3)] Technological position: Logic is a technology, the rules of a craft, to argue soundly.
\item[(4)] Mathematical position: Logic is about the validity of the form of argument rather than the content of argument.
\end{enumerate}
}

\xxparatran{冒頭の論理学の定義は，4つ目の数理的立場による．}{The definition of logic at the beginning is based on the fourth mathematical position.}
{
The definition of the study of logic at the beginning is based on the fourth mathematical position.
}{}

\xxparatran{人類史にそった論理の系統発生学的な観点から見れば，まず先史において(1) (2)の段階があり\footnote{%
直感的には(1)が(2)に先立つように思われるが，社会（文化）が人類を進化させた\cite{Henrick16}という見方が正しければ，
(1)と(2)は織り交ぜられているか，むしろ(2)が(1)に先行した可能性もある．}，
ついで有史において(3)の段階（伝統論理）にいたり，近代において(4)に昇華した（数理論理）と見ることができるだろう．
従って，(1)〜(4)は互いに独立した相反する観点というわけではなく，
関わり合いを持つ．
実際これまでに，(1)〜(3)に関わる人工知能研究の多くの課題に対して，(4)の立場で開発された命題論理，述語論理が利用されてきたのは周知の通りである（例えば\cite{Litman87}）．しかしウィトゲンシュタインによれば，数理論理においては『現実の事態を表現することについては全く考慮されていない』\cite{Janik73}（『』内の表現は邦訳による）．}
{From a phylogenetic point of view of logic along human history, there are stages (1) and (2) in prehistory. \footnote{%
Intuitively, (1) seems to precede (2). However, if the view that society (culture) has evolved humanity is correct, then (1) and (2) may be interwoven, or rather (2) may precede (1). There is also sex.} Then, it can be seen that it was in the stage (3) (traditional logic) in recorded history, or sublimated to (4) in modern times (mathematical logic). Therefore, (1) to (4) are not independent and contradictory viewpoints, but have a relationship. In fact, it is well known that propositional logic and predicate logic developed from the standpoint of (4) have been used for many issues of artificial intelligence research related to (1) to (3). There is (for example, \cite{Litman87}). However, according to Wittgenstein, in mathematical logic,``the expression of the actual situation is not considered at al'' \cite{Janik73} (the expression in "" is a Japanese translation).}
{From the perspective of Phylogeny along human history, in prehistory, there are stages (1) and (2)\footnote{%
Intuitively, (1) seems to precede (2), but if the view that society (culture) evolved human beings\cite{Henrick16}is correct，
(1) and (2) might be interwoven, and it is better to say that (2) may precede (1).},followed by the phase (3) (traditional logic), and then the phase (4) in modern times (mathematical logic). Thus, (1) to (4) are not mutually independent, opposite view, but they are related to each other. In fact, until now, it is well known that propositional logic and predicate logic developed from the point of (4) have been used for many subjects of artificial intelligence related to (1) to (3) (For example, \cite{Litman87}). 
However, according to Wittgenstein, in mathematical logic,``the expression of the actual situation is not considered at al'' \cite{Janik73} (the expression in "" is a Japanese translation).
\cite{Janik73}
}
{
From a phylogenetic point of view of logic along human history, we would say that there were the stages of (1) and (2) in prehistory,\footnote{%
Intuitively, the stage of (1) seems to precede the stage of (2), but if the view that society (culture) evolved human beings \cite{Henrick16} is correct, (1) and (2) might be interwoven, and it is better to say that (2) may precede (1).}
followed by the stage of (3) (traditional logic), and then the stage of (4) in modern times (mathematical logic). Therefore, the four positions of (1) to (4) are not independent and contradictory viewpoints, but have a relationship. 
In fact, until now, it is well known that propositional logic and predicate logic developed from the point of (4) have been used for many subjects of artificial intelligence related to (1) to (3) (for example, \cite{Litman87}). However, according to Wittgenstein, 
\textit{``In constructing symbolic logic, Frege, Peano and Russell always had their eye on its application to mathematics alone, and they never gave any thought on the representation 
 of real states of affairs
.''} 
\cite{Janik73}}

\xxparatran{本論文は，人工知能のための，ある種の記号論理とそれに基づく推論法の枠組みを提案する．
本研究は，(4)の立場を一旦放棄し，(1) (2)へと回帰する．つまり本論文は，述語論理でも命題論理でもないものを提案する\footnote{
(4)の立場を放棄しているので，数理論理が数学に対してなした貢献をそのまま肩代わりできるものではない．
}．従って，提案する枠組みは，論理学の範疇には属さない．
ただし，それは人工知能のための枠組みであるので，計算機実装が可能な程度に形式化されている必要がある．従って，形式論理的である．
本枠組みよって実現された人工知能はいずれ，(3) (4)に属する論理を，小学生が算数を習うように，文化として獲得する（社会的に学習する）目論見である．
その段階に行きつくまでの課程で，本研究が主に対象とするのは，一杉ら\cite{Ichisugi20}が例にあげるような，
「常に正しいとは限らないが特定の状況においては一定の有用性を持つ規則」\footnote{\label{fn:choco}
 「きのう戸棚にチョコレートがあった。ということは、きょうもあるはず。」
}をもとにして，人間が日常的に行っている推論である．
これをアリストテレスは，説得推論（蓋然的推論）とよんだ\cite{Nouchi02}．}
{This paper proposes a framework of some kind of symbolic logic and inference methods based on it for artificial intelligence. This study temporarily abandons the position of (4) and returns to (1) and (2). In other words, since this paper abandons the position of \footnote{(4), which proposes something that is neither predicate logic nor propositional logic, it cannot directly replace the contribution that mathematical logic has made to mathematics.}. Therefore, the proposed framework does not belong to the category of logic. However, since it is a framework for artificial intelligence, it needs to be formalized to the extent that it can be implemented on a computer. Therefore, it is formal logic. Artificial intelligence realized by this framework will eventually acquire (socially learn) the logic belonging to (3) and (4) as a culture, just as elementary school students learn mathematics. In the process leading up to that stage, the main target of this research is "although not always correct, but constant in certain situations," as Ichisugi et al. \cite {Ichisugi20} gives as an example. A rule with usefulness "\footnote{\label{fn: choco}" There was chocolate in the cupboard yesterday. That should be today. "} It is inference. Aristotle called this convincing reasoning (probable reasoning) \cite{Nouchi02}.
}{In this paper, we propose a kind of symbolic logic and a framework of reasoning method based on it. In this research, once the position of (4) is abandoned, we will return to (1) and (2). In short, this paper is neither predicate nor propositional logic. \footnote{Since it abandons the position of (4), it cannot directly take the place of the contribution that mathematical logic has made to mathematics.} Therefore, the proposed framework does not belong to the logical category. However, since it is a framework for artificial intelligence, it is necessary to be formalized to the extent that computer can implement it. Thus, it is formal logic. The artificial intelligence realized by this framework is a plan to eventually acquire (socially learn) the logic belonging to (3) and (4) as a culture, like elementary school students learning arithmetic. In the process leading up to that stage, the main object of this research is the ``rules that are not always correct but have a certain degree of usefulness in specific situations,'' as exemplified by Ichisugi et al. ”\cite{Ichisugi20}\footnote{\label{fn:choco} According to ``There was chocolate in the cupboard yesterday, and there should be chocolate today as well.''}. Aristotle called this convincing reasoning (probable reasoning)\cite {Nouchi02}.
}
{
This paper proposes a kind of symbolic logic and a framework of reasoning based on it for artificial intelligence. This study temporarily abandons the position of (4) and returns to (1) and (2). In short, this paper is neither about predicate nor propositional logic. Therefore, the proposed framework does not belong to the category of mathematical logic.\footnote{Since it abandons the position of (4), it cannot directly take the place of the contribution that mathematical logic has made to mathematics.} 
However, since it is a framework for artificial intelligence, it is necessary to be formalized to the extent that it can be implemented on a computer. Thus, it is formal logic. 
Artificial intelligence realized by this framework will eventually acquire (socially learn) the logic belonging to (3) and (4) as a culture, just as elementary school students learn mathematics. In the process leading up to that stage, the main target of this research is
``rules that are not always correct but have a certain degree of usefulness in specific situations,'' as exemplified by \cite{Ichisugi20}.\footnote{\label{fn:choco} ``There was chocolate in the cupboard yesterday, then, there should be chocolate today as well.''}
}

\xxparatran{本論文が提案する非公理的項論理（Non-Axiomatic Term Logic; NATL）の枠組みは，非公理的論理\cite{Wang94,Wang13}（Non-Axiomatic Logic; NAL）に着想を得ている．項論理または名辞論理（term logic）とは，これも先に触れたアリストテレスによって整理された三段論法（syllogism）に遡る論理の一種で，述語と個体変数の区別を持たず，2つの項をつなぐ形で文（命題）を表現するもので，
伝統論理の範疇に含まれるが，近代にフレーゲにより発明された述語論理によって事実上駆逐された状態にある．}
{The non-axiomatic term logic (NAL) framework proposed in this paper is inspired by the non-axiomatic logic \cite{Wang94, Wang13} (Non-Axiomatic Logic; NAL).
Term logic or term logic
(Term logic) is a kind of logic that goes back to the syllogism organized by Aristotle, which I mentioned earlier. It does not distinguish between predicates and individual variables, and it is a sentence (proposition) that connects two terms. ) Is expressed.
Although it is included in the category of traditional logic, it is virtually eliminated by the predicate logic invented by Frege in modern times.}
{The framework of Non-Axiomatic Term Logic (NATL) proposed by this paper is inspired by Non-Axiomatic Logic (NAL) \cite{Wang94, Wang13}. Term logic, which can go back a kind of logic of syllogism organized by Aristotle mentioned before, does not have discrimination between predicates, individual variables, and express as sentences combined by two connected items. Although it is included in the category of traditional logic, it is virtually eliminated by predicate logic invented by phase in modern times.
}
{
The framework of Non-Axiomatic Term Logic (NATL) proposed in this paper is inspired by Non-Axiomatic Logic (NAL) \cite {Wang94, Wang13}. Term logic goes back to the syllogism of Aristotle. It does not have discrimination between predicates and individuals, and expresses a sentence by connecting two terms by a copula. 
It is included in the category of traditional logic, which is virtually eliminated by predicate logic invented by Frege in modern times.}

\xxparatran{一方で，少数ながら，項論理の持つ簡素さや直感的な理解のしやすさなどに惹かれた内外の研究者らによって，述語論理に代わる独自の論理・推論法が，現代においても時折提案されてきた\cite{Sommers82,Morita87,Nishihara94,Wang94,Goertzel08,Moss10}．
本研究の出発点は，近年の深層学習による情報処理技術との融合により，前述のような人間の日常的推論に対し，項論理が有用性を示すとする仮定である．
ただし，本研究は項論理に関する上記の諸研究とは一線を画する．
従って本論文が提案するものは，これらの研究の延長線上にあるものでもない}
{On the other hand, a small number of researchers inside and outside the country, who were attracted to the simplicity of term logic and the ease of intuitive understanding, occasionally proposed original logic and reasoning methods to replace predicate logic. I came \cite{Sommers82, Morita87, Nishihara94, Wang94, Goertzel08, Moss10}.
The starting point of this research is the assumption that term logic is useful for the above-mentioned daily reasoning of human beings by fusing with information processing technology by deep learning in recent years.
However, this study is different from the above studies on term logic.
Therefore, what this paper proposes is not an extension of these studies.}
{On the other hand, a small number of domestic and overseas researchers who are attracted by the simplicity and ease of intuitive understanding propose original logic and reasoning methods to replace predicate logic in present age\cite{Sommers82,Morita87,Nishihara94,Wang94,Goertzel08,Moss10}. The starting point of this research is the assumption that term logic is useful for the mentioned daily reasoning of human beings by fusing with information processing technology based on recent deep learning technology. However, this research is different from the above-mentioned researches. Therefore, the proposal in this paper is not an extension of such studies.
}
{
Nevertheless, a small number of researchers who are attracted by the simplicity and ease of intuitive understanding of term logic occasionally proposed original logic frameworks and reasoning methods to replace predicate logic \cite{Sommers82,Morita87,Nishihara94,Wang94,Goertzel08,Moss10}. The starting point of this research is the assumption that term logic is useful for modeling the aforementioned daily reasoning of human beings by fusing with recent information processing technologies based on deep learning. However, this research is different from the above-mentioned previous studies, and the proposal in this paper is not an extension of those studies.}

\xxparatran{本研究では，人間的な思考と親和性が高いと思われる離散系としての項論理を骨格とし，データから暗黙的・経験的に構成される多次元空間内における連続系としての意味表象に基づく情報処理技術を血肉とすることで，人間が行うような日常的推論，創造的な記号処理，記号を介した文化学習を行うシステムの実現を目指す．}
{This research is based on term logic as a discrete system, which seems to have a high affinity with human thinking, as a framework, and is based on semantic representation as a continuous system in a multidimensional space implicitly and empirically constructed from data. By using information processing technology as flesh and blood, we aim to realize a system that performs human-like daily reasoning, creative symbol processing, and cultural learning through symbols.}
{This research is based on term logic as a discrete system, which seems to have a high affinity with human thinking, as a framework, and is based on semantic representation as a continuous system in a multidimensional space implicitly and empirically constructed from data. By using information processing technology as flesh and blood, we aim to realize a system that performs human-like daily reasoning, creative symbol processing, and cultural learning through symbols.
}
{
Being based on term logic as a discrete system, which seems to have a high affinity with human thinking, as the skeleton, and being based on semantic representation as a continuous system in a multidimensional space implicitly and empirically constructed from data, as the flesh and blood, we aim to realize an AI system that performs human-like daily reasoning, creative symbol processing, and cultural learning through symbols.}

\xxparatran{このようなシステムは，特定のタスクに対する性能だけを考えれば，深層学習のみに基づきそのタスクに特化して大量のデータで訓練されたシステムにおそらく劣るであろう．特に人間の存在とは独立して定義できるタスクにおいては，その差はより顕著になると予想される．しかしながら，物語の解釈や会話など人間の思考との類似性・親和性が求められる場面，推論の過程・根拠を人に示さなくてはならない場面，人間的な錯誤を予知しなければならない場面，記号的に与えられる極少数のアドホックな知識・規則をその場で運用することが求められる場面などで，独自の有用性を確立できる可能性がある．}
{Such a system, if we only consider its performance on a specific task, will probably be inferior to a task-specific trained system based on deep learning alone and with large amounts of data. Especially in tasks that can be defined independently of the presence of humans, the difference is expected to become more pronounced. However, there are situations where similarity and affinity with human thinking are required, such as interpretation of stories and conversations, situations where the process and basis of reasoning must be shown to people, situations where human error must be predicted, There is a possibility that unique usefulness can be established in situations where it is required to operate a very small number of ad-hoc knowledge and rules given symbolically on the spot.}
{If only the performance of specific tasks is considered, this system might be not good as the system that only uses deep learning technology and trained by a large amount of data for such tasks. Especially for the tasks that can be defined independently without human existence, the performance gap will be more obvious according to expectation. However, it may establish its own usefulness in situations that require similarity and affinity with human thought, such as interpretation of a story or conversation; in situations where the process and basis of inference must be presented to others; in situations where human errors must be predicted; and in situations where a very small number of ad hoc knowledge and rules given symbolically must be operated on the spot.
}
{
Such an AI system, if we only consider its performance on a specific task, will probably be inferior to a task-specificly trained system based on deep learning alone and with large amounts of data. Especially in tasks that can be defined independently of the presence of humans, the difference is expected to become more pronounced. However, it may establish its own usefulness in situations that require similarity and affinity with human thought, such as interpretation of a story or conversation; in situations where the process and basis of inference must be presented to others; in situations where human errors must be predicted; and in situations where a very small number of ad-hoc knowledge and rules given symbolically must be operated on the spot.}

\xxparatran
{本論文の構成は以下の通りである．
まず\S\ref{sec:logics}で項論理と述語論理の関係を整理し，項論理を発展させようとしたこれまでの研究のそれぞれの立場・目論見を整理し，
\S\ref{sec:nal}でWangによる非公理的論理の枠組みの概要とその課題を議論する\footnote{
\S\ref{sec:logics}及び\S\ref{sec:nal}の内容は，\cite{Funakoshi21}を基にしている．
}．
次に\S\ref{sec:trl}と\S\ref{sec:natl}で，\S\ref{sec:nal}で示した課題を踏まえて考案した
非公理的項論理とそれが用いる項表示言語を提示する．
非公理的項論理の有用性を部分的にでも示すため，\S\ref{sec:argumentation}にて，非公理的項論理を用いた議論の定性的な事例分析を示す．
構想の有望性を示すために，\S\ref{sec:applications}で，認知科学やロボティクスとのつながりにおいて，研究の展開を議論する．
最後に，\S\ref{sec:implementation_and_issues}で計算機実装の見通しと今後の課題について議論し，\S\ref{sec:conclusion}で本論文の内容をまとめるとともに，自然言語とは別の形式的表示手段と本論文の提案の意義を確認する．}
{The structure of this paper is as follows.
First, \S\ref{sec:logics} is used to sort out the relationship between term logic and predicate logic, and then sort out the positions and plans of each of the studies so far that have tried to develop term logic.\S\ref{sec:nal} discusses Wang's outline of the framework of axiomatic logic and its issues \footnote{The contents of \S\ref{sec:logics} and \S\ref{sec:nal} are based on \cite{Funakoshi21}.}.Next, \S\ref{sec:trl} and \S\ref{sec:natl} were devised based on the issues shown in \S\ref{sec:nal}.We present irrational term logic and the term display language it uses.
In order to show the usefulness of irrational term logic even partially, \S\ref{sec:argumentation} shows a qualitative case analysis of the argument using irrational term logic.To show the promise of the concept, \S\ref{sec:applications} discusses the development of research in connection with cognitive science and robotics.Finally, in \S\ref{sec:implementation_and_issues}, we discuss the prospects and future issues of computer implementation, and in \S\ref{sec:conclusion}, we summarize the contents of this paper, and use a formal presentation method other than natural language. and confirm the significance of the proposal in this paper.}{The structure of this paper is as follows. First, \S\ref{sec:logics} is used to sort out the relationship between term logic and predicate logic, and sort out then the positions and plans of each of the studies so far that have tried to develop term logic. \S\ref{sec:nal} discusses the outline of the framework of axiomatic logic and its issues based on Wang's \footnote{The content of \S\ref{sec:logics} and \S\ref{sec:nal} are based on \cite{Funakoshi21}．}．Next, in \S\ref{sec:trl} and \S\ref{sec:natl}, we present a non-axiomatic term logic and the term representation language it uses, based on the issues presented in \S\ref{sec:nal}. To show the usefulness of non-axiomatic term logic partially, the qualitative case analysis of discussion using non-axiomatic term logic from \S\ref{sec:argumentation} is shown. To demonstrate the promise of the concept, the development of research about connection between cognitive science and robotics will be discussed  in \S\ref{sec:applications}. Finally, in \S\ref{sec:implementation_and_issues}, we discuss the prospects and future issues of computer implementation, and in \S\ref{sec:conclusion}, we summarize the contents of this paper, and use a formal presentation method other than natural language, confirming the significance of the proposal in this paper.
}
{
The structure of this paper is as follows. First, \S\ref{sec:logics} is used to sort out the relationship between term logic and predicate logic, and sort out then the positions and plans of each of the studies so far that have tried to develop term logic. \S\ref{sec:nal} discusses the outline of Wang's Non-Axiomatic Logic and its issues. Next, in \S\ref{sec:trl} and \S\ref{sec:natl}, we present our Non-Axiomatic Term Logic and its knowledge representation language, based on the issues presented in \S\ref{sec:nal}. To show the usefulness of Non-Axiomatic Term Logic partially, the qualitative case analysis of arguments using Non-Axiomatic Term Logic is shown in\S\ref{sec:argumentation}. To demonstrate the promise of the concept, the possible development of research in connection with cognitive science and robotics will be discussed in \S\ref{sec:applications}. Finally, in \S\ref{sec:implementation_and_issues}, we discuss the prospects and future issues of computational implementation, and in \S\ref{sec:conclusion}, we summarize this paper.}

\xxparatran
{本論文の主要な貢献は，以下の2点である．}
{The main contributions of this paper are the following two points.}
{The main contributions of this paper are the following two points.}
{The main contributions of this paper are two-fold:}
\xxparatran
{\begin{enumerate}
    \item 離散系の構文論と連続系の意味論の統合を前提として，従来の項論理よりも高い表現能力を持つ独自の「項表示言語」を提示し，非公理的項論理として，項表示言語で表される3つのクラスの知識によって実施可能な5つの推論の型を形式的に定義した．
    \item 関連文献から採取した議論に対して非公理的項論理を適用して分析を行うことで，非公理的項論理の形式的記述力と説明力に関し，定性的な実証を与えた．
\end{enumerate}}
{\begin{enumerate}
     \item Based on the premise of integrating the syntactic theory of discrete systems and the semantics of continuous systems, we propose a unique "term representation language" that has higher expressive power than conventional term logic, and as a non-axiomical term logic, it is a term representation language. We formally defined five types of reasoning that can be performed by three classes of knowledge represented.
     \item By applying non-axiomatic term logic to arguments collected from related literature and analyzing them, we have provided qualitative proof of the formal descriptive power and explanatory power of non-axiomatic term logic.
\end{enumerate}}
{\begin{enumerate}
    \item Based on the premise of integrating the syntactic theory of discrete systems and the semantics of continuous systems, we propose a unique ``term representation language'' that has higher expressive power than conventional term logic, and as a  non-axiomatic term logic, it is a term representation language. We formally defined five types of reasoning that can be performed by three classes of knowledge represented.
    \item By applying non-axiomatic term logic to arguments collected from related literature and analyzing them, we have provided qualitative proof of the formal descriptive power and explanatory power of non-axiomatic term logic.
\end{enumerate}}
{\begin{enumerate}
    \item Based on the premise of integrating a discrete syntactic system and a continuous semantic system, we propose ``Term Representation Language'' (TRL) that has higher expressive power than conventional term logic.
    As ``Non-Axiomatic Term Logic'' (NATL), 
    we formally define five types of reasoning that can be performed on three classes of knowledge represented by using TRL.
    \item By applying NATL to the three arguments collected from related literature and analyzing them, we have provided a qualitative proof of the formal descriptive power and explanatory power of NATL.
\end{enumerate}}

\section{Term logic, Predicate logic, and Non-Axiomatic Logic}\label{sec:logics}

\xxparatran{アリストテレスに遡る三段論法は，大雑把に言えば，}{Roughly speaking, the syllogism that goes back to Aristotle}{Roughly speaking, the syllogism which goes back to Aristotle are as below,
}
{
Consider sentences (propositions) that have one of the following four patterns $A,I,E,O$:
\begin{itemize}
    \item[$A$:]\ all $X$ are $Y$.
    \item[$I$:]\ some $X$ are $Y$.
    \item[$E$:]\ all $X$ are not $Y$.
    \item[$O$:]\ some $X$ are not $Y$.
\end{itemize}
Roughly speaking, the syllogism that goes back to Aristotle distinguishes between valid and invalid inferences of the type that concludes one proposition from two propositions.\footnote{This explanation is more traditionally logical than Aristotle's own. See \cite{Lukasiewicz51, 
Englebretsen96} etc. for the difference between Aristotle's syllogism and traditional logic.}
Here, the symbols $X$ and $Y$ that appear in the propositions are called terms (names) and represent specific categories and properties such as ``human'', ``animal'', and ``immortality''.
For example, from the two propositions $A_1$: ``All humans are mammals'' and $A_2$: ``All mammals are animals'', we conclude $ A_3 $: ``All humans are animals''.
}
\xxparatran{の4つの型のいずれかを持つ文（命題）を考え，
2つの命題から1つの命題を結論する型の推論について，
妥当な推論と妥当でない推論を区別するものである\footnote{この説明はアリストテレス自身のものよりは伝統論理的である．アリストテレスの三段論法と伝統論理との違いについては\cite{Lukasiewicz51,Noya94,Englebretsen96}等を参照されたい．}．
このとき命題文に現れる記号$X$と$Y$が項（名辞）とよばれ，「人間」「動物」「不死」など特定のカテゴリや性質を表す．
例えば，$A_1$:「全ての人間は哺乳類である」，$A_2$:「全ての哺乳類は動物である」の2つの命題から，$A_3$:「全ての人間は動物である」を結論する．
このように，項論理は命題の中の一定の構造を表現することができる．
\citet{Lukasiewicz51}の形式化においては，命題間の論理はいわゆる命題論理に従う．}{
Consider a sentence (proposition) that has one of the four types of About type inference that concludes one proposition from two propositions, it distinguishes between valid and invalid reasoning. \footnote{This explanation is more traditional logic-flavored than Aristotle's own. See \cite{Lukasiewicz51, Noya94, 
Englebretsen96} etc. for the differences between Aristotle's syllogism and traditional logic.}.
At this time, the symbols $ X $ and $ Y $ that appear in the propositional sentence are called terms (names) and represent specific categories and properties such as "human", "animal", and "immortality".
For example, from the two propositions $ A_1 $: "All humans are mammals" and $ A_2 $: "All mammals are animals", we conclude $ A_3 $: "All humans are animals". do.
In this way, term logic can express a certain structure in a proposition.
In the formalization of \citet{Lukasiewicz51}, the logic between propositions follows the so-called propositional logic.}{
Considering sentences (propositions) that have one of four types, it distinguishes between valid and invalid inferences of the type that concludes one proposition from two propositions.\footnote {This explanation is more traditionally logical than Aristotle's own. See \cite{Lukasiewicz51, Noya94, Englebretsen96} etc. for the difference between Aristotle's syllogism and traditional logic.}
At this time, the symbols $X$ and $Y$ that appear in the propositions are called terms (names) and represent specific categories and properties such as ``human'', ``animal'', and ``immortality''.
For example, from the two propositions $A_1$: "All humans are mammals" and $A_2$: ``All mammals are animals'', we conclude $ A_3 $: ``All humans are animals''.
Thus, term logic can express certain structures in propositions. In the formalization of \cite{Lukasiewicz51}, the logic between propositions follows so-called propositional logic.
}
{
}


\xxparatran{述語論理も，命題の内部構造・命題間の内容的異同を形式的に記述することを可能にする道具立てである．
例えば「全ての人間は哺乳類である」という命題は，
\[(\forall x) (\mathrm{Human}(x) \Rightarrow \mathrm{Mammal}(x))\]
のように述語$\mathrm{Human}$と$\mathrm{Mammal}$を用いて記述できる．
述語論理においては，集合論に基づいて規定される「存在」，任意の数の変数をとることで様々な関係を定義できる「述語」，より柔軟なスコープの記述が可能な「量化子」（$\forall$と$\exists$）により，伝統論理が分析した2項間の4つの文型に囚われることなく，多様な命題，特に多重量化が必要となる数学的な定義・定理を，自然言語によらず厳密に表現することが可能になった．}{
Predicate logic is also a tool that makes it possible to formally describe the internal structure of propositions and the content differences between propositions.
For example, the proposition "All humans are mammals"
\[(\forall x)(\mathrm{Human}(x)\Rightarrow\mathrm{Mammal}(x))\] It can be described using the predicates $\mathrm {Human}$ and $\mathrm {Mammal}$.
In predicate logic, ``existence'' is defined based on set theory, ``predicate'' can define various relationships by taking any number of variables, and ``quantifier'' ($\forall$ and $\exists$) allows various propositions, especially mathematical definitions and theorems that require multiple weights, to be expressed in natural language without being bound by the four sentence patterns between two terms analyzed by traditional logic. It has become possible to express precisely without depending on natural language.}{
Predicate logic is also a tool that makes it possible to formally describe the internal structure of propositions and the different content between propositions. For example, the proposition ``All humans are mammals'',
\[(\forall x) (\mathrm{Human}(x) \Rightarrow \mathrm{Mammal}(x))\]can be described using the predicates $\mathrm {Human}$ and $\mathrm {Mammal}$. 
In predicate logic, ``existence'' which is specified based on set theory, ``predicates'' which can define various relations by taking any number of variables, and ``quantifiers''($\forall$ and $\exists$) which allow a more flexible description of scope, make it possible to strictly express various propositions, especially mathematical definitions and theorems that require weighting without natural language, and not be confined to the four sentence patterns between two terms that traditional logic has analyzed.
}
{
Predicate logic is also a tool that makes it possible to formally describe the internal structure of propositions and the different content between propositions. For example, the proposition ``All humans are mammals'',
\[(\forall x) (\mathrm{Human}(x) \Rightarrow \mathrm{Mammal}(x))\]can be described using the predicates $\mathrm {Human}$ and $\mathrm {Mammal}$. 
In predicate logic, ``existences'' (or individuals) which are specified based on set theory, ``predicates'' which can define various relations by taking any number of variables, and ``quantifiers''($\forall$ and $\exists$) which allow a more flexible description of scope, make it possible to strictly express various propositions, especially mathematical definitions and theorems that require multiple quantification, without natural language, and not be confined to the four sentence patterns between two terms that traditional logic has analyzed.}

\xxparatran{その代償として，述語論理で命題を表現するにあたっては，必ずしも直感的とは言い難い変換が要求される．述語論理では必ず個別の存在に立ち戻ってそれらの存在間の関係として命題を記述しなければならず，先の例で言えば，人間と哺乳類という概念間の関係を記述するにあたって，「全ての$x$について，それが人であるならばそれは哺乳類である」というような翻訳を要する.}{
The price is that when expressing a proposition with predicate logic, a transformation that is not always intuitive is required. In predicate logic, we must always go back to individual beings and describe the proposition as the relationship between those beings. In the previous example, in describing the relationship between the concepts of humans and mammals, "For all $x$, if it is a human, it is a mammal."}{
The price is that when expressing a proposition with predicate logic, a transformation that is not always intuitive is required. In predicate logic, we must always go back to individual beings and describe the proposition as the relationship between those beings. In the previous example, when describing the relationship between the concepts of human and mammal, we need a translation such as ``For all $x$, if it is a human, it is a mammal.''}
{The price is that when expressing a proposition with predicate logic, a transformation that is not always intuitive is required. In predicate logic, we must always go back to individual beings and describe the proposition as the relationship between those beings. In the previous example, when describing the relationship between the concepts of human and mammal, we need a translation such as ``For all $x$, if it is a human, it is a mammal.''}

\xxparatran{\citet{Sommers82,SommersEnglebretsen00}はこのような不自然さを嫌い，述語論理における個体変数を使用せずに，より自然に自然言語の中にある論理関係を形式的に表現できる枠組みを提案した．
これとは独立に，本邦においても\citet{Morita87}が，
述語論理が外延的であり項論理が内包的であることを強調
した上で，内包的な記述が直接可能な知識表現のための論理体系\footnote{
森田らは，モンタギュー意味論\cite{Thomason74}が，
述語論理体系の上に内包と高階の概念を導入したために非常に複雑になってしまったと指摘している．
\citet{Munemiya96}は，モンタギュー意味論が，述語論理に基づく外延主義故に，
そもそも「概念」を扱うことができない（動名詞や不定詞を用いた陳述に対して無力）
と指摘している．
}
の構築を目指した．
\citet{Nishihara94}はこれに動詞の取り扱いを導入して表現力を高めた．
\citet{Moss10}は西原らの提案に対しいくつかの改良を試みている．}{
\citet{Sommers82,SommersEnglebretsen00} dislikes such unnaturalness and proposes a framework that can formally express logical relationships in natural language more naturally without using individual variables in predicate logic.
Independently of this, \citet {Morita87} is also used in Japan.
Emphasize that predicate logic is extensional and term logic is intensional
Then, a logical system for knowledge representation that can be directly described inclusively \footnote{
Morita et al. Point out that Montague semantics \cite{Thomason74} has become very complicated due to the introduction of the concept of comprehension and higher order on the predicate logic system. \citet{Munemiya96} cannot deal with "concepts" in the first place because Montague semantics is intensional based on predicate logic (powerless for statements using gerunds and infinitives)
It is pointed out. } Aimed at the construction.
\citet{Nishihara94} introduced the handling of verbs into this to enhance its expressiveness.
\citet{Moss10} is trying some improvements to Nishihara et al.'S proposal.}{
\citet{Sommers82,SommersEnglebretsen00} dislikes such unnaturalness and proposes a framework that can formally express logical relationships in natural language more naturally without using individual variables in predicate logic.
Independently of this, \citet{Morita87} in Japan, emphasized the fact that predicate logic is extensional and term logic is intensional, and aimed to construct a logical system\footnote{Morita et al. point out that Montague semantics\cite {Thomason74} became very complicated because it introduced the concepts of implication and higher order on top of the predicate logic system, and \citet {Munemiya96} points out that Montague semantics cannot handle ``concepts'' in the first place because it is intensional based on predicate logic (powerless for statements using gerunds and infinitives).
} for knowledge representation in which intensional descriptions are directly possible.
\citet{Nishihara94} introduced the handling of verbs into this to enhance its expressiveness.
\citet{Moss10} is trying some improvements to Nishihara et al.'s proposal.
}
{
Sommers~\cite{Sommers82,SommersEnglebretsen00} dislikes such unnaturalness and proposes a framework that can formally express logical relationships in natural language more naturally without using individual variables in predicate logic.
Independently of this, \citet{Morita87}, emphasized the fact that predicate logic is extensional and term logic is intensional,\footnote{Morita et al. point out that Montague semantics \cite{Thomason74} became very complicated because it introduced the concepts of implication and higher orders on top of the predicate logic system, and \citet{Munemiya96} points out that Montague semantics is not capable of handling ``concepts'' (i.e., powerless for statements using gerunds and infinitives) in the first place because of its extensionalism based on predicate logic .
} and aimed to construct a logical system for knowledge representation in which intensional descriptions are directly possible.
\citet{Nishihara94} introduced the handling of verbs into this to enhance its expressiveness.
\citet{Moss10} is trying some improvements to Nishihara et al.'s proposal.}

\xxparatran{これらの項論理研究者の主な関心は自然言語の意味表示にあると思われる．
一方で，彼らが提示するものが論理体系として健全性と完全性を備えることにも関心を向けている．つまり自然言語を扱いつつも，公理主義の立場であり，
論理体系が少数の公理から出発して妥当な結論のみを推論でき（健全性），妥当な結論を全て推論できる（完全性）ことを求める．
これは数理論理あるいは計算機科学の立場としては真っ当である．}{
The main interest of these term logic researchers seems to be in the meaning display of natural language.
On the other hand, we are also interested in the fact that what they present has soundness and completeness as a logical system. In other words, while dealing with natural language, it is in the position of axiomism.
We request that the logical system starts from a small number of axioms and can infer only valid conclusions (soundness), and can infer all valid conclusions (completeness).
This is true from the standpoint of mathematical logic or computer science.
}{
The main interest of these term logic researchers seems to be in the semantic representation of natural language.
On the other hand, we are also interested in the fact that what they present has soundness and completeness as a logical system. In other words, while dealing with natural language, it is in the position of axiomism.
We request that the logical system starts from a small number of axioms and can infer only (soundness) and all (completeness) valid conclusions.
From an academic point of view, this is a mathematical logic or computer science.
}
{
The main interest of these term logic researchers seems to be in the semantic representation of natural language.
On the other hand, they are also interested in what they present has soundness and completeness as a logical system. 
In other words, while dealing with natural language, they take the position of axiomism.
}

\xxparatran{これに対し\citet{Wang94,Wang13}は，知能とはなにかという考察から出発し\footnote{%
Wangは，知能を「不十分な知識とリソースの元で環境に適応しうまく機能するための能力」として定義している．}，
数理論理の有用性を認めた上で\footnote{NAL自体は数理論理で記述されている．}，
汎用的な知能としての推論に対して非公理主義の重要性を主張した\footnote{%
物理学と生物学ではそれぞれの世界の法則・原理が異なるように，
計算機科学と（汎用）人工知能もそれぞれ異なる法則・原理に基づく，という仮説に立脚している．
}．
すなわち，完全な知識とリソースの存在を前提とする公理主義に基づくアプローチは，
知能のモデルとはなりえないという立場であり，
その視点から有用な推論の実現に取り組む上で，項論理的な知識表現方法が持つ特性の有用性に着目している．}{
On the other hand, \citet {Wang94, Wang13} starts from the consideration of what intelligence is, \footnote {%
Wang defines intelligence as "the ability to adapt and function well with inadequate knowledge and resources." },
After recognizing the usefulness of mathematical logic, \footnote{NAL itself is described in mathematical logic.},
He argued the importance of non-axiomism to reasoning as general intelligence \footnote{%
Just as the laws and principles of each world are different in physics and biology,
It is based on the hypothesis that computer science and (general purpose) artificial intelligence are also based on different laws and principles.
}.
In other words, they are of the position that an axiomatic approach that assumes the existence of perfect knowledge and resources cannot serve as a model of intelligence. We focus on the usefulness of the characteristics of knowledge representation methods.}
{
On the contrary, \citet{Wang94,Wang13} started from the consideration of what intelligence is \footnote{Wang defines intelligence as ``the ability to adapt and function well in an environment under inadequate knowledge and resources''.}, acknowledged the usefulness of mathematical logic \footnote{NAL itself is described in mathematical logic.}, and insisted on the importance of non-axioms for reasoning as a general-purpose intelligence\footnote{Just as the laws and principles of the world are different in physics and biology,, computer science and (general-purpose) artificial intelligence are also based on their different laws and principles.}.
In other words, the axiomatic approach, which assumes the existence of complete knowledge and resources, cannot serve as a model of intelligence. From this perspective, we focuses on the usefulness of the properties of term-logic methods of knowledge representation in realizing useful inferences.
}
{
On the contrary, \citet{Wang94,Wang13} started from the consideration of what intelligence is,\footnote{Wang defines intelligence as 
``the ability for a system to adapt to its environment and to work with insufficient knowledge and resources''. 
} acknowledged the usefulness of mathematical logic,\footnote{NAL itself is described in mathematical logic.} and insisted on the importance of non-axiomism for reasoning as a general-purpose intelligence.\footnote{\citet{Wang13} argues that, just as the laws and principles of the world are different in physics and biology, computer science and (general-purpose) artificial intelligence are also based on their different laws and principles.}
In other words, the axiomatic approach, which assumes the existence of complete knowledge and resources, cannot serve as a model of intelligence. From this perspective, he focuses on the usefulness of the properties of term-logic for knowledge representation in realizing useful inferences.
}

\xxparatran{Wangが提案する非公理的論理（Non-Axiomatic Logic; 以下NAL）は，汎用的な推論システムの実現を念頭に設計されており，
Sommersや森田・西原らのように自然言語文の意味表示に焦点を合わせているわけではない．
NALでは，推論規則は与えられるものの，公理（絶対的な命題）は存在しない．
同じ有限の知識源を元に推論しても，時間というリソースも制限されているため，
得られる推論結果は異なりうるし，更新され得る．
加えて，演繹以外の「弱い推論規則」が用意されており，妥当ではない結論を導く可能性がはじめから許容されている．
命題論理にも従わない．NALにおける命題間の含意関係（$P \Rightarrow Q$）は，
$P$という命題から$Q$という命題を導出できること（つまり$P \vdash Q$）と同義と定義され，
$\lnot P \lor Q$と同値ではない．
NALにおける「真」の概念は，過去の経験に照らして妥当といえること，である．
したがって，「人が人でないなら私は神である」というような無意味な命題は，
命題論理では恒真であるが，NALではそうではない．}{Wang's proposed non-axiomatic logic (NAL) is designed with the realization of a general-purpose inference system in mind.
It does not focus on the meaning display of natural language sentences like Sommers and Morita / Nishihara.
In NAL, inference rules are given, but axioms (absolute propositions) do not exist.
Even if inference is based on the same finite knowledge source, the inference results obtained can vary and can be updated due to the limited resource of time.
In addition, "weak reasoning rules" other than deduction are prepared, and the possibility of drawing invalid conclusions is allowed from the beginning.
It does not follow propositional logic. 
The entailment relation between propositions in NAL ($P \Rightarrow Q$) is defined as the ability to derive the proposition $Q$ from the proposition $P$ (that is, $P \vdash Q$), and Not equivalent to $\lnot P \lor Q$.
Not equivalent to $\lnot P \lor Q$.
The concept of "truth" in NAL is that which can be said to be valid in the light of past experience.
Therefore, nonsensical propositions such as ``If a man is not a man, then I am a god'' are true in propositional logic, but not in NAL.}
{
Non-Axiomatic Logic (referred to as NAL) proposed by Wang was designed with the realization of a general-purpose reasoning system in mind and does not focus on semantic representation of natural language sentences as Sommers, Morita, Nishihara, and others have done.
In NAL, inference rules are given, but axioms (absolute propositions) do not exist.
Even if inference is based on the same finite knowledge source, the inference results obtained can vary and can be updated due to the limited resource of time.
In addition, ``weak inference rules'' other than deduction are provided, and the possibility of drawing invalid conclusions is allowed from the beginning.
It does not follow propositional logic either.
The entailment relation between propositions in NAL ($P \Rightarrow Q$) is defined as the ability to derive the proposition $Q$ from the proposition $P$ (that is, $P \vdash Q$), and not equivalent to $\lnot P \lor Q$.
The concept of ``truth'' in NAL is what is reasonable in view of past experience.
Therefore, meaningless propositions such as ``If a man is not a man, then I am a god'' are true in propositional logic, but not in NAL.
}
{
Non-Axiomatic Logic (referred to as NAL) proposed by Wang is designed with the realization of a general-purpose reasoning system in mind and does not focus on semantic representation of natural language sentences as Sommers, Morita, Nishihara, and others have done.
In NAL, inference rules are given, but axioms (absolute propositions) do not exist.
Even if reasoning is based on the same finite knowledge source, the results obtained can vary and can be updated due to the limited resource of time.
In addition, ``weak inference rules'' other than deduction are provided, and the possibility of drawing invalid conclusions is allowed from the beginning.
It does not follow propositional logic either.
The entailment relation between propositions in NAL ($P \Rightarrow Q$) is defined as the ability to derive the proposition $Q$ from the proposition $P$ (that is, $P \vdash Q$), and not equivalent to $\lnot P \lor Q$.
The concept of ``truth'' in NAL is what is reasonable in view of past experience.
Therefore, meaningless propositions such as ``If a man is not a man, then I am a god'' are true in propositional logic, but not in NAL.
}

\section{Non-Axiomatic Logic}\label{sec:nal}

\xxparatran
{非公理的論理NAL~\cite{Wang13}の概要を，構文論，意味論，推論規則の側面から説明し，
NALの課題を議論する．}
{The outline of the axiom logic NAL ~ \cite{Wang13} is explained from the aspects of syntax, semantics, and inference rules, and the problems of NAL are discussed.}
{This section explains the outline of non-axiomatic logic (NAL)~\cite{Wang13} from the aspects of syntax, semantics, and inference rules, and then discusses the tasks of NAL.
}
{
This section explains the outline of Non-Axiomatic Logic (NAL)~\cite{Wang13} from the aspects of syntax, semantics, and inference rules, and then discusses the issues of NAL.}

\subsection{Syntax}\label{sec:nal_syntax}

\xxparatran
{$S \to P$という形式で，項$S$と$P$を繋辞
（copula）$\to$で結ぶことにより，「$S$は$P$である」という命題を表す．この形の命題を陳述（statement）とよぶ．
この陳述構文を基本とし，NAL言語（Narses）の表現能力を高めるための多数の拡張がなされているが，
本稿での議論に関係するものだけを以下に取り上げる．}
{The proposition ``$S$ is $P$'' is expressed by joining the terms $S$ and $P$ with the copula $\to$ in the form $S \to P$. This form of proposition is called a statement.
Based on this statement syntax, many extensions have been made to enhance the expressiveness of the NAL language (Narses).
Only those that are relevant to the discussion in this paper are discussed below.}
{The proposition ``$S$ is $P$'' is expressed by joining the terms $S$ and $P$ with the copula $\to$ in the form $S \to P$.. This form of proposition is called ``statement''. Based on this statement syntax, a number of extensions have been made to increase the expressive power of the NAL language (Narses), but only those that are relevant to the discussion in this paper are discussed below.
}
{
A proposition ``$S$ is $P$'' is expressed by joining the terms $S$ and $P$ with the copula $\to$ in the form $S \to P$. This form of proposition is called a statement.
Based on this statement syntax, many extensions have been made to enhance the expressiveness of the NAL language (Narses).
Only those that are relevant to the discussion in this paper are explained below.}

\xxparatran{前述のように，陳述$P$が陳述$Q$を含意するとき$P \To Q$と書く．
$P \To Q$も（高階の）陳述であり，陳述も項である．}
{As mentioned above, when the statement $P$ implies the statement $Q$, write $P \To Q$. $P \To Q$ is also a (higher) statement, and the statement is also a term.}
{As mentioned before, when the statement $P$ implies the statement $Q$, write $P \To Q$. $P \To Q$ is also a (higher-order) statement and statement is also a term.
}
{
As mentioned above, when statement $P$ implies statement $Q$, it is written as $P \To Q$. $P \To Q$ is also a (higher-order) statement, and the statement is also a term.}

\xxparatran
{３項以上の関係を表現したいときは，$U \times V \to R$のように記述する．
例えば``water resolves salt''は，$\mathrm{water} \times \mathrm{salt} \to \mathrm{resolve}$
と表現する．
$U \times V$も1つの項（積項）として扱われ，$R$は関係項とよばれる．}
{If you want to express the relationship of three or more terms, write as $U \times V \to R$. For example, `` water resolves salt'' is expressed as $ \mathrm {water} \times \mathrm {salt} \to \mathrm {resolve}$. $U \times V$ is also treated as one term (product term), and $R$ is called a relational term.}
{When we want to express the relationship of three or more terms, write as $U \times V \to R$. For example, ``water resolves salt'' is expressed as $\mathrm{water} \times \mathrm{salt} \to \mathrm{resolve}$. $U \times V$ is treated as one term (product term) and $R$ is called the ``linkage term''.
}
{
When expressing the relationship of three or more terms, it is expressed like $U \times V \to R$. For example, ``water resolves salt'' is expressed as $\mathrm{water} \times \mathrm{salt} \to \mathrm{resolve}$. $U \times V$ is treated as one term (product term) and $R$ is called a relational term.}

\subsection {Semantics (experience-grounded semantics)}

\xxparatran
{繋辞$\to$が項$S$と$P$の間の継承関係（上位下位関係）を表すという考えを基本に\footnote{転換の推論規則により，NALでは$S\to P$から$P\to S$も推論できる．
後者において$\to$が継承を表すといえるのかは疑問が残る．
伝統論理において，転換は$I,E$型の陳述への適用は妥当であるが，他の2つへの適用は妥当ではない．
}，
陳述の意味論（真偽）が，項の外延（下位の項）と内包（上位の項）の集合で定義される．
すなわち，$S \to S$が常に成り立つと定めた上で，
ある項$T$の外延集合$T^E$を$T^E = \{x|(x \in V_K) \land (x \to T)\}$，
内包集合$T^I$を$T^I = \{x|(x \in V_K) \land (T \to x)\}$としたとき，
$S \to P$の真偽は，$S^E \subseteq P^E$かつ$P^I \subseteq S^I$であるかどうかと同値であるとされる．
つまり$|S^E - P^E| + |P^I - S^I| = 0$であれば真である．
陳述の一つ一つが観測（経験）であるという意味で，これを経験接地意味論とよぶ．}
{Based on the idea that the concatenation $\to$ represents the inheritance relationship (hyponymy and hyperactivity relationship) between the terms $ S $ and $ P $, \footnote{By the inference rule of conversion, from $S\to P$ in NAL You can also infer $P\to S$.
In the latter case, it remains doubtful whether $ \ to $ represents inheritance.
In traditional logic, conversion is valid for $ I, E $ type statements, but not for the other two.
},
The semantics (true or false) of a statement is defined by a set of extension (lower term) and comprehension (upper term) of a term.
That is, after stipulating that $S \to S$ always holds.
The extension set $T^E$ of a term $T^E = \{x|(x \in V_K) \land (x \to T)\}$,
When the comprehension set $T^I = \{x|(x \in V_K) \land (T \to x)\}$
The truth of $S\to P$ is considered to be equivalent to whether it is $S^E \subseteq P^E$ and $P^I \subseteq S^I$.
In other words, it is true if $|S^E - P^E| + |P^I - S^I| = 0$.
In the sense that each statement is an observation (experience), this is called empirical grounding semantics.}
{Based on the thinking that using copula $\to$ express the inheritance relationship (hyponymy) between $S$ and $P$ \footnote{According to the influence rules of conversion, the $P \to S$ can be inferred from $ S \to P $ in NAL. It still remain doubt that whether $ \to $ can represents inheritance in $P \to S$. In traditional logic, it is reasonable to apply conversion for type $ I, E $ of statements whereas it is not suitable for other two types of statement.}, the semantics (true or false) of a statement is defined by a set of extensions (lower terms) and intensions (upper terms) of a term. That is, if $S \to S$ is always true, and then the extension set $T^I$ of a term $T$ is $T^I = \{x|(x \in V_K) \land (T \to x)\}$ and intension set $T^I$ is $T^I = \{x|(x \in V_K) \land (T \to x)\}$, then the truth of $S \to P$ is said to be equivalent to whether $S^E \subseteq P^E$ and $P^I \subseteq S^I$. In other words, it is true if $|S^E - P^E| + |P^I - S^I| = 0$. In the sense that each statement is an observation (experience), this is called experience-based semantics.
}
{
Based on the idea that copula $\to$ expresses the inheritance relationship (hyponymy) between $S$ and $P$,\footnote{By the influence rule of conversion, the $P \to S$ can be inferred from $ S \to P $ in NAL. It still remain doubt that whether $ \to $ can represents inheritance in $P \to S$. In traditional logic, it is reasonable to apply conversion for type $ I, E $ of statements whereas it is not suitable for other two types of statement.}
the semantics (true or false) of a statement is defined by a set of extensions (lower terms) and intensions (upper terms) of a term. 
That is, given that $S \to S$ is always true, and the extension set $T^E$ of a term $T$ is $T^E = \{x|(x \in V_K) \land (x \to T)\}$ and intension set $T^I$ is $T^I = \{x|(x \in V_K) \land (T \to x)\}$, then the truth of $S \to P$ is said to be equivalent to whether $S^E \subseteq P^E$ and $P^I \subseteq S^I$. In other words, it is true if $|S^E - P^E| + |P^I - S^I| = 0$. In the sense that each statement is an observation (experience), this is called experience-grounded semantics.}

\xxparatran
{ただし前述の真偽の定義は，Wangが継承論理（inheritance logic; IL）とよぶ理想状況（不確実性が存在しない状況）においてのみ成り立つ．
NALでは不確実性を前提としているため，
前記の同値関係を不確実性を取り込むように一般化している．
しかしその詳細は本論文に直接関係しないため説明を割愛する．}
{However, the above definition of truth and falsehood only in an ideal situation (a situation in which there is no uncertainty) called Wang's inheritance logic (IL). Since NAL presupposes uncertainty, the above equivalence relation is generalized to incorporate uncertainty. However, the details are not directly related to this paper, so the explanation is omitted.}
{However, the above definition of truth and falsehood can only work in an ideal situation (a situation in which there is no uncertainty) called inheritance logic by Wang. Since the premise of NAL is uncertainty, the above equivalence relation is generalized to incorporate uncertainty. However, such details are not related to this paper directly, so the explanation is omitted.
}
{
However, the above definition of truth and falsehood can only work in an ideal situation (a situation in which there is no uncertainty) called inheritance logic by Wang. Since the premise of NAL is uncertainty, the above equivalence relation is generalized to incorporate uncertainty. However, the details of this generalization are not related to this paper directly, so the explanation is omitted.}

\subsection{Inference rules}\label{sec:rules}

\xxparatran
{主なNALの推論は2つの前提から1つの結論を導く三段論法的規則によって行われるが，
他に，否定や転換など1つの陳述だけを取る規則，積項と関係項の間の構造変換を行う規則
なども定義されている．
ここではNALにおける三段論法的規則の中でも，
演繹（deduction），帰納（induction），アブダクション（abduction）の3規則だけを取り上げる．
すなわち，}
{The main reasoning in NAL is based on syllogistic rules that lead to one conclusion from two premises, but there are also rules that take only one statement, such as negation and conversion, and structural transformations between product terms and relation terms. Rules for doing so are also defined.
Among the syllogistic rules in NAL, only three rules, deduction, induction, and abduction, are taken up here.
That is,}
{The main reasoning in NAL is based on syllogistic rules that lead to one conclusion from two premises, but there are also rules that take only one statement, such as negation and conversion, and structural transformations between product terms and relation terms. Rules for doing so are also defined. Among the syllogistic rules in NAL, only three rules, deduction, induction, and abduction, are taken up here. That is,
}
{
The main reasoning in NAL is based on syllogistic rules that lead to one conclusion from two premises, but there are also rules that take only one statement, such as negation and conversion, and rules for structural transformations between product terms and relation terms. 
Among the syllogistic rules in NAL, only three rules, deduction, induction, and abduction, are taken up here. That is,}
\begin{align}
\nonumber  deduction   &:  \{S \to M, M \to P\} \vdash S \to P, \\
\nonumber  induction   &:  \{S \to M, S \to P\} \vdash M \to P, \\
\nonumber  abduction   &:  \{M \to P, S \to P\} \vdash S \to M. 
\end{align}

\xxparatran
{帰納とアブダクションは，演繹の前提と結論を入れ替えた形になっている．
Wangはそれぞれの規則ごとに，2つの前提陳述の真理値から結論陳述の真理値を与える
真理値関数を提示している．}
{Induction and abduction are forms in which the premises and conclusions of deduction are interchanged.
For each rule, Wang presents a truth-value function that gives the truth-value of the conclusion statement from the truth-values of the two premise statements.}
{Induction and abduction are forms in which the premises and conclusions of deduction are interchanged.
For each rule, Wang presents a truth-value function that gives the truth-value of the conclusion statement from the truth-values of the two premise statements.
}
{
Induction and abduction are forms in which the premises and conclusions of deduction are interchanged. For each rule, Wang presents a truth-value function that gives the truth-value of the conclusion statement from the truth-values of the two premise statements.}

\xxparatran
{帰納は事例の一般化である．
様々な自然言語推論に関するbAbi tasks~\cite{bAbi}のTask 16 Basic Inductionは，
例えば，
A: Lily is a swan.
B: Lily is white.
C: Greg is a swan.
という3文の入力が与えられた際のWhat color is Greg?という質問に対して white と
答えることを求めている．
ここで求められている推論は，$S$: Lily, $M$: a swan, $P$: white という対応において，
文A, Bから，a swan is white, すなわち$M \to P$という知識を帰納することである．
それと文CからGreg is white. を演繹すれば，正しい回答ができる．}
{Induction is a generalization of cases.
Task 16 Basic Induction of bAbi tasks~\cite{bAbi} related to various natural language inference
for example,
A: Lily is a swan.
B: Lily is white.
C: Greg is a swan.
In response to the question ``What color is Greg?''
I want to answer.
The inference sought here is that in the correspondence $S$: Lily, $M$: a swan, $P$: white, from sentences A and B, a swan is white, i.e., $M \to P$. It is the induction of knowledge.
And if we deduce Greg is white from sentence C, we get the correct answer.
}{Induction is a generalization of cases. Task 16 Basic Induction of bAbi tasks~\cite{bAbi} is related to various natural language inferences, for example, 
A: Lily is a swan.
B: Lily is white.
C: Greg is a swan.
For these three input sentences, ``white'' should be the answer for ``What color is Greg?''. The inference required here is ``a swan is white'', which is in the correspondence of $S$: Lily, $M$: a swan, $P$: white, from the sentences A and B. That is, the induction of the knowledge $M \to P$. And if we deduce ``Greg is white'' from sentence C, we get the correct answer.}
{
Induction is a generalization of cases. Consider an example from Task 16 Basic Induction of bAbi tasks~\cite{bAbi}: 
A: Lily is a swan;
B: Lily is white;
C: Greg is a swan.
For these three input sentences, ``white'' should be the answer for the question ``What color is Greg?'' 
The inference required from the sentences A and B is ``a swan is white.''
With the correspondences of $S$: Lily, $M$: a swan, $P$: white, this is the induction of the knowledge $M \to P$. And if we deduce ``Greg is white'' from sentence C and the last inferred sentence, we get the expected answer.}

\xxparatran{アブダクションは，パースが見出した「経験から知識を得る」ためのもう一つの形の推論であり，その本質は解釈（説明）の論理\cite{Arima14}である．
誤った推論結果を容易に導きうるが，
それが人間の思考の柔軟性・創造性にも寄与している．
パース自身が示したアブダクションの例は以下のようなものである\cite{Arima14}.}
{Abduction is another form of reasoning that Peirce found for ``obtaining knowledge from experience,'' and its essence is the logic of interpretation (explanation) \cite{Arima14}.
It can easily lead to erroneous inference results, which contributes to the flexibility and creativity of human thinking.
An example of abduction shown by Perth Peirce is the following\cite {Arima14}.}
{Abduction is another form of reasoning that Peirce found for ``obtaining knowledge from experience,'' and its essence is the logic of interpretation (explanation) \cite{Arima14}. Although it can easily lead to erroneous inference results, it contributes to the flexibility and creativity of human thinking. An example of abduction shown by  Peirce is the following \cite{Arima14}.
}
{
Abduction is another form of reasoning that Peirce found for ``obtaining knowledge from experience,'' and its essence is the logic of interpretation (explanation) \cite{Arima14}. Although it can easily lead to erroneous inference results, it contributes to the flexibility and creativity of human thinking. An example of abduction shown by  Peirce is the following \cite{Arima14}:}

\xxparatran
{\begin{itemize}
    \item[規則]： この袋から出る豆は全て白い．
    \item[結果]： ここにある豆は白い．
    \item[仮説]： ここにある豆はこの袋から出たものだ．
\end{itemize}}
{\begin {itemize}
     \item [Rule]: All the beans that come out of this bag are white.
     \item [Result]: The beans here are white.
     \item [Hypothesis]: The beans here are from this bag.
\end {itemize}}
{\begin {itemize}
     \item[Rule]: All the beans from this bag are white.
     \item[Result]: The beans here are white.
     \item[Hypothesis]: The beans here are from this bag.
\end {itemize}}
{\begin{description}
     \item[Rule]: \hspace{0.83cm} All the beans from this bag are white.
     \item[Result]: \hspace{0.57cm} The beans here are white.
     \item[Hypothesis]: The beans here are from this bag.
\end{description}}

\xxparatran
{これは，$M$:この袋から出る豆，$P$:白い，$S$:ここにある豆，という対応において，
まさに$\{M \to P, S \to P\} \vdash S \to M$ の形になっている．}
{This corresponds to $M$: beans from this bag, $P$: white beans, $S$: beans here. exactly $\{M \to P, S \to P\} \vdash S \to M$.}
{That is, $M$: beans from this bag, $P$: white, $S$: beans here. The correspondence of this case is $\{M \to P, S \to P\} \vdash S \to M$.}
{With the correspondences of $M$: beans from this bag, $P$: white, $S$: beans here, this case is exactly in the form of $\{M \to P, S \to P\} \vdash S \to M$.}

\xxparatran
{この3つの規則は，継承繋辞$\to$を含意繋辞$\To$に置き換えても成立する．
アブダクションの説明はしばしば事態間の因果（含意）関係の例でもなされる．
例えば，「風が吹くと木が揺れる」という因果的知識をもって，「木が揺れている」
という結果の観測から「風が吹いている」という原因を推測をする，というものである．
$ \{M \To P, S \To P\} \vdash S \To M$という推論形式は，
一見この例にうまく当てはまらないように見えるが，
$S \To$の部分を文脈条件として背景化すると，$\{M \To P,  P\} \vdash M$となり，
所与の文脈のもとで観測した結果から原因を推定（解釈）する形になっていることがわかる．}
{These three rules hold even if the inheritance copula $\to$ is replaced with the implication copula $\To$.
The explanation of abduction is often given as an example of causal (implication) relationships between situations.
For example, with the causal knowledge that ``a tree sways when the wind blows'', ``a tree sways''.
From the observation of the result, the cause of ``the wind is blowing'' is inferred.
The inference form of $ \{M \To P, S \To P \} \vdash S \To M $ is
At first glance it doesn't seem to fit this example well,
When the $ S \ To $ part is used as the context condition, it becomes $\{M \To P,  P\} \vdash M$.
It can be seen that the cause is inferred (interpreted) from the observed results in a given context.}
{These three rules hold even if the inheritance copula $\to$ is replaced with the implication copula $\To$. The explanation of abduction is often given as an example of causal (implication) relationships between situations. For example, with the causal knowledge that ``a tree sways when the wind blows'', the result ``a tree sways'' is observed and the cause ``the wind blows'' is inferred. At first glance, the $ \{M \To P, S \To P\} \vdash S \To M$ inference form does not seem to apply well to this example, but when the $S \To$ part is used as the context condition, it becomes $\{M \To P,  P\} \vdash M$. It can be seen that the cause is inferred (interpreted) from the observed results in a given context.
}
{These three rules hold even if the inheritance copula $\to$ is replaced with the implication copula $\To$. The explanation of abduction is often given as an example of causal (implication) relationships between situations. For example, with the causal knowledge that ``a tree sways when the wind blows'', the result ``a tree sways'' is observed and the cause ``the wind blows'' is inferred. At first glance, the inference form of $ \{M \To P, S \To P\} \vdash S \To M$ does not seem to apply well to this example, but when the $S \To$ part is contexualized, it becomes $\{M \To P,  P\} \vdash M$. It can be seen that the cause is inferred (interpreted) from the observed results in a given context.
}

\subsection{Issues of Non-Axiomatic Logic} \label{sec:issues}

\xxparatran
{Wangが提示した非公理的論理NALは，
直感的な簡潔性を持ち人間的な内包的記述が容易（述語論理の適用に際して求められる「何を個別存在とし何を述語とするか」の判断はそれほど自明ではない），
アブダクションなどの非演繹的
推論を行う枠組みを備えている，など，
本研究の目的である「常に正しいとは限らないが特定の状況においては一定の有用性を持つ規則をもとにして人間が日常的に行っている推論」
のモデル化にあたって望ましいと思われる特徴を備えている．
また，自然論理\cite{vanBenthem08,MacCartney09}
とは異なり，あくまで記号推論のモデルであって，
自然言語のみを対象とするものではない．
自然論理は，自然言語の表層形をそのまま用いて推論を行うことを目指す．}
{The non-axiomatic logical NAL presented by Wang is intuitively concise and easy to describe intensively (the judgment of "what is an individual entity and what is a predicate" required when applying predicate logic is It is not so obvious), and it has a framework for non-deductive reasoning such as abduction. Desirable features for modeling the purpose of this research, ``inferences that humans make on a daily basis based on rules that are not always correct but have certain usefulness in specific situations'' It has
Also, natural logic \cite{vanBenthem08, MacCartney09} Unlike, it is just a model of symbolic reasoning.It is not just for natural language.
Natural logic aims to make inferences using the surface form of natural language as it is.}
{The non-axiomatic logical NAL presented by Wang is intuitively concise and easy to describe intensively (the judgment of ``what is an individual entity and what is a predicate'' required when applying predicate logic is not so obvious), and it has a framework for non-deductive reasoning such as abduction. Based on this, it has features that are considered desirable for modeling the purpose of this research, ``inferences that humans make on a daily basis based on rules that are not always correct but have certain usefulness in specific situations''. Also, unlike natural logic \cite{vanBenthem08, MacCartney09}, it is still a model of symbolic inference and not just for natural language. Natural logic aims to make inferences by using a surface form of natural language directly.
}
{
Non-Axiomatic Logic presented by Wang is intuitively concise and easy to describe intensional knowledge (the judgment of ``what is an individual entity and what is a predicate'' required when applying predicate logic is not so obvious), and it has a framework for non-deductive reasoning such as abduction. It has features that are considered desirable for modeling the purpose of this research, ``inferences that humans make on a daily basis based on rules that are not always correct but have certain usefulness in specific situations''. Also, unlike natural logic \cite{vanBenthem08, MacCartney09}, it is a general model of symbolic reasoning and not just for natural language. 
Natural logic aims to make inferences by using a surface form of natural language directly.}

\xxparatran
{一方でNALに内在する課題も指摘できる．
本節では，それらの課題とそれらに対するアプローチを提示する．}
{On the other hand, we can point out the problems inherent in NAL.
This section presents these issues and approaches to them.}
{On the other hand, we can point out the issues inherent in NAL. This section presents these issues and approaches to them.
}
{
On the other hand, we can point out some issues inherent in NAL. This section presents these issues and approaches to them.}

\subsubsection{Assignment of latent space semantic representations to terms}\label{sec:latent_repr}
\xxparatran
{
NALの推論過程はある項を別の項に代入あるいは置換する過程とみることができる（\cite{Wang13} p.22）．
NALにおけるこの代入処理は，項つまり記号の表層的一致や，記号間の類似性の知識（$S \leftrightarrow T$）
によって導かれることになるが，表層的一致による処理は精度を担保できるものの実用的な頑健性を得る点で問題があることは従来からの記号処理AIに共通する課題であったし，記号間の類似性を評価する関数を問題領域の特性を十分考慮して適切に設計することも容易ではなかった．}
{
The inference process of NAL can be regarded as the process of substituting or replacing one term with another (\cite {Wang13} p. twenty two). NAL This assignment process in this p.22 NAL is guided by the superficial match of terms or symbols and the knowledge of similarity between symbols ($ S \leftrightarrow T$), but the process by superficial match is Although the accuracy can be guaranteed, there is a problem in obtaining practical robustness, which is a problem common to conventional symbol processing AI, and the function to evaluate the similarity between symbols is sufficient for the characteristics of the problem domain. It was also not easy to properly design a function that evaluates the similarity between symbols by fully considering the characteristics of the problem domain.}
{
The inference process of NAL can be seen as the process of substituting or replacing one term with another (\cite{Wang13} p.22). This substitution processing in NAL is guided by the superficial match of terms or symbols and knowledge of similarity between symbols ($S \leftrightarrow T$). Although it can ensure accuracy, there is a problem in obtaining practical robustness, which is a common topic of symbol processing AI in the past. It is also not easy to properly design a function that evaluates the similarity between symbols by fully considering the characteristics of the problem domain.
}
{
The reasoning process of NAL can be seen as the process of substituting or replacing one term with another (\cite{Wang13} p.22). This substitution process in NAL is guided by the superficial match of terms or symbols and knowledge of similarity between symbols ($S \leftrightarrow T$). Although it can ensure accuracy, there is a problem in obtaining practical robustness, which is a common topic of symbol processing AI in the past. It is also not easy to properly design a function that evaluates the similarity between symbols by fully considering the characteristics of the problem domain.}

\xxparatran
{近年，記号の集合を特定の制約の下で多次元連続空間（潜在空間）内の点に写像する「埋め込み」（embedding）によって，空間内の位置関係で記号間の意味的関係性・類似性を評価することにより，自然言語処理や知識グラフ処理のタスクを高精度かつ頑健に行うことができるようになっている（例えば\cite{word2vec,transE}）．
NALにおいても各項に埋め込み表象を付与することで，推論の精度と頑健性を高めることが期待できる．
同様の考え方による定理証明の研究もある（例えば\cite{Arabshahi21}）.}
{In recent years, the semantic relationship/similarity between symbols through the positional relationship in the space has been investigated by ``embedding'', which maps a set of symbols to points in a multidimensional continuous space (latent space) under specific constraints. By evaluating , it is possible to perform natural language processing and knowledge graph processing tasks with high accuracy and robustness (for example, \cite {word2vec, transE}).
In NAL, we can expect to increase the accuracy and robustness of inference by giving each term an embedded representation. There is also research on theorem proving based on a similar idea (for example, \cite{Arabshahi21}).}{
In recent years, the semantic relationship/similarity between symbols through the positional relationship in the space has been investigated by ``embedding'', which maps a set of symbols to points in a multidimensional continuous space (latent space) under specific constraints. By evaluating , it is possible to perform natural language processing and knowledge graph processing tasks with high accuracy and robustness (for example, \cite {word2vec, transE}). In NAL, it is expected to increase the accuracy and robustness of inference by adding embedded representations to each term. There are also theorem proving researches based on the same idea ( for example, \cite{Arabshahi21}).}
{
In recent years, the semantic relationship/similarity between symbols has been investigated by ``embedding'' as the positional relationship in the space, which maps a set of symbols to points in a multidimensional continuous space (latent space) under specific constraints. 
By embedding, it has become possible to perform natural language processing and knowledge graph processing tasks with high accuracy and robustness (for example, \cite {word2vec, transE}). For NAL, it is expected to increase the accuracy and robustness of inference by adding embedded representations to each term. There are also theorem proving researches based on the same idea (for example, \cite{Arabshahi21}).}

\subsubsection{Training of reasoners based on reinforcement learning}\label{sec:reinforcement_on_reasoner}
\xxparatran
{
NALを元にしてWangが提案する推論システムNARSの推論アルゴリズムは非常にナイーブなものであり，無駄な推論を省き，有意な推論結果を実効性のある時間内に得ることは困難である．
NALにおける推論は処理対象となる陳述と適用する規則の選択の繰り返しであるが，これはある種の暗黙知的な技能であり，強化学習による推論器の訓練が有効であると思われる．
この点については，記号推論と強化学習の組み合わせに関する先行研究（\cite{Ichisugi20}など）で得られた知見を援用できるだろう．
前述した項別の潜在空間意味表象や，それらをグラフ畳み込みなどで集約したものには，強化学習における状態空間表現としての利用価値もあるだろう．}
{
The inference algorithm of the inference system NARS proposed by Wang based on NAL is very naive, and it is difficult to eliminate useless inference and obtain significant inference results within a practical time. Inference in NAL is a repetition of selection of statements to be processed and rules to be applied, but this is a kind of implicit intellectual skill, and training of reasoners by reinforcement learning seems to be effective.
On this point, we can use the knowledge obtained from previous research on the combination of symbolic reasoning and reinforcement learning (such as \cite{Ichisugi20}).
The above-mentioned latent space semantic representations for each term and those aggregated by graph convolution may be useful as state space representations in reinforcement learning.}
{
The inference algorithm of the inference system NARS proposed by Wang based on NAL is very naive, and it is difficult to eliminate useless inference and obtain significant inference results within a valid time. Inference in NAL is an iterative process of selecting statements to be processed and rules to be applied, which is a kind of tacit intellectual skill, and training the reasoner by reinforcement learning seems to be effective. In this regard, we can cite the knowledge obtained from the previous research related to the combination of symbolic reasoning and reinforcement learning (\cite {Ichisugi20}, etc.). The above-mentioned latent space semantic representations classified by terms and those aggregated by graph convolution might be valuable to represent as state space in reinforcement learning.
}
{
The algorithm of the reasoning system NARS proposed by Wang based on NAL is very naive, and it is difficult to eliminate useless inference and obtain significant inference results within a valid time. Inference in NAL is an iterative process of selecting statements to be processed and rules to be applied, which is a kind of tacit intellectual skill, and training the reasoner by reinforcement learning seems to be effective. 
The above-mentioned latent space semantic representations classified by terms and those aggregated by graph convolution might be valuable to represent as state space in reinforcement learning.}

\subsubsection{The continuity and prototype hypothesis of copula}\label{sec:cont_copula} 
\xxparatran
{
NALでは，継承繋辞・含意繋辞の他にも，主に表現能力上の要請から8つの繋辞が導入されている．
工学的な立場から見ればシステムを複雑化してしまっている懸念があり，
一方で科学的な立場からみればこれだけで十分なのかという疑念がある．}
{
In addition to inheritance and implication copula, NAL introduces 8 other copula mainly for expressive capacity requirements.
From an engineering point of view, there is a concern that the system is becoming complicated, while from a scientific point of view, there is a suspicion that this alone is sufficient.}
{
In NAL, in addition to the inherited copula and implied copula, eight types of copula were introduced mainly for the requirement of expressing ability. From an engineering point of view, there is a concern that the system has become complicated, while from a scientific point of view, there is a suspicion that this alone is sufficient.}
{
In NAL, in addition to the inherited copula and implied copula, eight types of copula were introduced mainly for the requirement of expressing ability. From an engineering point of view, there is a concern that the system has become complicated, while from a scientific point of view, there is a suspicion that this alone is sufficient.}

\xxparatran
{例えば$S \isInstanceOf P$は$S$が$P$の個体事例（固有名詞を持つもの）であること，つまり$S$が下位の項を1つも持たないことと定義されている（例：$\it{Tweety}\isInstanceOf \it{bird}$）．
しかし無生物の場合，名称を持つ個体的なものがその下位に個体的な亜種を持つと考えることが自然な場合がしばしばある（例えば，特定の楽曲が多くの編曲を持ち，それらがさらに個々の演奏を持ち，さらにその演奏の録音の編集があり，等）．
$\isInstanceOf$が必要だとするWangの説明も理解できる一方，
線引後の両側に静的・決定的な質の違いを定めながら，境界線の引きどころが曖昧な区別を要請することは，Wangが批判した述語論理の述語と存在の区別に通じるものを自ら導入して，項論理の簡潔性を損なっているように思われる．}{
For example, $S \isInstanceOf P$ is defined to mean that $S$ is an individual instance of $P$ (that has a proper noun), that is, $S$ has no subterms (e.g. : $\it{Tweety}\isInstanceOf \it{bird}$).
However, in the case of inanimate objects, it is often natural to think that an individual with a name has an individual subspecies below it (for example, a particular piece of music has many arrangements, and they are even more individual. Has a performance of, and has edited the recording of that performance, etc.).
While I can understand Wang's explanation that $\isInstanceOf$ is necessary, it is not possible to require a vague distinction where the boundary line is drawn, while defining static and definitive quality differences on both sides after the line is drawn, seems to impair the succinctness of term logic by introducing something similar to the distinction between predicate and existence in predicate logic, which Wang criticized.}{
For example, $S \isInstanceOf P$ means $S$ is an individual case (that has a proper noun) of $P$, that is, $S$ has no subterms (for example, $\it{Tweety}\isInstanceOf \it{bird}$). 
However, in the case of nonliving things, it is often natural to think that an individual with a name has individual subspecies under it (for example, a particular piece of music has many arrangements, and they also have individual performances, and further there is a compilation of recordings of the performance, etc.).
Wang's explanation that $\isInstanceOf$ is necessary is understandable, and the request for a fuzzy distinction between where the boundary is drawn, while defining a difference in static and definite quality on both sides after the line is drawn, seems to impair the succinctness of term logic by introducing something similar to the distinction between predicate and existence in predicate logic, which Wang criticized.}
{
For example, $S \isInstanceOf P$ means $S$ is an individual case (that has a proper noun) of $P$, that is, $S$ has no sub-terms (for example, $\it{Tweety}\isInstanceOf \it{bird}$). 
However, in the case of nonliving things, it is often natural to think that an individual with a name has individual subspecies under it (for example, a particular piece of music has many arrangements, and they also have individual performances, and further there is a compilation of recordings of the performance, etc.).
While Wang's explanation that $\isInstanceOf$ is necessary is understandable, the request for a fuzzy distinction that defines a difference in static and definite quality on both sides after the line is drawn, seems to impair the succinctness of term logic by introducing something similar to the distinction between predicate and existence in predicate logic, which Wang criticized.}

\xxparatran
{一方で，経験的な意味とそれに依拠する推論能力を重視する時に，「麺類が好きならばうどんが好き」というような論理包含的関係と，「風が吹くならば木が揺れる」というような因果的関係を同じ含意の繋辞で一括に扱ってよいのかは，自明ではないように思える．
繋辞に関する疑問点は他にもある．
先にも触れた変換規則で$S \to P$を$P \to S$に変換したときの変換後の繋辞も，
もはや継承繋辞ではなく連想繋辞や代表性繋辞とよぶべき異質の繋辞ではないか．
\cite{Wang13}では継承繋辞が伝統論理の$A$型の繋辞のように（暗黙的に）説明される一方で$I,E,O$型については議論されないが，$I,E,O$型の繋辞も必要ではないか．
そもそも人間の日常的な推論においては，「全ての（all）」と「とある（some）」の間には，「ほとんどの」「多くの」「少しの」など連続的で多様な量化の認識があるのではないか．}{
On the other hand, when empirical meaning and reasoning ability based on empirical meaning are emphasized, logical inclusive relations such as ``If you like noodles, you like udon'' and ``If the wind blows, trees sway.'' It seems non-trivial whether causal relations can be treated collectively with the same implicative suffix.
The converted copula when $S \to P$ is converted to $P \to S$ by the conversion rule mentioned earlier should no longer be called an associative copula or a representative copula. Isn't it a foreign copula?
In \cite {Wang13}, inheritance copulas are explained (implicitly) like $A$ type copulas in traditional logic, while $I, E, O$ types are not discussed, but $I, E, O$ type copulas?
In the first place, in human daily reasoning, there is a continuous and diverse quantification such as "most", "many", and "a little" between "all" and "some". Isn't there recognition?}{
Meanwhile, when emphasizing empirical meaning and inference ability based on emphasizing empirical meaning, it is not obvious whether the conditional inclusive relations such as ``If you like noodles, you like udon" and the causal relationship like ``If the wind blows, the tree shakes'' can be treated collectively with the same implicative copula. 
There are other questions about copulas. In the above-mentioned conversion rules, the converted copula of the time that $S \to P$ is converted into $P \to S$ to be called associative copulas, representative copulas, or the heterogeneous copulas, rather than inheritance copulas. In \cite {Wang13}, inheritance copulas are explained (implicitly) like $A$-type copulas in traditional logic, while $I,E,O$-type copulas are not discussed, but I think $I,E,O$-type connectives are also necessary. First of all, in everyday human reasoning, there may be a continuous and diverse quantification recognition between ``all'' and ``some'', such as ``most'', ``a lot'', ``a little'', and so on.
}
{
Meanwhile, when emphasizing inference ability based on empirical meaning, it is not obvious whether the conditional inclusive relations such as ``If one likes pasta, the one likes spaghetti" and the causal relationship like ``If the wind blows, the tree shakes'' can be treated collectively with the same implicative copula. 
There are other questions about copulas. In the above-mentioned conversion rule, the copula of $P \to S$ converted from $S \to P$ may have better be called the associative, representative, or heterogeneous copula, rather than the inheritance copula. In \cite {Wang13}, the inheritance copula is explained (implicitly) like the $A$-type copula in traditional logic, while $I,E,O$-type copulas are not discussed, but $I,E,O$-type copulas would be also necessary. 
In the first place, in human daily reasoning, there might be a continuous and diverse quantification recognition between ``all'' and ``some'', such as ``most'', ``a lot'', ``a little'', and so on.}

\xxparatran
{以上を総合すると，繋辞も連続的な空間の中で家族的・プロトタイプ的なカテゴリ\cite{Wittgenstein53,Taylor12}をなすと考えるのが良いように思える．
先験的に与える繋辞は
最低限に留め， その他のニュアンス（例えば，$\isInstanceOf$）は経験的に学習させる，
つまり繋辞についても\S\ref{sec:latent_repr}で述べたように連続的な表象を与えられる枠組みにし，それを経験的に学習させ，状況に応じた繋辞のニュアンスの区別を推論器に任せる（ある場面では区別するし別の場面では区別しないということがおきる），というアプローチである．
ただし，真理値関数の設計方法が大きな課題となる．
人手によるトップダウンで詳細な設計はもはや困難である.}
{Taken together, it seems better to think of the copula as a family / prototype category \ cite {Wittgenstein53, Taylor12} in a continuous space.
Minimize a priori copulas and let other nuances (eg $ \isInstanceOf $) be learned empirically, that is, copulas as described in \S\ref{sec:latent_repr}. Make a framework that gives continuous representations, let them learn empirically, and let the inferencer distinguish the nuances of copulas according to the situation. ), The approach.
However, the design method of the truth value function is a big issue.
Manual top-down and detailed design is no longer difficult.}
{Overall, it seems better to think that copula is also a family and prototype category in continuous space \cite{Wittgenstein53,Taylor12}. The approach is that the given copulas are kept minimally in a prior way and other nuances (For example, $\isInstanceOf$) are learned empirically, that is, copulas as described in \S\ref{sec:latent_repr} make a framework to give continuous representation and learn it empirically, and the differences of nuances of copulas according to the situations are handed over to the reasoner (the distinction may be made in some situations and not in others). However, a major challenge is how to design the truth function. It is no longer feasible to design the functions in a top-down, detailed manner by hand.
}
{
Taken together, it seems better to think of copula as a family/prototype category \cite{Wittgenstein53,Taylor12} in a continuous space. The approach is that the copulas given in a prior manner are kept minimally and other nuances (for example, $\isInstanceOf$) are learned empirically, that is, copulas as described in \S\ref{sec:latent_repr} take continuous representations, which are learned empirically, and the differences of nuances of copulas according to the situations are handed over to the reasoner (the distinction may be made in some situations and not in others). However, a major challenge is how to design the truth functions with regard to copulas. It is no longer feasible to design the truth functions in a top-down, detailed manner by hand.}

\subsubsection{Separation and coordination of reasoning and language processing} \label{sec:separation_of_logic_and_language}

\xxparatran
{形態素単位で解析した自然言語文の意味を論理言語で表示し，その上で妥当な推論を行えることは，
論理・推論システムの原理実証・汎用性の観点からは望ましいかもしれないが，目的と実態に照らせば必ずしも必要なことではないだろう．
\S\ref{sec:rules}で示したパースの例を見ても，
個々の単語が個別の項となるまで文構造を解体しないと推論ができないわけではない．
必要なことは，状況に応じて適当な粒度の意味・概念を表す項同士の類似性・関連性の認識である．
かつては形式記号の単一化しかその手段がなく，最小粒度での文の解体が必要不可欠だったかもしれないが，今はそうではない．}
{It may be desirable from the point of view of proof of principle and versatility of logic/inference systems to display the meaning of natural language sentences analyzed by morpheme unit in logic language, and to be able to make reasonable inferences on that basis. Considering the actual situation, this is not necessarily necessary.
Even if you look at the example of parsing shown in \S\ref{sec:rules}, it is not impossible to make inferences without deconstructing the sentence structure until each word becomes an individual term.
What is necessary is recognition of similarities and relationships between terms that express meanings and concepts with appropriate granularity according to the situation.
In the past, the only way to do this was to unify formal symbols, and deconstructing sentences at the smallest possible granularity might have been essential, but that is no longer the case.}
{It might be desirable that express the meaning of natural language sentences analyzed by morphemes in logical language and make appropriate inferences on this basis according to the view of principle proof or universality of logic or inference system. 
Considering the actual situation, this is not necessary.
From the perspective example shown in \S\ref{sec:rules}, it is not impossible to make inferences without deconstructing the sentence structure until each word becomes an individual term. 
What is necessary is recognition of similarities and relationships between terms that express meanings and concepts with appropriate granularity according to the situation.
In the past, the only way to do this was to unify formal symbols, and deconstructing sentences at the smallest possible granularity might have been essential, but this is no longer the case.
}
{
It may be desirable from the point of view of proof of principle and versatility of logic/reasoning systems to represent the meaning of natural language sentences analyzed by morpheme units in a logical language, and to be able to make reasonable inferences on that basis. 
However, this is not always necessary.
Pierce's example shown in \S\ref{sec:rules} demonstrates that it is possible to make inferences without decomposing the sentence structure until each word becomes an individual term. 
What is necessary is recognition of similarities and relationships between terms that express meanings and concepts with appropriate granularity according to the situation.
In the past, the only way to do this was to unify formal symbols while decomposing sentences at the smallest granularity, but this is no longer the case.}

\xxparatran
{人間は必ずしも事態／状況を言語的に表示しなくても前述のパースの例が示すような推論は行えるように思えるし，
日常の言語使用はかなりの部分が反射的・自動的で，熟考的な記号推論が理解と応答のためにその都度行われているようにも見えない．
子供の発達を見れば，人間は論理的思考ができるようになってから言葉が話せるようになるのでもないし，流暢に会話できるからといって論理的な思考ができるのでもない．}
{It seems that humans can make inferences such as those shown in Peirce's example above, even if they do not necessarily express the situation verbally. Nor does it appear that symbolic reasoning is performed each time for comprehension and response.
Looking at the development of children, we can see that humans do not become able to speak after being able to think logically, nor are they able to think logically just because they can speak fluently.}
{
It seems that humans can make the kind of inferences that the Perth's example above demonstrates without necessarily having to verbally indicate the situation, and that daily language use is to a large extent reflexive and automatic, with no deliberative symbolic inference being made each time for understanding and response.
From the perspective of the development of children, humans are not able to speak until they are able to think logically, nor are they able to think logically just because they can speak fluently.}
{
It seems that humans can make inferences such as those shown in Peirce's example above, even if they do not necessarily express the situation verbally. Nor does it appear that symbolic reasoning is performed each time for comprehension and response in our daily language use, which seems greatly \textit{relfexive and automatic}.
Looking at the development of children, we can see that humans do not become able to speak after being able to think logically, nor are they able to think logically just because they can speak fluently.}

\xxparatran{その一方で，必要に応じて，必要な粒度で，表現の意味するところを論理的に思慮したり，
複雑な状況を言語化することでより妥当な推論を行うことも，人間はできているように見える．
人間の「推論のシステム」と「言語のシステム」はそれぞれにあり，相互依存していると捉えるのが妥当なアプローチではないだろうか．
それぞれだけでも一定程度有効に機能するが，より効率的・効果的（つまり知的）であるためには，相互の活用・連携が鍵となる．}
{On the other hand, human beings are also able to make more reasonable inferences by logically thinking about the meaning of expressions and verbalizing complicated situations, if necessary, with the necessary particle size. looks like.
A reasonable approach would be to regard the human ``inference system'' and ``language system'' as separate and mutually dependent.
Each function works effectively to a certain extent, but mutual utilization and cooperation are the key to being more efficient and effective (in other words, intellectual).}
{On the other hand, human beings are also able to make more reasonable inferences by logically thinking about the meaning of expressions and verbalizing complicated situations based on the needs and necessary granularity. 
A reasonable approach would be to regard the human's ``inference system'' and ``language system'' as separate and mutually dependent. 
Each function works effectively to a certain extent, but mutual utilization and cooperation are the key to being more efficient and effective (in other words, intellectual).
}
{
On the other hand, human beings are also able to make more reasonable inferences by logically thinking about the meaning of expressions and verbalizing complicated situations based on the needs at necessary granularity. 
A reasonable approach would be to regard the human's ``reasoning system'' and ``language system'' as separate and mutually dependent. 
Each system individually works effectively to a certain extent, but mutual utilization and coordination are the key to being more efficient and effective, i.e., intelligent.}

\xxparatran
{前述の反射的・自動的な言語能力については，近年の深層学習技術によって，かなりの部分が既に十分に人間を近似できるレベルでモデル化されているように思われる．
これを「言語のシステム」とし，それを所与として，残りの「推論のシステム」の実現を考えることが，本研究のとるべきアプローチと考える．}
{Recent deep learning technology seems to have already modeled a large part of the above-mentioned reflexive and automatic language ability at a level that can sufficiently approximate humans.
The approach we should take in this research is to consider this as a ``language system'' and to consider the realization of the remaining ``system of reasoning'' with it as a given.}
{Recently, deep learning technology seems to have already modeled a large part of the above-mentioned reflexive and automatic language ability at a level that can sufficiently approximate humans. 
We consider this as a ``language system'' and as a precondition.  
The implementation of the remaining ``inference system'' should be considered as the approach to be taken in this research.
}
{
Recent deep learning technology seems to have already modeled a large part of the above-mentioned reflexive and automatic language ability at a level that can sufficiently approximate humans.
The approach we should take in this research is to consider this as a ``language system'' and to consider the realization of the remaining ``reasoning system''.}

\xxparatran
{NALの積項と関係項で文法構造の細かな表現を行うことは不可能ではないかもしれないが，複雑な仕組みの導入を必要とするだろう．
文法的な要素の処理は「言語のシステム」に任せることができれば，「推論のシステム」の構成は簡潔に保つことができる．}
{It may not be impossible to express grammatical structures finely with NAL product terms and relatative terms, but it would require the introduction of complex mechanisms.
If the processing of grammatical elements can be left to the "system of language", the construction of the "system of reasoning" can be kept simple.}
{It may not be impossible to express grammatical structures finely with NAL product terms and linkage terms, but it would require the introduction of complex mechanisms.
If the processing of grammatical elements can be left to the ``language system'', the structure of the ``inference system'' can be kept concise.
}
{
It may not be impossible to express grammatical structures finely with NAL product terms and relational terms, but it would require the introduction of complex mechanisms. If the processing of grammatical elements can be left to the ``language system'', the construction of the ``reasoning system'' can be kept simple.}

\xxparatran
{同様の考え方の有効性は，既に\citet{Nye_Dual-System_Neurips21}によって，極限られた問題設定の中でではあるが実証されている\footnote{\label{fn:DualSystem}
\citet{Nye_Dual-System_Neurips21}は，直感的な言語的応答を返すシステム（いわゆるSystem 1~\cite{Kahnemann11}）と，その結果に対して論理的なフィルタをかけるシステム（System 2）を用いることで，
言語的に自然で，かつ整合した物語生成ができることを示した．System 1として，GPT-3~\cite{GPT3}を用いている．

GPT-3は，人間のSystem 1とSystem 2の働きをテストする次の問題
``A ball and a bat cost \$1.10. The bat costs one dollar more than the ball. How much does the ball cost?''を与えられたとき，時間制限を課された多くの人間と同じ「間違った回答」（10 cents）を返す（正しい答えは5 cents）\cite{Nye_Dual-System_Neurips21}．
Taylar~ら認知言語学者が考える言語知識とは，「言葉の正しい使用方法に関する知識」であり，「語の意味を知っている」とはその言語におけるその語の「用法を知っている」ということに等しい\cite{Taylor12}．
この意味で，このテストにおいて，GPT-3は完全に人と同等に「語の意味」を知っていることを実証している．
本論文も，言語に対して，認知言語学の立場を取る．
本文で述べた「反射的・自動的な言語能力」とは，このような言語能力を指している．

一方で，GPT-3はこのテストに失敗している．
それはこのテストが「言語のシステム」，言語の用法だけでは解けない問題だからである．
本研究の焦点もその部分にある．
}．
本研究の構想はこれを，この後で見るように，仮説推論，論証，比喩・アナロジー，発見的問題解決など，人間の多様な記号的思考に広げ，非公理的項論理という形式的枠組みの上で統一的に扱おうとするものである．}
{The effectiveness of a similar idea has already been demonstrated by \citet{Nye_Dual-System_Neurips21}, albeit in an extremely limited set of problems
\footnote{\label{fn:DualSystem}
\citet{Nye_Dual-System_Neurips21} uses a system that returns intuitive verbal responses (so-called System 1~\cite{Kahnemann11}) and a system that logically filters the results (System 2). showed that it is possible to generate linguistically natural and consistent stories. As System 1, GPT-3~\cite{GPT3} is used.
GPT-3 tests the functioning of human System 1 and System 2 with the following problem ``A ball and a bat cost \$1.10. The bat costs one dollar more than the ball. gives the same ``wrong answer'' (10 cents) as many humans with time limits (correct answer is 5 cents) \cite{Nye_Dual-System_Neurips21}.
Cognitive linguists such as Taylar consider linguistic knowledge to be ``knowledge about the correct usage of words,'' and ``knowing the meaning of words'' means ``knowing the usage'' of the words in the language. Equivalent to \cite{Taylor12}.
In this sense, this test demonstrates that GPT-3 knows the ``meaning of words'' exactly as well as humans.
This paper also takes the standpoint of cognitive linguistics with respect to language.
The ``reflexive/automatic linguistic ability'' mentioned in the text refers to this kind of linguistic ability.
On the other hand, GPT-3 fails this test.
This is because this test is a "language system," a problem that cannot be solved by language usage alone.
The focus of this research is also on that part.
}. As we will see later, the concept of this research is to expand this to various symbolic thinking of human beings such as hypothetical reasoning, argumentation, metaphor/analogy, and heuristic problem solving. This is what we are going to treat in a unified way.
}
{The effectiveness of a similar idea has already been demonstrated by \citet{Nye_Dual-System_Neurips21}, but in a very limited problem setting
\footnote{\label{fn:DualSystem} 
\citet{Nye_Dual-System_Neurips21} uses a system that returns intuitive verbal responses (so-called System 1~\cite{Kahnemann11}) and a system that logically filters the results (System 2) to show that it is possible to generate linguistically natural and consistent stories. System 1 uses GPT-3~\cite{GPT3}.
When GPT-3 is given the following question that is used to test the functioning of human System 1 and System 2, ``A ball and a bat cost \$1.10. The bat costs one dollar more than the ball. How much does the ball cost?'', it returns the same ``wrong answer'' (10 cents) as many humans do in a time limit (the correct answer is 5 cents) \cite{Nye_Dual-System_Neurips21}. 
Cognitive linguists such as Taylar \cite{Taylor12} consider linguistic knowledge to be ``knowledge about the correct use of language'',  and ``knowing the meaning of a word'' is equal to ``knowing the usage of the word'' in the language.
In this sense, this test confirmed that GPT-3 could understand ``the meaning of words'' exactly as well as humans. 
This paper also takes the standpoint of cognitive linguistics with respect to language.
The ``reflective and automatic language ability'' mentioned in this paper refers to such kind of linguistic ability.
Meanwhile, GPT-3 has failed this test. It is because this test is a problem that cannot be solved by a ``language system'', that is, language usage alone. The focus of this research is also on this part.
}. 
As we will see later, the concept of this research is to expand this to various symbolic thinking of human beings such as hypothesis inference, argumentation, metaphor/analogy, heuristic problem solving, and so forth, and try to handle them uniformly on the formal framework of non-axiomatic term logic.
}
{
The effectiveness of a similar idea has already been demonstrated by \citet{Nye_Dual-System_Neurips21}, but in a limited problem setting. \footnote{\label{fn:DualSystem} 
\citet{Nye_Dual-System_Neurips21} uses a system that returns intuitive verbal responses (so-called System 1~\cite{Kahnemann11}) and a system that logically filters the results (System 2) to show that it is possible to generate linguistically natural and consistent stories. System 1 uses GPT-3~\cite{GPT3}.
When GPT-3 is given the following question that is used to test the functioning of human System 1 and System 2, ``A ball and a bat cost \$1.10. The bat costs one dollar more than the ball. How much does the ball cost?'', it returns the same ``wrong answer'' (10 cents) as many humans do in a time limit (the correct answer is 5 cents) \cite{Nye_Dual-System_Neurips21}. 
Cognitive linguists such as Taylar consider linguistic knowledge to be ``knowledge about the appropriate use of language'', and ``knowing the meaning of a word'' is equal to ``knowing the usage of the word'' in the particular language \cite{Taylor12}. 
In this sense, this test confirmed that GPT-3 could understand ``the meaning of words'' exactly as well as humans do. The present paper also takes the standpoint of cognitive linguistics with respect to language. The ``reflective and automatic language ability'' mentioned in this paper refers to such kind of linguistic ability. Meanwhile, GPT-3 has failed this test. It is because this test is a problem that cannot be solved by language usage alone. The focus of our research is also on that part.
}
As we will see later, the concept of this research is to expand this to various symbolic thinking of human beings such as hypothesis inference, argumentation, metaphor/analogy, heuristic problem solving, and so forth, and try to handle them uniformly on the formal framework of Non-Axiomatic Term Logic.
}

\section{Term Representation Language}\label{sec:trl}


\xxparatran{前章で示したNALの課題を受け，
本章と次章で本論文が提案する非公理的項論理を提示する．
本章ではまず非公理的項論理が用いる項表示言語を示す．}
{In response to the NAL issues presented in the previous section, we present the non-axiomatic term logic proposed in this paper in this and the next section.
In this chapter, we first show the term representation language used by non-axiomical term logic.
}
{In response to the NAL issues shown in the previous section, this section, and the next section we present the non-axiomatic term logic proposed in this paper.
In this chapter, we first present the term representation language used by the non-axiomatic term logic. 
}
{
In response to the issues in NAL presented in the previous section, we present Non-Axiomatic Term Logic proposed in this paper in this and the next sections.
In this section, we first define Term Representation Language used by Non-Axiomatic Term Logic.
}

\xxparatran{項表示言語は，人間の認知的内部表象の表示モデルであり，記号を用いて認識と知識を表現する．
次章で示す非公理的項論理は，項表示言語によって明示的に構造化された知識表現を用いて，推論の過程を表示する．
ここでは項表示言語の構文論を示す．}{
A term display language is a display model of human cognitive internal representation, and uses symbols to express cognition and knowledge.
Non-axiomatic term logic, which will be presented in the next chapter, expresses the process of reasoning using knowledge representations explicitly structured by a term representation language.
Here we present the syntax of the term display language.}
{The term representation language is a display model of human cognitive internal representations that use symbols to represent recognition and knowledge.
The non-axiomatic term logic, shown in the next chapter, expresses the process of reasoning using knowledge representations explicitly structured by the term representation language.
Here we present the syntax of the term representation language.
}
{
Term Representation Language (hereafter, TRL) is a description model of human cognitive internal representations that use symbols to represent recognition and knowledge. Non-Axiomatic Term Logic, shown in the next section, expresses the process of reasoning using knowledge representations explicitly structured by TRL. Here we present the syntax of TRL.
}

\xxparatran{大雑把に言えば，項表示言語はNAL~\cite{Wang13}の表示言語であるNarsesの陳述の表示を基本に，述語論理と同様の任意の数の項をとる命題表示を導入したものである．}
{Roughly speaking, the term display language is based on Narses' statement display, which is the display language of NAL~\cite{Wang13}, and introduces a propositional display with an arbitrary number of terms similar to predicate logic.}
{Roughly speaking, the term representation language is based on the representation of statements by Narses, which is also the representation language of NAL\cite{Wang13} and introduces a propositional display that takes an arbitrary number of terms, similar to predicate logic.
}
{
Roughly speaking, TRL is based on the notation of statements by Narses, which is the representation language of NAL~\cite{Wang13}, and it introduces a proposition notation that takes an arbitrary number of terms, similar to predicate logic.}

\subsection{Term}
\xxparatran{
項（term）は，今まさに知覚されている経験の認識（例えば，眼前の1つのりんご，あるいはそのりんごが木から落下する様子），
それがエピソード化された記憶，エピソード記憶の集積から抽出・抽象化された意味記憶（いわゆる概念，例えば一般的な意味としての「りんご」や「落下」）など，
人間の心理的処理によって1つのまとまりをもって意識の中に立ち現れるものを表す．
任意の項を表すのに，$T_1, T_2, T_3, \ldots$の記号を用いる．}{
A term is a perception of an experience that is being perceived right now (e.g., an apple in front of one's eyes or the appearance of an apple falling from a tree), its episodic memory, and extraction and extraction from episodic memory accumulation. Abstracted semantic memories (so-called concepts, such as ``apple'' and ``falling'' in general terms) represent things that appear in the consciousness as a unity through human psychological processing.
We use the symbols $T_1, T_2, T_3, \ldots$ to denote arbitrary terms.}{
A term is a perception of the experience that is being perceived right now (for example, one apple in front of you, or one apple falling from the tree), an episodic memory of that experience, a semantic memory extracted and abstracted from an accumulation of episodic memories (they can be called concepts, such as ``apple'' and ``falling'' in the general sense), which emerges in consciousness as a unity through human psychological processing. The symbols $T_1, T_2, T_3, \ldots$ are used to denote arbitrary terms.
}
{
A term represents an immediate perception of an experience (for example, one apple in front of you, or one apple falling from a tree), an episodic memory of that experience, a semantic memory extracted and abstracted from an accumulation of episodic memories (they can be called concepts, such as ``apple'' and ``falling'' in the general sense), which emerges in consciousness as a unity through human psychological processing. The symbols $T_1, T_2, T_3, \ldots$ are used to denote arbitrary terms.}

\xxparatran{ある英単語と結びついた特定の概念を項の指示対象として想定する場合，
可読性を高めるために，$\term{sheep}$, $\term{human}$などのように斜体の英単語で表す．
それらの概念（タイプ）が具象化された対象をトークンとして区別する場合は，$\term{human}_1$, $\term{human}_2$のようにする．}
{If we assume that the referent of a term is a specific concept associated with a word, italicize the word, such as $\term{sheep}$, $\term{human}$, etc., to improve readability.
When distinguishing objects in which those concepts (types) are instantiated as tokens, use $\term{human}_1$, $\term{human}_2$.}
{When assuming a specific concept associated with a certain English word as the referent of a term, italicized English words such as $\term {sheep}$, $\term {human}$ are used to improve readability.
To distinguish the objects whose concepts (types) are instantiated as tokens, use terms like $\term{human}_1$, $\term{human}_2$.
}
{
When assuming a specific concept associated with a certain English word as the referent of a term, italicized English words such as $\term {sheep}$, $\term {human}$ are used to improve readability.
To distinguish the objects whose concepts (types) are instantiated as tokens, use terms like $\term{human}_1$, $\term{human}_2$.}

\subsection{Basic term and composed term}
\xxparatran{
認識の最小単位として扱われる項を基本項（basic term）とよぶ．
例えば，先に挙げた$\term{sheep}$, $\term{human}$, $\term{human_1}$, $\term{human_2}$は基本項である．
ところで，ある概念が，自然言語によって1語で表現されうるかどうかは，概念の単独性と関係しない\cite{Hofstadter2013}．
複数の語で表現される概念・対象であっても，ハイフンでつないで$\term{polar\mh bear}, \term{getting\mh wet}$のようにし，1つの基本項として表示する．}{
The term treated as the minimum unit of recognition is called the basic term.
For example, $\term {sheep}$, $\term {human}$, $\term {human_1}$, $\term {human_2}$ mentioned above are basic terms.
By the way, whether or not a concept can be expressed in one word by natural language has nothing to do with the singularity of the concept\cite {Hofstadter2013}.
Even if the concept / object is expressed in multiple words, connect it with a hyphen to make it $\term{polar\mh bear}, \term{getting\mh wet}$, and display it as one basic term.}{
The term treated as the minimum unit of recognition is called a basic term.
For example, $\term {sheep}$, $\term {human}$, $\term {human_1}$, $\term {human_2}$ mentioned above are basic terms.
Besides, whether or not a concept can be expressed in one word by natural language has nothing to do with the singularity of the concept\cite{Hofstadter2013}.
Even if a concept or an object is expressed by multiple words, they can be hyphenated like $\term{polar\mh bear}, \term{getting\mh wet}$ and presented as a single basic term.
}
{
A term treated as the minimum unit of recognition is called a basic term. For example, $\term {sheep}$, $\term {human}$, $\term {human_1}$, $\term {human_2}$ mentioned above are basic terms. Besides, whether or not a concept can be expressed in one word by natural language has nothing to do with the singularity of the concept \cite{Hofstadter2013}. Even if a concept or an object is expressed by multiple words, they can be hyphenated like $\term{polar\mh bear}, \term{getting\mh wet}$ and used as a single basic term.}

\xxparatran{基本項であることを一般的に記述するときには，$B_1, B_2, \ldots$ の記号を用いる．
次に説明する複合項・陳述項・連関項を総称して構成項（composed term）とよぶ．
基本項は複合項・陳述項を構成する典型的な単位である．}{
When describing that it is a basic term, the symbols $ B_1, B_2, \ ldots $ are used.
The compound terms, statement terms, and relational terms described below are collectively called composed terms.
The basic term is a typical unit that constitutes a compound term /statement term.}{
The symbols $B_1, B_2, \ldots$ are generally used to describe what is a basic term.
The compound terms, statement terms, and relational terms described below are collectively called composed terms.
A basic term is a typical unit that constitutes compound terms and statement terms.
}{
Symbols $B_1, B_2, \ldots$ are generally used to denote basic terms. The compound terms, statement terms, and linkage terms to be described below are collectively called composed terms. A basic term is a typical unit that constitutes compound terms and statement terms.}

\subsection{Compound term} \label{sec:compound-term}
\xxparatran{
複数の項の間の関係，または単独の項に関する2次的な認識を表現する構成項を複合項（compound term）とよぶ．
複合項を構成する要素（要素項）は基本項に限らず，項全般を取りうる．
これらの認識・関係の概念を表す項を$R$で表し，複合項を$C$とすると，
それを構成する関係項と要素項で表すときは，$C: (R, T_1, T_2, \ldots, T_n)$ のようにする．
複合項の要素数$n$は$R$に依る．
NALにおいては3項関係を積項と関係項を用いた陳述$U \times V \to R$として表示したが（\S\ref{sec:nal_syntax}参照），
本論の項表示言語では複合項$C:(R, U, V)$として表示される．}
{
A composed term that expresses a relationship between multiple terms or a secondary perception of a single term is called a compound term.
The elements (element terms) that make up a compound term are not limited to the basic terms, but can be all terms.
If the terms that represent these concepts of recognition and relationships are represented by $ R $ and the compound terms are represented by $ C $, then when the terms and element terms that compose them are represented, $C: (R, T_1, T_2, \ldots, T_n)$.
The number of elements in the compound term $n$ depends on $R$.
In NAL, the three-term relation is expressed as a statement $U \times V \to R$ using the product and relational terms, but in the term display language of this paper, it is expressed as a compound term $C:(R, U, V)$. Is displayed.
}
{
A composed term that expresses a relationship between multiple terms or a secondary perception of a single term is called a compound term.
The elements (element terms) that make up a compound term are not limited to the basic terms, but can be all terms.
Let $R$ denote the term representing these concepts of recognition and relationship, and let $C$ denote the compound term, then when the terms and element terms that compose them are represented, they are like $C: (R, T_1, T_2, \ldots, T_n)$.
The number of elements in the compound term $n$ depends on $R$.
In NAL, the three-term relation is expressed as a statement $U \times V \to R$ using the product and relational terms (see \S\ref{sec:nal_syntax}), but in the term representation language of this paper, it is expressed as a compound term $C:(R, U, V)$.
}
{
A composed term that expresses a relationship between multiple terms or a secondary recognition of a single term is called a compound term.
The elements (element terms) that make up a compound term are not limited to basic terms, but can be all kinds of terms.
Let $R$ denote the term representing these concepts of recognition and relationship, and let $C$ denote the compound term $C: (R, T_1, T_2, \ldots, T_n)$. Here $T_1, T_2, \ldots$ are the element terms of $C$.
The number of elements in the compound term, $n$, depends on $R$.
In NAL, a three-term relation is expressed as a statement $U \times V \to R$ using the product and relational terms (see \S\ref{sec:nal_syntax}), but in TRL, it is expressed as a compound term $C:(R, U, V)$.}

\xxparatran{自然言語において一般動詞を用いて表現される「事態」は，
$(\term{rotate}, \term{Earth})$，$(\term{eat}, \term{sheep}, \term{grass})$
のように，いわゆる述語項構造と同じ要領で，複合項として表示できる．
項の連言・選言も，$(\term{and}, T_1, T_2, \ldots)$のように表示できる．
夫婦のような特定の関係性を持つ集合体も，
$(\term{couple}, \term{husband_1}, \term{wife_2})$のように表示できる．}{
``Situation'' expressed using general verbs in natural language is $(\term{rotate}, \term{Earth})$, $(\term{eat}, \term{sheep},\term{grass })$ can be displayed as a compound term in the same manner as the so-called predicate-argument structure.
Conjunctions and disjunctions of terms can also be displayed as $(\term{and}, T_1, T_2, \ldots)$.
Aggregates with specific relationships such as couples can also be displayed as $(\term{couple}, \term{husband_1}, \term{wife_2})$.
}{
In natural language, a ``situation'' expressed with a general verb like $(\term{rotate}, \term{Earth})$, $(\term{eat}, \term{sheep}, \term{grass})$, and so on, can be displayed as compound terms in the same way as predicate term structures.
The conjunctions and disjunctions can be displayed as $(\term{and}, T_1, T_2, \ldots)$. And aggregates with specific relationships, such as couples, can also be displayed as$(\term{couple}, \term{husband_1}, \term{wife_2})$.
}
{
A ``situation'' expressed with a general verb in a natural language
can be expressed as compound terms like $(\term{rotate}, \term{Earth})$, $(\term{eat}, \term{sheep}, \term{grass})$ in the same way as predicate argument structures. The conjunctions and disjunctions can be expressed as $(\term{and}, T_1, T_2, \ldots)$. Aggregates with specific relationships, such as couples, can also be expressed as $(\term{couple}, \term{husband_1}, \term{wife_2})$.}

\subsection{Statement term}
\xxparatran{
陳述項（statement term）$S$は，2つの任意の項を繋辞でつないだ構成項である．
すなわち $S: (T_1 \to T_2)$．$T_1$を$S$の左項，$T_2$を右項とよぶ，

陳述項は，NALにおける陳述におよそ対応するものであり，
項に対する分類（categorization）の認識を表す．
もっとも基本的な繋辞は，「ソクラテスは男である」のようなis-a関係を表すものであるが，
そのバリエーションとして，is-aの否定（例えば「ソクラテスは女ではない」）を表す繋辞の他に，
同値（$T_1 = T_2$）の繋辞，伝統論理のA/I/E/Oの4文型を表す繋辞などが派生的に複数存在すると考える．
繋辞が何種類存在するのかは表示言語としては規定しない．}{
The statement term $ S $ is a component of two arbitrary terms connected by a copula. That is, $S: (T_1 \to T_2)$．$T_1$ is called the left term of $S$, $T_2$ is called the right term,
The statement term corresponds roughly to the statement in NAL and represents the recognition of the classification of the term.
The most basic suffix expresses an is-a relationship, such as ``Socrates is a man'', but variations express the negation of is-a (e.g. ``Socrates is not a woman''). In addition to suffixes, we think that there are multiple suffixes that have the same value ($T_1 = T_2$) and suffixes that represent the four sentence patterns of traditional logic A/I/E/O.
The display language does not specify how many types of suffixes exist.
}{
The statement term $S$ is a composed term of two arbitrary terms connected by a copula. That is, $S: (T_1 \to T_2)$, where $T_1$ is called the left term of $S$ and $T_2$ the right term,
The statement term corresponds roughly to the statement in NAL and represents the recognition of the classification of the terms.
The most basic copula expresses an is-a relationship, such as ``Socrates is a man''. However, there are several variations of the is-a relation, other than the negation of is-a (for example, ``Socrates is not a woman''), equivalence ($T_1 = T_2$), the four sentence patterns of A/I/E/O in traditional logic, and so on are all the derivatives from the is-a relationship.
The representation language does not specify how many types of copulas exist.
}
{
A statement term $S$ is a composed term of two arbitrary terms connected by a copula. That is, $S: (T_1 \to T_2)$, where $T_1$ is called the left term of $S$ and $T_2$ the right term. A statement term corresponds roughly to a statement in NAL and represents the  categorization of a term. The most basic copula expresses an is-a relationship, such as ``Socrates is a man''. However, there are several variations of the is-a relation, that is, the negation of is-a (for example, ``Socrates is not a woman''), equivalence ($T_1 = T_2$), the four sentence patterns of A/I/E/O in traditional logic, and so on. All these types are the derivatives from the is-a relationship. TRL does not specify how many types of copulas exist.}

\subsection{Linkage term}
\xxparatran{
連関項（linkage term）$L$は，NALにおいて高階陳述（higher-order statement）あるいは含意陳述（implication statement）とよぶものに対応する．
陳述項同様，2項を要素項としてとる構成項である．すなわち $L: (T_1 \To T_2)$．

連関項の典型的な要素項は，事態を表す複合項や陳述項である．
連関項に用いられる繋辞は規則や因果関係などを表し，
種々の繋辞が存在し得るが，
いくつ存在するのかは表示言語としては規定しない．}{
The linkage term $ L $ corresponds to what is called a higher-order statement or implication statement in NAL.
Like the statement term, it is a constituent term that takes two terms as element terms. That is, $ L: (T_1 \ To T_2) $.

Typical element terms of relational terms are compound terms and statement terms that represent situations.
The copulas used in the relational terms represent rules, causal relationships, etc., and various copulas can exist, but the number of copulas does not specify as a display language.
}{
The linkage term $L$ corresponds to what is called a higher-order statement or implication statement in NAL.
Like the statement term, it is a composed term that takes two terms as element terms. That is, $L: (T_1 \To T_2)$.

Typical element terms of linkage terms are compound terms and statement terms that represent situations.
The copulas used in the linkage terms represent rules, causal relationships, etc., and can have many different kinds, but the representation language does not specify how many kinds there are.
}
{
A linkage term $L$ corresponds to what is called a higher-order statement or implication statement in NAL.
Like a statement term, it is a composed term that takes two terms as element terms. That is, $L: (T_1 \To T_2)$.
Typical element terms of a linkage term are statement terms and compound terms that represent situations.
The copulas used in the linkage terms represent rules, causal relationships, etc., and can have many different kinds, but TRL does not specify how many kinds there are.
}

\subsection{Variable term}
\xxparatran{
具体的な個別の項に縛られない知識を表現するために，変数項を導入する．
変数項は小文字英字$x,y,\ldots$で表す．
例えば，「$x$が人であるならば，$x$は物語が好き」という知識は，
連関項$L_1$として
\[\ L_1: \ \ S_1: (x \to \term{human}) \To C_1: (\term{favor}, x, \term{narrative})\]
と表示できる．
このとき，$S_1$中の$x$と$C_1$中の$x$は同じ項として束縛される．}{
We introduce variable terms to express knowledge that is not tied to specific individual terms.
Variable terms are represented by lowercase letters $x,y,\ldots$.
For example, the knowledge ``If $x$ is a person, $x$ likes stories'' is represented by \[\ L_1: \ \ S_1: (x \to \term{human}) \To C_1: (\term{favor}, x, \term{narrative})\]. In this case, $x$ in $S_1$ and $x$ in $C_1$ are bound as the same term.
}{
In order to express knowledge that is not tied to specific individual terms, variable terms are introduced.
Variable terms are represented by lowercase letters $x,y,\ldots$.
For example, the knowledge that ``if $x$ is a person, then $x$ likes stories'' can be displayed as the linkage term $L_1$ as
\[\ L_1: \ \ S_1: (x \to \term{human}) \To C_1: (\term{favor}, x, \term{narrative})\]
At this time, $x$ in $S_1$ and $x$ in $C_1$ are bound as the same term.
}
{
In order to express knowledge that is not tied to specific individual terms, variable terms are introduced. Variable terms are represented by lowercase letters $x,y,\ldots$ For example, the knowledge that ``if $x$ is human, then $x$ likes narratives'' can be expressed as linkage term $L_1$ as
\[\ L_1: \ \ S_1: (x \to \term{human}) \To C_1: (\term{likes}, x, \term{narratives}).\]
Here, $x$ in $S_1$ and $x$ in $C_1$ are bound as the same term.
}

\section{Non-Axiomatic Term Logic}\label{sec:natl}

\xxparatran
{\S\ref{sec:trl}で提示した項表示言語を土台として，非公理的項論理（Non-Axiomatic Term Logic; NATL）を提示する．
NATLはNAL同様，モデル理論的意味論（model-theoretic semantics）ではなく，証明理論的意味論（proof-theoretic semantics）を指向している．}
{Non-Axiomatic Term Logic (NATL) is presented based on the term display language presented in \S\ref{sec:trl}.
Like NAL, NATL is oriented towards proof-theoretic semantics rather than model-theoretic semantics.}
{Non-Axiomatic Term Logic (NATL) is presented based on the term representation language presented in \S\ref{sec:trl}. Like NAL, NATL is oriented towards proof-theoretic semantics rather than model-theoretic semantics.
}
{
In this section, Non-Axiomatic Term Logic (NATL) is presented based on TRL introduced in \S\ref{sec:trl}. 
Like NAL, NATL is oriented towards proof-theoretic semantics rather than model-theoretic semantics.}

\subsection{Classes of terms and distinction of thing/logical terms}

\xxparatran
{陳述項のクラスを$\mathcal{S}$，連関項のクラスを$\mathcal{L}$で表し，
複合項と基本項をまとめたクラスを$\mathcal{C}$で表す\footnote{
基本項は，要素項数が1で，関係項が何も意味的作用を持たない特殊な複合項とみなすことができる．}．}
{$\mathcal{S}$ represents the class of statement terms, $\mathcal{L}$ represents the class of associated terms, and $\mathcal{C}$ represents the class of compound terms and elementary terms\footnote{A basic term can be regarded as a special compound term whose number of element terms is 1 and whose relational term has no semantic effect.}.}
{$\mathcal{S}$ represents the class of statement terms, $\mathcal{L}$ represents the class of linkage terms, and $\mathcal{C}$ represents the class of compound terms and basic terms\footnote{A basic term can be regarded as a special composed term with one element term and the linkage term having no semantic effect.}.
}
{
$\mathcal{S}$ denotes the class of statement terms, $\mathcal{L}$ denotes the class of linkage terms, and $\mathcal{C}$ denotes the class of compound terms and basic terms.\footnote{A basic term can be regarded as a special composed term with one element term and one empty relational term having no semantic effect.}}

\xxparatran
{クラス$\mathcal{C}$の項を事物項（thing term），
クラス$\mathcal{S}$と$\mathcal{L}$の項を合わせて論理項（logic term）とよぶ．
\ref{tab:term_system}に，\S\ref{sec:trl}と本節で導入した項の区別と対応関係を示す．}{
A term of class $\mathcal{C}$ is called a thing term, and a term of classes $\mathcal{S}$ and $\mathcal{L}$ together is called a logic term.
\ref{tab:term_system} shows the distinction and correspondence between \S\ref{sec:trl} and the terms introduced in this section.}
{
The term of class $ \mathcal {C} $ is called thing term, and a term of classes $\mathcal{S}$ and $\mathcal{L}$ together is called a logic term. \ref{tab:term_system} shows the distinction and correspondence between \S\ref{sec:trl} and the terms introduced in this section.}
{
A term of class $ \mathcal {C} $ is called a thing term, and a term of classes $\mathcal{S}$ and $\mathcal{L}$ together is called a logic term. Table \ref{tab:term_system} shows the distinction of terms and correspondences to the term classes introduced in this section.}

\xxparatran
{\begin{table}[t]
    \centering
\begin{tabular}{ll|l||ll}
\hline
\multicolumn{3}{c||}{項表示言語}                                           & \multicolumn{2}{c}{非公理的項論理}                            \\ \hline\hline
\multicolumn{1}{l|}{\multirow{5}{*}{項}} & \multicolumn{2}{l||}{基本項}   & \multirow{2}{*}{$\mathcal{C}$} & \multirow{2}{*}{事物項} \\ \cline{2-3}
\multicolumn{1}{l|}{}                   & \multirow{3}{*}{構成項} & 複合項 &                                &                      \\ \cdashline{3-3} \cline{4-5} 
\multicolumn{1}{l|}{}                   &                      & 陳述項 & $\mathcal{S}$                  & \multirow{2}{*}{論理項} \\
\multicolumn{1}{l|}{}                   &                      & 連関項 & $\mathcal{L}$                  &                      \\ \cline{2-5} 
\multicolumn{1}{l|}{}                   & \multicolumn{2}{l||}{変数項}   &                                &                      \\ \hline
\end{tabular}    \caption{項の種別と対応関係}
    \label{tab:term_system}
\end{table}}
{\begin{table}[tb]
    \centering
\begin{tabular}{ll|l||ll}
\hline
\multicolumn{3}{c||}{term display language}                                           & \multicolumn{2}{c}{non-axiomic term logic}                            \\ \hline\hline
\multicolumn{1}{l|}{\multirow{5}{*}{item}} & \multicolumn{2}{l||}{basic item}   & \multirow{2}{*}{$\mathcal{C}$} & \multirow{2}{*}{things} \\ \cline{2-3}
\multicolumn{1}{l|}{}                   & \multirow{3}{*}{composed term} & Compound Term &                                &                      \\ \cdashline{3-3} \cline{4-5} 
\multicolumn{1}{l|}{}                   &                      & Statement & $\mathcal{S}$                  & \multirow{2}{*}{Logic} \\
\multicolumn{1}{l|}{}                   &                      & Relations & $\mathcal{L}$                  &                      \\ \cline{2-5} 
\multicolumn{1}{l|}{}                   & \multicolumn{2}{l||}{variable term}   &                                &                      \\ \hline
\end{tabular}    \caption{Item types and correspondence}
    \label{tab:term_system}
\end{table}}
{\begin{table}[tb]
    \centering
\begin{tabular}{ll|l||ll}
\hline
\multicolumn{3}{c||}{Term Representation Language}                                           & \multicolumn{2}{c}{Non-Axiomatic Term Logic}                            \\ \hline\hline
\multicolumn{1}{l|}{\multirow{5}{*}{Term}} & \multicolumn{2}{l||}{Basic Term}   & \multirow{2}{*}{$\mathcal{C}$} & \multirow{2}{*}{Thing Term} \\ \cline{2-3}
\multicolumn{1}{l|}{}                   & \multirow{3}{*}{Composed term} & Compound Term &                                &                      \\ \cdashline{3-3} \cline{4-5} 
\multicolumn{1}{l|}{}                   &                      & Statement Term & $\mathcal{S}$                  & \multirow{2}{*}{Logic Term} \\
\multicolumn{1}{l|}{}                   &                      & Linkage Term & $\mathcal{L}$                  &                      \\ \cline{2-5} 
\multicolumn{1}{l|}{}                   & \multicolumn{2}{l||}{Variable Term}   &                                &                      \\ \hline
\end{tabular}    \caption{Term Types and Corresponding Relationship}
    \label{tab:term_system}
\end{table}
}
{\begin{table}[tb]
    \centering
\begin{tabular}{ll|l||ll}
\hline
\multicolumn{3}{c||}{Term Representation Language}                                           & \multicolumn{2}{c}{Non-Axiomatic Term Logic}                            \\ \hline\hline
\multicolumn{1}{l|}{\multirow{5}{*}{Term}} & \multicolumn{2}{l||}{Basic term}   & \multirow{2}{*}{$\mathcal{C}$} & \multirow{2}{*}{Thing term} \\ \cline{2-3}
\multicolumn{1}{l|}{}                   & \multirow{3}{*}{Composed term} & Compound term &                                &                      \\ \cdashline{3-3} \cline{4-5} 
\multicolumn{1}{l|}{}                   &                      & Statement term & $\mathcal{S}$                  & \multirow{2}{*}{Logic term} \\
\multicolumn{1}{l|}{}                   &                      & Linkage term & $\mathcal{L}$                  &                      \\ \cline{2-5} 
\multicolumn{1}{l|}{}                   & \multicolumn{2}{l||}{Variable term}   &                                &                      \\ \hline
\end{tabular}    \caption{Term Types and Correspondences between TRL and NATL}
    \label{tab:term_system}
\end{table}
}

\subsection{Reasoning and logic in NATL}

\xxparatran
{NATLに基づく推論は，
1つの論理項と，もう1つの項（事物項または論理項）との間で，
代入操作を行うことにより前提知識から結論を導く．
代入操作が行えるのは，2つの項が同一である場合を含め，2つの項の単一化が可能なときである．
（ここでは文字列の厳密な一致によるハードな単一化ではなく，
\cite{Arabshahi21}が行っているような埋め込みベクトルに基づくソフトな単一化を想定している．）}
{Reasoning based on NATL draws conclusions from prior knowledge by performing substitution operations between one logical term and another term (thing term or logical term).
Substitution operations are allowed when two terms can be unified, including when the two terms are identical.
(Here we assume soft unification based on embedding vectors, as \cite{Arabshahi21} does, rather than hard unification by exact string matching.)}
{Reasoning based on NATL draws conclusions from prior knowledge by performing substitution operations between one logical term and another term (thing term or logical term). 
Substitution operations are allowed when two terms can be unified, including when the two terms are identical. (Here we assume soft unification based on embedding vectors, as \cite{Arabshahi21} does, rather than hard unification by exact string matching)
}
{
Reasoning based on NATL draws conclusions from prior knowledge by performing substitution operations between one logical term and another term (thing term or logical term). 
Substitution operations are allowed when two terms can be unified, including when the two terms are identical. Here, we assume soft unification based on embedding vectors, as \cite{Arabshahi21} does, rather than hard unification by exact string matching.}

\xxparatran
{このとき，NATLにおける推論は2つの項のクラスの組み合わせとして次の5つの類型に分けられる．
すなわち，
$\mathcal{S}\cdot\mathcal{S}$，
$\mathcal{S}\cdot\mathcal{C}$，
$\mathcal{S}\cdot\mathcal{L}$，
$\mathcal{L}\cdot\mathcal{C}$，
$\mathcal{L}\cdot\mathcal{L}$．
これらの5つの型における推論は全て，この後で例示するように，単一化可能な共通項を介した代入操作によって導かれる.}
{At this time, inference in NATL is divided into the following five types as a combination of classes of two terms.
That is,
$ \mathcal {S} \cdot \mathcal {S} $,
$ \mathcal {S} \cdot \mathcal {C} $,
$ \mathcal {S} \cdot \mathcal {L} $,
$ \mathcal {L} \cdot \mathcal {C} $,
$ \mathcal {L} \cdot \mathcal {L} $.
All inferences in these five types are guided by assignment operations via unifying common terms, as illustrated below.}
{At this time, inference in NATL as a combination of classes of two terms is divided into the following five types. That is, 
$\mathcal{S}\cdot\mathcal{S}$，
$\mathcal{S}\cdot\mathcal{C}$，
$\mathcal{S}\cdot\mathcal{L}$，
$\mathcal{L}\cdot\mathcal{C}$，
$\mathcal{L}\cdot\mathcal{L}$．
The inferences in these five types are all derived based on the substitution operations by common terms that can be simplified, as illustrated later.
}
{
Reasoning in NATL as a combination of classes of two terms is divided into the following five types. That is, 
$\mathcal{S}\cdot\mathcal{S}$,
$\mathcal{S}\cdot\mathcal{C}$,
$\mathcal{S}\cdot\mathcal{L}$,
$\mathcal{L}\cdot\mathcal{C}$,
$\mathcal{L}\cdot\mathcal{L}$.
The reasoning in these five types are all derived based on the substitution operations by common terms that can be unified, as illustrated later.}

\xxparatran
{$\mathcal{S,L,C}$間の可能な組み合わせの残りの1つ
$\mathcal{C}\cdot\mathcal{C}$においては，
型全体にわたって一般化可能な操作を定義できない\footnote{%
正確に言えば，推論という観点から意味のある操作を見いだせない．
例えば，NALにおいて$U \times V \to R$の形式を用いて表示されていたものが，
NATLにおいては$C:(R, U, V)$の形式を用いて表示されることとの対応から，
$C_1:(R_1, U_1, V_1)$と$C_2:(R_2, U_2, V_2)$をそれぞれ
$S_1:(U_1, V_1) \to R_1$と$S_2:(U_2, V_2) \to R_2$と変形した上で，
$\{S_1,S_2\} \vdash C_3:((U_2, V_2), U_1, V_1)$ という結論を得る
形式的操作を定義することはできるが，この操作により得られた$C_3$がどういう意味を持ちうるのかは不明である．}．
一方で，自然言語を用いて伝達される物語や議論では多くの場合，
複合項すなわち事態の系列だけが提示される．
この複合項の系列の間，つまり事物項の連鎖$\mathcal{C}\cdot\mathcal{C}$を，
論理項を用いて補間的に接続することがNATLにおける「推論」の本質であり，
その接続のあり方を規定するものが，
冒頭で触れた心理的・社会的・技術的な立場での，NATLにおける「論理」である%
\footnote{%
例えば ``Ann likes pandas. She goes to the zoo everyday.''というようなナラティブはありふれたものだが，
これは，$C_1:(\term{likes}, \term{Ann}, \term{pandas})$, $C_2:(\term{goes\mhyp to}, \term{she}, \term{the\mhyp zoo}, \term{everyday})$ として事物項の連鎖で表示される．
事物項$C_1, C_2$から，何のために動物園に行くのか，動物園の中のどこにいるのか，といったことに関する多様な推論を行えるが，それらは「パンダと動物園の関係」や「動物園における人の行動」に関する一般的な知識を表現する論理項（陳述項と連関項）を用いて，事物項$C_1, C_2$を接続することで実現されるとNATLでは考える．}．
すなわちNATLは，人間が日常的に行う推論が，
3つのクラスの知識表現上の5つの型の記号操作で記述し尽くせると主張する理論である．
この主張が妥当であるかは，今後の検証を待つ必要があるが，
次章\S\ref{sec:argumentation}でその有望性を例証する．}
{In the remaining one of the possible combinations between $\mathcal{S,L,C}$, $\mathcal{C}\cdot\mathcal{C}$, we cannot define an operation that is generalizable across types\footnote{%
Strictly speaking, we cannot find any operations that make sense from the point of view of reasoning.
For example, what was displayed using the format $U \times V \to R$ in NAL is now displayed using the format $C:(R, U, V)$ in NATL. From the correspondence, $C_1:(R_1, U_1, V_1)$ and $C_2:(R_2, U_2, V_2)$ are changed to $S_1:(U_1, V_1) \to R_1$ and $S_2:(U_2, V_2) \to R_2$ and then define a formal operation that leads to the conclusion $\{S_1,S_2\}\vdash C_3:((U_2, V_2), U_1, V_1)$, but this operation It is unclear what the meaning of $C_3$ obtained by is.
}.
On the other hand, in stories and discussions conveyed using natural language, in many cases, only compound terms, that is, sequences of events, are presented.
The essence of ``reasoning'' in NATL is to interpolate the series of compound terms, that is, the chain of matter terms, $\mathcal{C}\cdot\mathcal{C}$, using logical terms. The ``logic'' of NATL from the psychological, social, and technical standpoints mentioned at the beginning is what defines how these connections should be%
\footnote {%
For example, ``Ann likes pandas. She goes to the zoo everyday.'' is a common narrative. $C_1:(\term{likes}, \term{Ann}, \term{pandas})$, $C_2:(\term{goes\mhyp to}, \term{she}, \term{the\mhyp zoo}, \term{everyday})$ in a chain of things.
From the items $C_1 and C_2$, we can make various inferences about why we go to the zoo and where we are in the zoo. NATL considers that it is realized by connecting the matter terms $C_1 and C_2$ using logical terms (statement term and association term) that express general knowledge about "behavior of a person".}.
In other words, NATL is a theory that asserts that human reasoning can be fully described by five types of symbolic operations on three classes of knowledge representations.
Whether or not this assertion is valid needs to be verified in the future, but its promise is illustrated in the next chapter \S\ref{sec:argumentation}.}
{In the remaining one possible combination $\mathcal{C}\cdot\mathcal{C}$ between $\mathcal{S,L,C}$, we cannot define an operation that is generalizable over the whole types\footnote {%
To be exact, meaningful operation cannot be found from the perspective of reasoning. For example, from what is represented by using the format of $U \times V \to R$ in NAL corresponding to the what is represented by using the format of $C:(R, U, V)$ in NATL. From the correspondence, $C_1:(R_1, U_1, V_1)$ and $C_2:(R_2, U_2, V_2)$ are changed to $S_1:(U_1, V_1) \to R_1$ and $S_2:(U_2, V_2) \to R_2$ and then define a formal operation that leads to the conclusion $\{S_1,S_2\}\vdash C_3:((U_2, V_2), U_1, V_1)$ respectively, whereas the meaning of $C_3$ obtained by this operation is unclear.}. 
Meanwhile, stories and argumentation conveyed in natural language often present only compound terms, like a series of events. The essence of ``inference'' in NATL is the complementary connection between this series of compound terms, that is, the chain of thing terms $\mathcal{C}\cdot\mathcal{C}$, using logical terms. And the ``logic'' in NATL, from the psychological, social and technical standpoints mentioned at the beginning is what defines how these connections should be
\footnote{%
For example, ``Ann likes pandas. she goes to the zoo every day.'' such a narrative is very common because $C_1:(\term{likes}, \term{Ann}, \term{pandas})$, $C_2:(\term{goes\mhyp to}, \term{she}, \term{the\mhyp zoo}, \term{everyday})$ are presented by the chain of thing terms. 
From the thing terms $C_1, C_2$, related, various inferences can be made such as why to go to the zoo and where to be in the zoo. NATL considers that it is realized by chaining thing terms $C_1, C_2$ using logical terms (statement terms and linkage terms) presented by generalized knowledge related to ``The relationship between pandas and the zoo'' and ``The action of people who are in the zoo''. }.
In other words, NATL is a theory that argues that inferences that humans make in daily life can be completely described by five types of symbolic operations on three classes of knowledge representations. 
Although whether this claim is valid needs to be verified in the future, the next chapter \S\ref{sec:argumentation} will illustrate its promise.
}
{
In the remaining one possible combination type $\mathcal{C}\cdot\mathcal{C}$ between $\mathcal{S,L,C}$, we cannot define an operation that is generalizable over the type.\footnote {%
To be precise, a meaningful operation cannot be found from the perspective of reasoning. For example, from what is represented by using the format of $U \times V \to R$ in NAL corresponding to the what is represented by using the format of $C:(R, U, V)$ in NATL. From the correspondence, $C_1:(R_1, U_1, V_1)$ and $C_2:(R_2, U_2, V_2)$ are transformed to $S_1:(U_1, V_1) \to R_1$ and $S_2:(U_2, V_2) \to R_2$, and then we can define a formal operation that leads to the conclusion $\{S_1,S_2\}\vdash C_3:((U_2, V_2), U_1, V_1)$ respectively, whereas the meaning of $C_3$ obtained by this operation is unclear.} 
Meanwhile, narratives and argumentation conveyed in natural language often present a series of events, i.e., compound terms. 
The essence of ``reasoning'' in NATL is the complementary connection between this series of compound terms, that is, the chain of thing terms $\mathcal{C}\cdot\mathcal{C}$, using logical terms. 
And the ``logic'' in NATL, from the psychological, social and technological standpoints mentioned at the beginning of this paper is what defines how these connections should be.\footnote{%
Consider, for example, a common narrative such as ``Ann likes pandas. She goes to the zoo every day.''
This narrative can be expressed as $C_1:(\term{likes}, \term{Ann}, \term{pandas})$, $C_2:(\term{goes\mhyp to}, \term{she}, \term{the\mhyp zoo}, \term{everyday})$ as a series of thing terms. 
From the thing terms $C_1, C_2$, related, various inferences can be made such as why to go to the zoo and where to be in the zoo. 
NATL considers that it is realized by the connection of thing term $C_1, C_2$ and using logic terms (statement terms and linkage terms) presented by generalized knowledge related to ``the relationship between pandas and the zoo'' and ``the action of people who are in the zoo''. }
In other words, NATL is a theory that argues that inferences that humans make in daily life can be completely described by five types of symbolic operations on three classes of knowledge representations. 
Although whether this claim is valid needs to be verified in the future, the next section \S\ref{sec:argumentation} will illustrate its promise.}

\xxparatran
{ここで，\S\ref{sec:separation_of_logic_and_language}でも触れた，
明示的・熟考的な推論と，非明示的・反射的な推論との関係性について，
NATLの理論的立場を示す．
事態間の繋がりに関する「論理」が個人の中で習慣化されると，
論理項による明示的な推論経路の導出を経ずに，
反射的にすばやく推論が行われるようになると考える．
そしてその論理が社会的に慣習化されると，言語化されなくなる．
（このことは次章\S\ref{sec:argumentation}で触れるwarrantの暗黙性に関係する．）
そして，大規模生成言語モデルによる推論（例えば\cite{Bhagavatula20}）は
まさにこれを行っていると捉えるのがNATLの立場である．}{
Here, we present the theoretical position of NATL on the relationship between explicit/deliberative reasoning and implicit/reflective reasoning, which was touched on in \S\ref{sec:separation_of_logic_and_language}.
We believe that when the ``logic'' regarding the connections between situations becomes habitual within an individual, inferences will be made reflexively and quickly without derivation of explicit inference paths by logical terms.
And when that logic becomes socially customary, it ceases to be verbalized.
(This is related to the implicitity of warrants, which is covered in the next chapter, \S\ref{sec:argumentation}.)
NATL's position is that inference by large-scale generative language models (eg \cite{Bhagavatula20}) does exactly this.}
{
As also mentioned in \S\ref{sec:separation_of_logic_and_language}, we present the theoretical position of NATL on the relationship between explicit, deliberative reasoning and implicit, reflective reasoning. We believe that when the ``logic'' regarding the connections between situations becomes habitual within an individual, inferences will be made reflexively and quickly without derivation of explicit inference paths by logical terms. 
And when that logic becomes socially customary, it ceases to be verbalized (This is related to the implicity of warrants, which is covered in the next chapter, \S\ref{sec:argumentation}). NATL's position is that inference based on large-scale generative language models (for example, \cite{Bhagavatula20}) exactly catches that phenomenon.
}
{
As also mentioned in \S\ref{sec:separation_of_logic_and_language}, we present the theoretical position of NATL on the relationship between explicit, deliberative reasoning and implicit, reflective reasoning. We believe that when the ``logic'' regarding the connections between situations becomes habitual within an individual, inferences will be made reflexively and quickly without derivation of explicit inference paths by logic terms. 
And when that logic becomes socially customary, it ceases to be verbalized (This is related to the implicity of warrants, which is covered in the next section, \S\ref{sec:argumentation}). 
NATL's position is that inference based on large-scale generative language models (for example, \cite{Bhagavatula20}) exactly catches that phenomenon.}

\xxparatran
{一方で，社会的に慣習化・共有化されていない論理について他者の了解を得るには，
論理項による明示的な推論経路の導出と提示が必要である．
そのような場合に，導出された推論経路に基づいて実際に
提示される「説明」に対して知覚されるその質の良し悪しは，
共有されていない部分について明示的な推論経路（論理項）
が必要十分に示されているかどうか，ということと強く相関することをNATLは予想する．
この「説明」において，十分に慣習化されている部分まで事細かに示せば，
逆に冗長という負の評価をくだされると予想する．
これらのことから，NATLは実際の人間の推論・論理のあり方に関する理論としても，
一定の反証可能性を備える．}
{On the other hand, in order to gain the understanding of others about logic that is not socially customary or shared, it is necessary to derive and present an explicit inference path using logic terms.
In such cases, the perceived quality of the ``explanation'' actually presented based on the derived inference path depends on the explicit inference path (logical term) for the unshared part. NATL predicts that there is a strong correlation between whether or not it is adequately represented.
In this ``explanation'', if I show in detail the parts that are sufficiently customary, I expect that it will be negatively evaluated as redundant.
For these reasons, NATL has a certain degree of falsifiability as a theory of how human reasoning and logic should be.}
{On the other hand, in order to gain the understanding of others about logic that is not socially customary or shared, it is necessary to derive and present an explicit inference path using logic terms. 
In such cases, the perceived quality of the ``explanation'' actually presented based on the derived inference path depends on the explicit inference path (logical term) for the unshared part. NATL predicts that there is a strong correlation between whether or not it is adequately represented. 
In this ``explanation'', if the part that has been fully habitual is shown in detail, we will expect that it is negatively evaluated as redundant. 
For these reasons, NATL has a certain degree of falsifiability as a theory of how human reasoning and logic should be.
}
{
On the other hand, in order to gain the understanding of others about logic that is not socially customary or shared, it is necessary to derive and present an explicit inference path using logic terms. 
In such cases, NATL predicts that there is a strong correlation between the perceived quality of the ``explanation'' actually presented based on the derived inference path and the explicitly explained inference path (logic terms) for the unshared part. 
In this ``explanation'', if the part that has been fully habitual is shown in detail, NATL expects that it is negatively evaluated as redundant. 
Thus, NATL has a certain degree of falsifiability as a theory of how human reasoning and logic should be.}


\subsection{Reasoning types}\label{sec:reasoning_patterns}

\xxparatran
{以下の例では，陳述項の繋辞を継承関係（is-a），連関項の繋辞を含意関係を表すと想定する．
一般に得られた結論の妥当性は前提の繋辞の意味によって異なるはずだが，
ここでは推論の操作として可能な類型を示す．}
{In the following examples, we assume that the affix of the statement term represents the inheritance relationship (is-a) and the affix of the association term represents the entailment relationship.
In general, the validity of the conclusion obtained should differ depending on the meaning of the premise affixes, but here we show possible types of inference operations.}
{In the following examples, we assume that the copula of the statement term represents the inheritance relationship (is-a) and the copula of the linkage term represents the entailment relationship. In general, the validity of the conclusion obtained should be different depending on the meaning of the premise copula, but here we show possible types of inference operations.
}
{
In the following examples, we assume that the copula of statement terms represents the inheritance relationship (is-a) and the copula of linkage terms represents the entailment relationship. In general, the validity of the conclusion obtained should be different depending on the meaning of the premise copulas, and here we show possible types of inference operations regardless of the validity of conclusion.}

\subsubsection{$\mathcal{S}\cdot\mathcal{S} \to \mathcal{S}$}

\xxparatran 
{陳述項2つを取る推論で，三段論法の基本形である．
結論は陳述項になる．}{
An inference that takes two statements, and is the basic form of a syllogism.
The conclusion becomes a statement.}
{
Taking two statements to make an inference is the basic form of a syllogism. The conclusion is also a statement term.}
{
Taking two statements to make an inference is the basic form of a syllogism. The conclusion is also a statement term.}

\xxparatran
{\S\ref{sec:logics}の冒頭で示した演繹の例は以下のようになる．
\begin{quote}
\begin{tabular}{ll}
$S_1:$& $\term{human} \to \term{animal}$ \\
$S_2:$& $\term{animal} \to \term{mammal}$ \\
\hline
$S_3:$& $\term{human} \to \term{mammal}$
\end{tabular}
\end{quote}}
{The deductive example given at the beginning of \S\ref{sec:logics} looks like this:
\begin{quote}
\begin{tabular}{ll}
$S_1:$& $\term{human} \to \term{animal}$ \\
$S_2:$& $\term{animal} \to \term{mammal}$ \\
\hline
$S_3:$& $\term{human} \to \term{mammal}$
\end{tabular}
\end{quote}}
{The deductive example given at the beginning of \S\ref{sec:logics} looks like this:
\begin{quote}
\begin{tabular}{ll}
$S_1:$& $\term{human} \to \term{animal}$ \\
$S_2:$& $\term{animal} \to \term{mammal}$ \\
\hline
$S_3:$& $\term{human} \to \term{mammal}$
\end{tabular}
\end{quote}
}
{The deductive example given at the beginning of \S\ref{sec:logics} looks like this:
\begin{quote}
\begin{tabular}{ll}
$S_1:$& $\term{human} \to \term{animal}$ \\
$S_2:$& $\term{animal} \to \term{mammal}$ \\
\hline
$S_3:$& $\term{human} \to \term{mammal}$
\end{tabular}
\end{quote}
}

\xxparatran
{\S\ref{sec:rules}で示した帰納とアブダクションの例を同じく示す．
\begin{quote}
\begin{tabular}{ll}
$S_4:$& $\term{Lily} \to \term{swan}$ \\
$S_5:$& $\term{Lily} \to \term{white}$ \\
\hline
$S_6:$& $\term{swan} \to \term{white}$
\end{tabular}
\end{quote}}
{We also show examples of induction and abduction given in \S\ref{sec:rules}.
\begin{quote}
\begin{tabular}{ll}
$S_4:$& $\term{Lily} \to \term{swan}$ \\
$S_5:$& $\term{Lily} \to \term{white}$ \\
\hline
$S_6:$& $\term{swan} \to \term{white}$
\end{tabular}
\end{quote}}
{We also show examples of induction and abduction given in \S\ref{sec:rules}.
\begin{quote}
\begin{tabular}{ll}
$S_4:$& $\term{Lily} \to \term{swan}$ \\
$S_5:$& $\term{Lily} \to \term{white}$ \\
\hline
$S_6:$& $\term{swan} \to \term{white}$
\end{tabular}
\end{quote}}
{We also show examples of induction and abduction given in \S\ref{sec:rules}.
\begin{quote}
\begin{tabular}{ll}
$S_4:$& $\term{Lily} \to \term{swan}$ \\
$S_5:$& $\term{Lily} \to \term{white}$ \\
\hline
$S_6:$& $\term{swan} \to \term{white}$
\end{tabular}
\end{quote}}

\begin{quote}
\def\tf{\term{these\mhyp beans}}
\def\ts{\term{beans\mhyp from\mhyp the\mhyp bag}}
\begin{tabular}{ll}
$S_7:$& $\tf \to \term{white}$ \\
$S_8:$& $\ts \to \term{white}$ \\
\hline
$S_9:$& $\tf \to \ts$
\end{tabular}
\end{quote}

\xxparatran
{$S_6$や$S_9$が逆向きでない（例えば，$S_6':\term{white}\to\term{swan}$ではない）
形式的・理論的な理由はない．
$S_6'$よりも$S_6$が選好されるならば，
それは推論器に事前に組み込まれたアドホックな基準
（人間であれば認知バイアス）や，
推論器が経験から獲得した傾向によるものと考える．}{
There is no formal or theoretical reason why $S_6$ and $S_9$ should not be backwards (eg $S_6':\term{white}\to\term{swan}$).
If $S_6$ is preferred over $S_6'$, it may be due to ad-hoc criteria (cognitive bias in humans) built into the reasoner in advance, or tendencies acquired by the reasoner from experience.}
{There is no formal or theoretical reason why $S_6$ and $S_9$ should not be backwards (for example, $S_6':\term{white}\to\term{swan}$). 
If $S_6$ is preferred over $S_6'$, it may be due to ad-hoc criteria (cognitive bias in humans) built into the reasoner in advance, or tendencies that the reasoner has acquired through experience.
}
{
There is no formal or theoretical reason why $S_6$ and $S_9$ should not be backwards (for example, $S_6':\term{white}\to\term{swan}$). 
If $S_6$ is preferred over $S_6'$, it may be due to ad-hoc criteria built into the reasoner in advance (cognitive bias in humans), or tendencies that the reasoner has acquired through experience.}

\subsubsection{$\mathcal{S}\cdot\mathcal{C} \to \mathcal{C}$}\label{sec:SCtoC}

\xxparatran
{複合項の要素項が陳述項に従って置換される．}
{Component terms of a compound term are replaced according to the statement term.}
{The element term of the compound term is replaced according to the statement term.
}
{
An element term of the given compound term is replaced according to the given statement term.}

\begin{quote}
\begin{tabular}{ll}
$S_1:$& $\term{polar\mhyp bear} \to \term{white}$ \\
$C_1:$& $(\term{likes}, \term{John}, \term{white})$ \\
\hline
$C_2:$& $(\term{likes}, \term{John}, \term{polar\mhyp bear})$
\end{tabular}
\end{quote}

\xxparatran
{以下のような逆方向の推論も許容されるが，一般に推論結果の確からしさは低くなると予想される．}
{Reverse inferences such as the following are also allowed, but generally the certainty of the inference results is expected to be low.}
{Although reverse inferences such as the following are also allowed, generally the accuracy of inference results is expected to be low.
}
{
Although reverse inferences such as the following are also allowed, generally the certainty of inference results is expected to be low.}

\begin{quote}
\begin{tabular}{ll}
$S_1:$& $\term{polar\mhyp bear} \to \term{white}$ \\
$C_2:$& $(\term{likes}, \term{John}, \term{polar\mhyp bear})$ \\
\hline
$C_1:$& $(\term{likes}, \term{John}, \term{white})$
\end{tabular}
\end{quote}

\subsubsection{$\mathcal{S}\cdot\mathcal{L} \to \mathcal{C/S/L}$}\label{sec:SCtoCSL}

\xxparatran
{連関項の2つの要素項のうちのどちらかと陳述項が単一化可能な場合，
連関項の残りの要素項が結論として取り出される．
この場合，帰結される項のクラスは与えられた連関項に依存する
（$\mathcal{C/S/L}$の$/$は排他的選言を意味する）．
また，単一化された変数項があれば結論でもその束縛が維持される．}
{If either of the two constituent terms of the associated term and the statement term are unifiable,
The remaining component terms of the associated term are taken as the conclusion.
In this case the class of resulting terms depends on the associated terms given ($/$ in $\mathcal{C/S/L}$ means exclusive disjunction).
Also, if there are unification variable terms, their bindings are preserved in the conclusion.}
{In the case that one of the two element terms of the linkage term and the statement term are unifiable, the remaining element terms of the linkage term are taken as the conclusion. 
In this case, the category of the consequent terms depends on the given linkage item ($\mathcal{C/S/L}$の$/$ means exclusive disjunction). 
Also, if there are unification variable terms, their bindings are preserved in the conclusion.
}
{
In the case that one of the two element terms of the linkage term and the statement term are unifiable, the remaining element terms of the linkage term are taken as the conclusion. 
In this case, the category of the consequent terms depends on the given linkage item ($/$ in $\mathcal{C/S/L}$ means exclusive disjunction). 
Also, if there are unified variable terms, their bindings are preserved in the conclusion.}

\begin{quote}
\begin{tabular}{ll}
$S_1:$& $\term{polar\mhyp bear} \to \term{white}$ \\
$L_1:$& $(x \to \term{white}) \To (\term{likes}, \term{John}, x)$ \\
\hline
$C_2:$& $(\term{likes}, \term{John}, \term{polar\mhyp bear})$
\end{tabular}
\end{quote}

\xxparatran
{$L_1$は「ある$x$が白いならばJohnは$x$が好き」という意味であり，
上記の例は，\S\ref{sec:SCtoC}における$\{S_1, C_1\}\vdash C_2$の例と内容的に同じである．}
{$L_1$ means ``John likes $x$ if some $x$ is white'', and the above example is equivalent to $\{S_1, C_1\}\ in \S\ref{sec:SCtoC} The content is the same as the vdash C_2$ example.}
{$L_1$ means ``If there are x which are white, John will like $x$'', and the above example is equivalent to the content of $\{S_1, C_1\}\vdash C_2$ in \S\ref{sec:SCtoC}.}
{
$L_1$ means ``If there are $x$ which are white, John likes $x$'', and the above example is equivalent to the content of $\{S_1, C_1\}\vdash C_2$ in \S\ref{sec:SCtoC}.}

\xxparatran
{このように，変数項を用いることで複合項は連関項に翻訳することができる．
ただし，下の例の$L_2$が示すように任意の連関項を複合項に翻訳できるわけではないので，
$\mathcal{S}\cdot\mathcal{C}$型の推論ができれば
$\mathcal{S}\cdot\mathcal{L}$型を扱う必要がないことにはらないし，その逆も同様である．}{
In this way, compound terms can be translated into associated terms by using variable terms.
However, as shown by $L_2$ in the example below, it is not possible to translate an arbitrary associated term into a compound term. 
It doesn't mean that you don't have to deal with $\mathcal{S}\cdot\mathcal{L}$ types, and vice versa.}{
In this way, compound terms can be translated to linkage terms by using variable terms. However, it is impossible to translate any linkage terms into compound terms as shown by $L_2$. 
If we can infer the $\mathcal{S}\cdot\mathcal{C}$ type, it doesn't mean we don't need to deal with the $\mathcal{S}\cdot\mathcal{L}$ type, and the reverse is true as well.
}
{
In this way, some compound terms can be translated to linkage terms by using variable terms. However, as shown by $L_2$, it is not always possible to translate arbitrary linkage terms into compound terms.
Therefore, we need both the $\mathcal{S}\cdot\mathcal{C}$ and $\mathcal{S}\cdot\mathcal{L}$ types of reasoning.}

\def\slf{\term{weather\mhyp of\mhyp today}}
\def\tlf{\term{weather\mhyp of\mhyp the\mhyp day}}
\begin{quote}
\begin{tabular}{ll}
$S_2:$& $\slf \to \term{bad}$\\
$L_2:$& $(\tlf \to \term{bad}) \To \term{no\mhyp school}$ \\
\hline
$B_1:$& $\term{no\mhyp school}$
\end{tabular}
\end{quote}

\xxparatran
{この例の$L_2$は「天気が悪いならば休校」という規則を表している．
$\slf$と$\tlf$は完全に同一の概念ではないが，
適切な推論の文脈のもとでは単一化が許されるとここでは前提している．
いつ許されるのかは，\S\ref{sec:reasoner}で後述する推論器が経験から判断する．}
{$L_2$ in this example represents the rule "If the weather is bad, the school will be closed."
Although $\slf$ and $\tlf$ are not exactly the same concept, we assume here that unification is permissible under the appropriate context of reasoning.
When it is permissible is determined empirically by the reasoner described later in \S\ref{sec:reasoner}.}
{
$L_2$ in this example represents the rule of ``If the weather is bad, the school will be closed''. Although $\slf$ and $\tlf$ are not exactly the same concept, the premise here is to allow simplification in an appropriate inferential context. When it is permissible is determined empirically by the reasoner described later in \S\ref{sec:reasoner}.}
{
$L_2$ in this example represents the rule of ``If the weather is bad, the school will be closed''. Although $\slf$ and $\tlf$ are not exactly the same concept, the premise here is to allow the unification of them in an appropriate inferential context. When it is allowed is determined empirically by the reasoner described later in \S\ref{sec:reasoner}.}

\subsubsection{$\mathcal{C}\cdot\mathcal{L} \to \mathcal{C/S/L}$}

\xxparatran
{連関項の2つの要素項のうちのどちらかと事物項が単一化可能な場合，
連関項の残りの要素項が結論として取り出される．
この場合も，帰結される項のクラスは与えられた連関項に依存する．}{
If either of the two component terms of the associated term is unifiable with the matter term, the remaining component term of the associated term is taken as the conclusion.
Again, the class of resulting terms depends on the associated terms given.}
{
When one of the two element terms of linkage terms and the thing terms are unifiable, the remaining element terms of the linkage items are taken as a conclusion. 
In this case, the category of the consequent terms depend on the given linkage terms.}
{
When one of the two element terms of a linkage term and the thing term are unifiable, the remaining element terms of the linkage items are taken as a conclusion. 
In this case, the class of the consequent terms depend on the given linkage terms.}

\begin{quote}
\begin{tabular}{ll}
$C_1:$& $(\term{likes}, \term{John}, \term{polar\mhyp bear})$\\
$L_1:$& $(\term{likes}, x, \term{polar\mhyp bear}) \To (\term{likes}, x, \term{penguin})$ \\
\hline
$C_2:$& $(\term{likes}, \term{John}, \term{penguin})$
\end{tabular}
\end{quote}

\subsubsection{$\mathcal{L}\cdot\mathcal{L}\to\mathcal{L}$}

\xxparatran
{$\mathcal{S}\cdot\mathcal{S}\to\mathcal{S}$と同じように，
単一化可能な要素項2項を介して，残りの要素項2項が新たな連関項を構成する．
ここでは演繹の例だけを示す．}{
As with $\mathcal{S}\cdot\mathcal{S}\to\mathcal{S}$,
Through the unifiable elemental 2 terms, the remaining elemental 2 terms form a new associated term.
Here we only give a deductive example.}
{
Like $\mathcal{S}\cdot\mathcal{S}\to\mathcal{S}$, two remaining element terms construct a new linkage term via two element terms that are unifiable. Here we only give a deductive example.}
{
Like $\mathcal{S}\cdot\mathcal{S}\to\mathcal{S}$, two remaining element terms construct a new linkage term via the two element terms that are unifiable. Here we only give a deductive example.}

\begin{quote}
\begin{tabular}{ll}
$L_1:$& $(\term{likes}, x, \term{polar\mhyp bear}) \To (\term{likes}, x, \term{penguin})$ \\
$L_2:$& $(\term{likes}, y, \term{penguin}) \To (\term{likes}, y, \term{dolphin})$ \\
\hline
$L_3:$& $(\term{likes}, x, \term{polar\mhyp bear}) \To (\term{likes}, x, \term{dolphin})$ \\
\end{tabular}
\end{quote}

\section{Elucidating the Internal Structures of Arguments} \label{sec:argumentation}

\xxparatran{本章では
非公理的項論理NATLの正当性・有効性を
「議論」の定性的分析を通じて示すことを試みる．
アブダクションと繋がりの深い関連性理論\cite{RelevanceTheory}を提出したSperberも，
Mercierと伴に，議論こそが推論（および理性）の主たる機能であると主張している\cite{Mercier2017}．
NATLは人間の日常的推論を扱おうとする理論であるから，
その理論の枠組みのなかで，人間の日常の議論の内容を表示し，組み立てることができなければならない．}{
This chapter attempts to demonstrate the validity and effectiveness of non-axiomatic term logic NATL through qualitative analysis of ``arguments''. Sperber, who proposed a relevance theory closely related to abduction, \cite{RelevanceTheory}, also argues with Mercier that argument is the main function of reasoning (and reasoning)\cite{Mercier2017}. Since NATL is a theory that deals with everyday human reasoning, it must be possible to display and construct the contents of human everyday arguments within the framework of the theory.
}{
This chapter attempts to demonstrate the validity and effectiveness of non-axiomatic term logic NATL through qualitative analysis of ``argumentation''.
Sperber, proposed a theory of relevance\cite{RelevanceTheory} closely related to abduction, also argues along with Mercier, that argumentation is primary function of inference (or reasoning)\cite{Mercier2017}．
Since NATL is a theory that attempts to deal with human daily reasoning, it must be able to represent and construct the content of human daily argumentation within the framework of that theory.
}
{
This sectionr attempts to demonstrate the validity and effectiveness of NATL through qualitative analysis of arguments. Sperber, who proposed the relevance theory \cite{RelevanceTheory} closely related to abduction, also argues with Mercier, that argumentation is the primary function of reasoning (or human reason) \cite{Mercier2017}. Since NATL is a theory that attempts to deal with human daily reasoning, it must be able to represent and construct the content of human daily arguments within the framework of the theory.}

\xxparatran{\begin{figure}
    \centering
  \begin{tikzpicture} 
    \node (D) {D};
    \node[right=3cm of D] (C) {So, C};
    \node[below right=0.5cm and 0.75cm of D] (W) {Since W};
    \draw[->] (D) -- (C);
    \draw (W.north) -- ($(D)!(W.north)!(C)$);
  \end{tikzpicture}
    \caption{Toulminの論証図式~\cite{Toulmin58}}\label{fig:toulmin_diagram}
\end{figure}}
{\begin{figure}
    \centering
  \begin{tikzpicture} 
    \node (D) {D};
    \node[right=3cm of D] (C) {So, C};
    \node[below right=0.5cm and 0.75cm of D] (W) {Since W};
    \draw[->] (D) -- (C);
    \draw (W.north) -- ($(D)!(W.north)!(C)$);
  \end{tikzpicture}
    \caption{Toulmin's demonstrative diagram~~\cite{Toulmin58}}\label{fig:toulmin_diagram}
\end{figure}}
{\begin{figure}
    \centering
  \begin{tikzpicture} 
    \node (D) {D};
    \node[right=3cm of D] (C) {So, C};
    \node[below right=0.5cm and 0.75cm of D] (W) {Since W};
    \draw[->] (D) -- (C);
    \draw (W.north) -- ($(D)!(W.north)!(C)$);
  \end{tikzpicture}
    \caption{Toulmin's demonstrative diagram~\cite{Toulmin58}}\label{fig:toulmin_diagram}
\end{figure}
}
{
\begin{figure}
    \centering
  \begin{tikzpicture} 
    \node (D) {D};
    \node[right=3cm of D] (C) {So, C};
    \node[below right=0.5cm and 0.75cm of D] (W) {Since W};
    \draw[->] (D) -- (C);
    \draw (W.north) -- ($(D)!(W.north)!(C)$);
  \end{tikzpicture}
    \caption{Toulmin's diagram~\cite{Toulmin58}}\label{fig:toulmin_diagram}
\end{figure}
}

\xxparatran{\S\ref{sec:introduction}で参照したToulminも，
社会の中で一般的に行われる論証的議論の構造について，「Toulminモデル」とよばれる
図式的パターン（\ref{fig:toulmin_diagram}）を提示している\cite{Toulmin58}．
ここで，D (Datum) は論証の根拠，C (Claim) は主張，W (Warrant) は論拠を表す．
（Toulminの図式モデルにはこの他に，Wを裏付ける backing B,  
主張を限定する rebuttal Rなどの要素が含まれうる．）}{Toulmin, referred to in \S\ref{sec:introduction}, also called the ``Toulmin model'' about the structure of discursive arguments commonly held in society.
\cite{Toulmin58} presenting a schematic pattern (\ref{fig:toulmin_diagram}). where D (Datum) is the basis for the argument, C (Claim) is the assertion, and W (Warrant) is the argument.
(Toulmin's schematic model also includes backing B,Elements such as rebuttal R that qualify claims can be included. ))
}{
Toulmin referred to in \S\ref{sec:introduction} also presents a diagrammatic pattern called the ``Toulmin model'' （\ref{fig:toulmin_diagram}）for the structure of demonstrative argumentation commonly used in society\cite{Toulmin58}.
Here, D (Datum) is the basis of the argumentation, C (Claim) is the claim, and W (Warrant) is the argumentation.
(Toulmin's diagrammatic model also includes other elements such as backing B to support W and rebuttal R to limit the claim.)
}
{
Toulmin referred to in \S\ref{sec:introduction} also presents a diagrammatic pattern called the ``Toulmin model'' (Fig. \ref{fig:toulmin_diagram}) for the structure of demonstrative argumentation commonly used in society~\cite{Toulmin58}. Here, D, datum, is the basis of an argument, C is the claim of the argument, and W is the warrant that enables the claim to be drawn from the datum. (Toulmin's model also includes other elements such as backing B to support W and rebuttal R to limit the claim.)
}

\xxparatran{例えば，
D: ハリーはバミューダで生まれた，
C: ハリーはイギリス臣民だ，
W: バミューダで生まれた人間はイギリス臣民になる，という
議論\cite{Toulmin58}をNATLの推論$\{S_D, L_W\} \vdash S_C$として表示すると，次のようになる．}{
For example, D: Harry was born in Bermuda, C: Harry is a British subject, W: A person born in Bermuda is a British subject.
Viewing the argument \cite{Toulmin58} as the NATL reasoning $\{S_D, L_W\} \vdash S_C$ looks like this:
}
{
For example,
D: Harry was born in Bermuda,
C: Harry is a British subject,
W: A person born in Bermuda is a British subject.
Viewing the argumentation \cite{Toulmin58} as the NATL inference $\{S_D, L_W\} \vdash S_C$ looks like this:
}
{
Consider the following argumentation \cite{Toulmin58}:
D: Harry was born in Bermuda,
C: Harry is a British subject,
W: A person born in Bermuda is a British subject.
As the NATL inference $\{S_D, L_W\} \vdash S_C$, it is formalized as follows:
}
\begin{quote}
\begin{tabular}{ll}
$S_D:$& $\term{Harry} \to \term{born\mhyp in\mhyp Bermuda}$ \\
$L_W:$& $(x \to \term{born\mhyp in\mhyp Bermuda}) \To (x \to \term{British\mhyp subject})$ \\
\hline
$S_C:$& $\term{Harry} \to \term{British\mhyp subject}$ 
\end{tabular}
\end{quote}

\xxparatran{このようにToulminの論証図式も三段論法が元になっているが，
W，つまり$L_W$は絶対的規則（真理）ではないという点で，(4) 数理的立場 （\S\ref{sec:introduction}参照）から外れている．
（\S\ref{sec:introduction}で触れたような規則（脚注\ref{fn:choco}参照）も，このwarrantに該当する．）
また，ToulminはWに相当する論拠は一般に暗黙化されやすいことを指摘している．
Toulminモデルをもとに一般人によるweb上の議論データを収集・分析した\citet{habernal-2017-argumentation}も，
明確にwarrantといえる事例が見当たらないとして，
議論テキストのアノテーション用に彼らが作成した「修正Toulminモデル」からはwarrantを除外している．}
{In this way, Toulmin's argumentation scheme is also based on syllogisms.
W, that is, $L_W$, is not an absolute rule (truth), and thus deviates from (4) the mathematical standpoint (see \S\ref{sec:introduction}).
(The rule mentioned in \S\ref{sec:introduction} (see footnote \ref{fn:choco}) also falls under this warranty.) In addition, Toulmin argues that arguments corresponding to W are generally implied. He points out that it is easy.
\citet{habernal-2017-argumentation}, which collects and analyzes web discussion data by ordinary people based on the Toulmin model, was also created by them for annotation of discussion texts, as there were no cases that could be clearly warranted. "Modified Toulmin model" excludes warranties.}
{In this way, Toulmin's argumentation scheme is also based on the three-stage argumentation, but since W, the $L_W$ is not an absolute rule (truth), and thus deviates from the mathematical position (4)（see \S\ref{sec:introduction}）.
(The rule mentioned in \S\ref{sec:introduction} (see footnote \ref{fn:choco}) also falls under this warranty.)
In addition, Toulmin argues that the rationale for W is generally implied.
\citet{habernal-2017-argumentation}, who collected and analyzed web-based argumentation data by the general public based on the Toulmin model, also excluded warrants from the ``modified Toulmin model'' they created for the annotation of argumentation texts, because they did not find any cases that could be clearly called warrants.
}
{
In this way, Toulmin's argumentation scheme is also based on syllogism, but since W is not an absolute rule (truth), and thus deviates from the mathematical position (4) (see \S\ref{sec:introduction}). (The rule mentioned in \S\ref{sec:introduction} (see footnote \ref{fn:choco}) also falls under this warrant.) In addition, Toulmin argues that the rationale as W is generally implied. \citet{habernal-2017-argumentation}, who collected web-based argument data by the general public and analyzed them based on the Toulmin model, also excluded warrants from their ``modified Toulmin model'' they created for the annotation of argument texts, because they did not find any cases that could be clearly identified as warrants.
}

\xxparatran{一方で，Habernalらは議論の適切なwarrantを2択で認識するArgument Reasoning Comprehension タスク (ARCT)~\cite{habernal-etal-2018-ARCT,niven-kao-2019-probing}を提案している．
しかしここでのwarrantは上の$L_W$のような規則的なものに限られず，「議論をサポートする主張であり，
その主張を反転させることで結論が変わるもの」，としてより広く定義されている．
従って，ARCTにおけるwarrantは，
それが議論の中の論理構造（推論過程）にどのように位置づけられているのかが自明ではない．
以下では，\cite{habernal-etal-2018-ARCT,niven-kao-2019-probing}からの3つの例を用いて，
NATLがこれらの推論過程を説明しwarrantの位置づけを明確にできることを示す．
一方で，\S\ref{sec:natl}で述べた基本的な枠組みだけでは，議論およびその背後の推論を
実現するには足りない要素があることも確認する．}
{On the other hand, Habernal et al. proposed the Argument Reasoning Comprehension Task (ARCT)~\cite{habernal-etal-2018-ARCT,niven-kao-2019-probing} that recognizes the appropriate warrant of the argument with two alternatives.
However, the warrant here is not limited to regular ones like $L_W$ above. Therefore, it is not obvious how the warrant in ARCT is positioned in the logical structure (inference process) in the argument.
Below, we use three examples from \cite{habernal-etal-2018-ARCT,niven-kao-2019-probing} to show how NATL can explain these inference processes and clarify the placement of warranties. On the other hand, we also confirm that the basic framework described in \S\ref{sec:natl} is insufficient to realize the discussion and the reasoning behind it.
}{
On the other hand, Habernal et al. proposes the Argument Reasoning Comprehension Task (ARCT) \cite {habernal-etal-2018-ARCT, niven-kao-2019-probing} that recognizes the appropriate warrant of the argument with two alternatives.
However, the warrant here is not limited to regular ones such as $L_W$ above, but is defined more broadly as ``A claim that supports an argumentation and whose conclusion changes when the claim is reversed.''.
Therefore, it is not obvious how the warrant in ARCT is positioned in the logical structure (inference process) in the argumentation.
Below, we show that NATL can explain these inference processes and clarify the position of warrants using three examples from  \cite{habernal-etal-2018-ARCT,niven-kao-2019-probing}.
At the same time, we also confirm that the basic framework described in \S\ref{sec:natl} is insufficient to realize the argumentation and the reasoning behind it.
}
{
On the other hand, Habernal et al. proposes the Argument Reasoning Comprehension Task (ARCT) \cite {habernal-etal-2018-ARCT, niven-kao-2019-probing} that recognizes the appropriate warrant of the argument with two alternatives.
However, the warrant here is not limited to regular ones such as $L_W$ above, but is defined more broadly as ``A claim that supports an argument and whose conclusion changes when the claim is reversed.''.
Therefore, it is not obvious how the warrant in ARCT is positioned in the logical structure (inference process) in the argumentation.
Below, we show that NATL can explain these inference processes and clarify the position of warrants using three examples from \cite{habernal-etal-2018-ARCT,niven-kao-2019-probing}.
At the same time, we also confirm that the basic framework described in \S\ref{sec:natl} is insufficient to fully realize the argumentation and the reasoning behind it.}

\subsection{Example 1: Should I bring an umbrella because it will rain?}\label{sec:ex_Umbrella}

\xxparatran{最初の例は以下の3文からなる（\cite{niven-kao-2019-probing}，p.1）．
D: It is raining.
C: You should take an umbrella.
W: It is bad to get wet.
この例では，Wを It is good to get wet. にすると，結論も You should not take an umbrella. になる．}
{The first example consists of the following three sentences (\cite{niven-kao-2019-probing}, p.1).
D: It is raining.
C: You should take an umbrella.
W: It is bad to get wet.
In this example, if W is It is good to get wet, the conclusion is You should not take an umbrella.
}
{The first example consists of the following three sentences(\cite{niven-kao-2019-probing}，p.1).
D: It is raining.
C: You should take an umbrella.
W: It is bad to get wet.
In this example, if W sets to ``It is good to get wet.'', the conclusion will be ``You should not take an umbrella.''.
}
{
The first example consists of the following three sentences (\cite{niven-kao-2019-probing}, p.1).
D: It is raining.
C: You should take an umbrella.
W: It is bad to get wet.
In this example, if W sets to ``It is good to get wet.'', the conclusion will be ``You should not take an umbrella.''
}

\xxparatran{これを項表示言語に翻訳すると，例えば以下の$S_D$, $C_C$, $S_W$ のようにできる．
Cの文が結論であるので，$C_C$を最後に示している．}{
If this is translated into term display language, for example, $S_D$, $C_C$, $S_W$ below. $C_C$ is shown last because the C statement is the conclusion.
}{
If this is translated into a term representation language, it can be made as follows, for example, $S_D$, $C_C$, $S_W$.
Since the sentence of C is the conclusion, $C_C$ is shown at the end.
}
{
This can be translated in TRL as $S_D$, $C_C$, $S_W$ as follows.
$C_C$ is shown last because sentence C is the conclusion}
\begin{quote}
\begin{tabular}{ll}
$S_D$:& $\tlf \to \term{raining}$ \\
$S_W$:& $\term{getting\mh wet} \to \term{bad}$ \\
$C_C$:& $(\term{take}, \term{You}, \term{umbrella})$ \\
\end{tabular}
\end{quote}

\xxparatran{明らかに$C_C$を直接$S_D$, $S_W$から導出することはできない．
ここでwarrant $S_W$も含めて明示的に示されていることだけから仮定できるのは.}{
Obviously $C_C$ cannot be directly derived from $S_D$, $S_W$.
What can be assumed only from the fact that the warranty $S_W$ is explicitly indicated here.
}{
Obviously, $C_C$ cannot be derived directly from $S_D$ and $S_W$.
All of that can be assumed from what is explicitly shown here, including the warrant $S_W$, is the existence of the rule (wisdom of life) below.
}
{
Obviously, $C_C$ cannot be derived directly from $S_D$ and $S_W$.
All what can be assumed from what is explicitly shown here, including warrant $S_W$, is the existence of the rule (wisdom of life) below.}
\begin{quote}
\begin{tabular}{rl}
$L_0:$& $(and, (\tlf \to \term{raining}), (\term{getting\mh wet} \to \term{bad})) \To (\term{take}, \term{one}, \term{umbrella})$
\end{tabular}
\end{quote}
\xxparatran{という規則（生活の知恵）の存在である．$L_0$で述べられていることは常識的に間違っていないし，このような規則が
一般的に人々に共有されていたとしてもおかしくはなさそうである．
しかし，そもそも雨というものがなにか，傘というものがなにかを知らない人間には，これだけでは説明していることにならない．
同様にこの規則があるということだけからは，
WがW': It is good to get wet. になったときに，規則の帰結が
C': You should not take an umbrella. となるべきであることも説明できない．
これではこのD, C, Wの3文から構成される議論を理解・説明できていることにはならない．}{This is the existence of the rule (wisdom of life). What is stated in $L_0$ is not wrong in common sense, and it would not be surprising if such rules were commonly shared by people.
However, for people who do not know what rain is or what an umbrella is, this alone is not enough to explain.
Similarly, just because there is this rule, when W becomes W': It is good to get wet., the consequence of the rule should be C': You should not take an umbrella. cannot be explained.
This does not mean that the argument consisting of the three sentences D, C, and W can be understood and explained.}
{This is the existence of the rule (wisdom of life). What is stated in $L_0$ is not wrong in common sense, and it would not be surprising if such rules were commonly shared by people.
However, for people who do not know what rain is or what an umbrella is, this alone is not enough to explain.
Similarly, the mere existence of this rule does not explain why, when W becomes W': ``It is good to get wet.'', the consequence of the rule should be C': ``You should not take an umbrella.''.
So, only by these rules, the argumentation consisting of the three sentences D, C, and W can not be completely understood or explained.
}
{
What is stated in $L_0$ is not wrong in common sense, and it would not be surprising if such rules were commonly shared by people. However, for people who do not know what rain is or what an umbrella is, this alone is not an enough explanation. Similarly, the mere existence of this rule does not explain why, when W becomes W': ``It is good to get wet.'', the consequence of the rule should be C': ``You should not take an umbrella.''. So, only by these rules, the argument consisting of the three sentences D, C, and W can not be completely understood or explained.}

\xxparatran{なぜ$L_0$，あるいは$S_D \To C_C$を主張できるのかは，
以下の3つの知識$L_1, L_2, L_3$があればNATLにより説明（導出）できる．
$L_1$は人々の一般的な振る舞いに関する常識的知識，
$L_2$は雨というものについての常識的知識，
$L_3$は傘というものについての常識的知識である．}{
The reason why $L_0$ or $S_D \To C_C$ can be asserted can be explained (derived) by NATL with the following three pieces of knowledge $L_1, L_2, L_3$.
$L_1$ is common sense knowledge about people's general behavior,
$L_2$ is common knowledge about rain,
$L_3$ is common sense knowledge about umbrellas.
}{
The reason why $L_0$ or $S_D \To C_C$ can be claimed can be explained (derived) by NATL with the following three pieces of knowledge $L_1, L_2, L_3$.
$L_1$ is common sense knowledge about general behavior of people, $L_2$ is common knowledge about rain, and $L_3$ is common knowledge about umbrellas.
}{
NATL can explain the reason why $L_0$ or $S_D \To C_C$ can be claimed with the following three pieces of knowledge $L_1, L_2, L_3$.
$L_1$ is common sense knowledge about general behavior of people, $L_2$ is common knowledge about rain, and $L_3$ is common knowledge about umbrellas.
}
\begin{quote}
\begin{tabular}{ll}
$L_1$: & $(\term{causal\mh and}, x, bad) \To (\term{avoid}, \term{people}, x)$\\
$L_2$: & $(\tlf \to \term{raining}) \To  \term{getting\mh wet}$\\
$L_3$: & $(\term{have}, x, \term{umbrella}) \To (\term{avoid}, x, \term{getting\mh wet})$\\
\end{tabular}
\end{quote}

\xxparatran{このとき，$\{S_D, S_W, L_1, L_2, L_3\}$を前提として以下の推論を行える．}{
At this time, we can make the following inference based on $\{S_D, S_W, L_1, L_2, L_3\}$.
}{At this time, the following inference can be made on the premise of $\{S_D, S_W, L_1, L_2, L_3\}$.
}
{
Then, the following reasoning can be made on the premise of $\{S_D, S_W, L_1, L_2, L_3\}$.
}
\begin{quote}
\begin{tabular}{ll}
$\{S_D, L_2\} \vdash$ & $ B_1: \term{getting\mh wet}$\\
$\{S_W, B_1\} \vdash$ & $ B_2: \term{bad}$\\
$\{B_1, B_2\} \vdash$ & $ C_1: (\term{causal\mh and}, B_1, B_2)$\\
$\{C_1, L_1\} \vdash$ & $ C_2: (\term{avoid}, \term{people}, \term{getting\mh wet})$\\
$\{C_2, L_3\} \vdash$ & $ C_3: (\term{have}, people, \term{umbrella})$\\
\end{tabular}
\end{quote}

\xxparatran{この文脈では$C_3$は$C_C$と同義と見なせるだろう．
これを認めれば，$S_D$から$C_C$を説明できことになる．
ただし$\{B_1, B_2\}$ $\vdash$ $C_1$ のステップは，
ここまでに説明していないもので，
因果的な読みを伴う連言を導入している．}{
In this context $C_3$ could be considered synonymous with $C_C$. If we accept this, we can explain $S_D$ to $C_C$. However, the steps $\{B_1, B_2\}$ $\vdash$ $C_1$ have not been explained so far.
Conjunctions with causal readings are introduced.}
{In this context, $C_3$ can be considered synonymous with $C_C$.
If we accept this, we can explain $C_C$ to $S_D$.
However, the steps $\{B_1, B_2 \}$ $\vdash$ $C_1$ have not been explained so far.Conjunctions with causal readings are introduced.
}
{
In this context, $C_3$ can be considered synonymous with $C_C$.
If we accept this, we can explain $C_C$ from $S_D$.
However, the step $\{B_1, B_2 \}$ $\vdash$ $C_1$ has not been explained so far.
It introduced a conjunction with a causal reading.}

\xxparatran{$\{S_D, L_2\}$ $\vdash$ $B_1$のステップが順方向の演繹である一方で, $\{C_2, L_3\}$ $\vdash$ $C_3$ のステップは逆向きのアブダクションである．
このように順方向と逆方向の推論が混在しうることは，非公理的論理NALと共有するNATLの大きな特徴である．}
{The $ \{S_D, L_2 \} $ $ \vdash $ $ B_1 $ step is a forward deduction, while the $ \{C_2, L_3 \} $ $ \vdash $ $ C_3 $ step is a reverse deduction. It is an abduction.
The fact that forward and reverse reasoning can coexist in this way is a major feature of NATL shared with irrational logic NAL.}
{While the $\{S_D, L_2\}$ $\vdash$ $B_1$ step is a forward abduction, the $\{C_2, L_3\}$ $\vdash$ $C_3$ step is a reverse abduction.
This possibility of mixing forward and reverse inference is a major feature of NATL, shared with the non-axiomatic logic NAL.
}
{
While the step of $\{S_D, L_2\}$ $\vdash$ $B_1$ is a forward deduction, the $\{C_2, L_3\}$ $\vdash$ $C_3$ step is a backward abduction. 
The fact that forward and reverse reasoning can coexist in this way is a major feature of NATL shared with NAL.}

\xxparatran{この説明では，もとのC文にあった`should'の意味は（少なくとも見かけ上は）消えてしまっている．
自然言語から項表示言語への翻訳方法は本稿では詳細には議論しないが，
今後本提案の工学的な実用性を実証するためには無視できない課題の1つであり，\S\ref{sec:nl2natl}にて見通しを述べる．
また，規則$L_1, L_2, L_3$
もきれいに連鎖するようになっているが，実際のシステム上で，
知識データベース全体に渡って完全に整合した記号体系を運用することは困難であり，
その困難を克服できる仕組みが必要になる．
意味ベクトルに基づくソフトな単一化の仕組みだけでそれを克服できるのかを明らかにするには，
今後の研究を要する．}
{
In this explanation, the meaning of `should' in the original C statement has disappeared (at least in appearance). Although the method of translating from natural language to term display language is not discussed in detail in this paper, it is one of the issues that cannot be ignored in order to demonstrate the engineering practicality of this proposal in the future. \S\ref{sec:nl2natl} gives a perspective. Also, the rules $L_1, L_2, and L_3$ are chained neatly, but on the actual system,
It is difficult to operate a completely consistent symbology over the entire knowledge database, and a mechanism that can overcome this difficulty is required.
Future research is required to clarify whether the mechanism of soft unification based on semantic vectors alone can overcome this problem.
}
{In this explanation, the meaning of `should' in the original C sentence has disappeared (at least in appearance). 
The translation method from natural language to term display language is not discussed in detail in this paper, but it is one of the issues that cannot be ignored in order to demonstrate the engineering practicality of the proposal in the future, and the outlook is described in \S\ref{sec:nl2natl}.
Though the rules $L_1, L_2, L_3$ are also neatly chained, it is still difficult to operate a perfectly consistent symbol system throughout the knowledge database in an actual system, and a mechanism is needed to overcome this difficulty.
Future research is required to clarify whether the mechanism of soft unification based on semantic vectors alone can overcome this problem.
}
{
In this explanation, the meaning of `should' in sentence C has disappeared (at least in appearance). 
The translation method from natural language to TRL is not given in detail in this paper, but it is one of the issues that cannot be ignored in order to demonstrate the technical feasibility of the proposal in the future, and thus the outlook will discussed in \S\ref{sec:nl2natl}.
Though the rules $L_1, L_2, L_3$ are also neatly chained, it is difficult to operate a perfectly consistent symbol system throughout the knowledge database in an actual system, and a mechanism is needed to overcome this difficulty.
Future research is required to clarify whether the mechanism of soft unification based on semantic vectors alone can overcome this problem.}

\xxparatran{また，$L_1, L_2, L_3$
は$L_0$よりも深い（より基本的な）知識であるとはいえ，
例えば$L_2$について「雨がふると濡れるのはなぜか？」と更に問うことは可能で，
その意味で上記の説明で全ての事柄が完全に説明されているわけではない．
あらゆる因果関係知識については，いくらでもなぜと問いつづけることができるが，
多くの知識（例えば雨が水であり，水に触れれば濡れること）については，
人は直接的に世界とのインタラクションの中で学んでおり，
当然のこととして，なぜと問うことなく了解するようになるのだろう．
これは身体性の問題でもあり，認知以前の問題に踏み込むので，NATLの理論射程の外側の問題になる．}
{Also, even though $L_1, L_2, and L_3$ are deeper (more fundamental) knowledge than $L_0$, it is still possible to ask, for example, $L_2$, ``Why does it get wet when it rains?'' is possible, and in that sense the above description does not fully explain everything.
We can continue to ask why about all kinds of causal knowledge, but for many types of knowledge (for example, rain is water, and if you touch water, you get wet), people directly interact with the world. I wonder if they will come to understand without asking why as a matter of course.
This is also a matter of physicality, and since it goes beyond cognition, it falls outside the scope of NATL's theory.}
{Even though $L_1, L_2, and L_3$ are deeper (more fundamental) knowledge than $L_0$, it is possible to ask, for example, $L_2$, ``Why do we get wet when it rains?''. In this sense, the above explanation does not fully explain everything.
For all causal knowledge, we can keep asking why as much as we want, but for much knowledge (for example, the rain is water and we get wet when we touch the water), we learn directly in our interactions with the world, and naturally come to understand without asking why.
This is also a problem of physicality, which is outside the theoretical range of NATL because it goes into issues before cognition.
}
{
Even though $L_1, L_2,$ and $L_3$ are deeper (more fundamental) knowledge than $L_0$, it is possible to ask a further question, for example, ``Why do we get wet when it rains?'' 
In this sense, the above explanation does not fully explain everything.
For all causal knowledge, we can keep asking ``why'' as much as we want, but for much knowledge (for example, the rain is water and we get wet when we touch the water), we learn it directly in our interactions with the world, and naturally come to understand without asking why.
This is a problem of embodiment, which is outside the theoretical scope of NATL because it goes into issues before cognition.}

\subsection{Example 2: Is marijuana a gateway drug?}\label{sec:ex_Marijuana}

\xxparatran{2つ目の例は以下の3文（\cite{habernal-etal-2018-ARCT}，図1）からなる．}{
The second example consists of the following three sentences (\cite{habernal-etal-2018-ARCT}, Figure 1).
}
{
The second example consists of the following three sentences (\cite{habernal-etal-2018-ARCT}, Figure 1).
}
{
The second example consists of the following three sentences (\cite{habernal-etal-2018-ARCT}, Figure 1).
}
That is, 
D: Milk isn't a gateway drug even though most people drink it as children.
C: Marijuana is not a gateway drug.
W: Milk is similar to marijuana.

\xxparatran{項表示言語には以下の$C_D, C_W, S_C$のように翻訳できる．（簡単のためD中のitの照応は解決済みとする．``as children''についても略す．）
ここで，継承否定の繋辞$\notto$と，類似の繋辞$\rightsquigarrow$ を導入している．}{
It can be translated into the term display language as follows: $C_D, C_W, S_C$. (For simplicity, it is assumed that the anaphora of it in D has already been resolved. We also omit ``as children''.) Here, the not-inheritance suffix $\notto$ and the analogous suffix $\rightsquigarrow$ are have introduced.
}
{
It can be translated into the term representation language as follows: $C_D, C_W, S_C$. (For simplicity, we assume that the coreference of it in D has been resolved. The``as children'' is also abbreviated.)
Here, the inheritance negation copula $\notto$ and the similar copula $\rightsquigarrow$ are introduced.
}
{
It can be translated into TRL as $C_D, C_W, S_C$. (For simplicity, we assume that the anaphora of `it' in D has been resolved. We also omit ``as children''.)
Here, the inheritance negation copula $\notto$ and the similarity copula $\rightsquigarrow$ are introduced.
}
\begin{quote}
\begin{tabular}{ll}
$C_D$:& ($\term{even\mh though}, C_1: (\term{drink}, \term{most\mh people}, \term{milk}), S_1: (\term{milk} \notto \term{gateway\mh drug}))$\\
$S_W$:& $\term{milk} \rightsquigarrow \term{marijuana}$ \\
$S_C$:& $\term{marijuana} \notto \term{gateway\mh drug}$ \\
\end{tabular}
\end{quote}

\xxparatran{D文は連関項として表現することも考えられるが，
条件規則的ではないのでここでは複合項として扱う．
このような曖昧性をどのように考えるべきかについては，
工学的というよりは，認知科学的観点で，
一貫性のある基準を明確化できることが望まれる．}
{D-sentences can be expressed as associated terms, but since they are not conditional, they are treated as compound terms here. Regarding how to think about such ambiguity, it is desirable to clarify a consistent standard from a cognitive scientific point of view rather than from an engineering point of view.
}
{The D statement could be expressed as a linkage term, but since it is not a conditional rule, it is treated as a compound term.
Regarding how to think about such ambiguity, it is desirable to clarify a consistent standard from a cognitive scientific point of view rather than from an engineering point of view.
}
{
Sentence D could be expressed as a linkage term, but since it is not a conditional rule, it is treated as a compound term in the present analysis.
Regarding how to think about such ambiguity, it is desirable to clarify a consistent standard from a cognitive science point of view rather than from an engineering point of view.}

\xxparatran{本題に戻ると，この議論の主目的は，$C_D$，特に$C_D$中の$S_1$から，$S_C$を導くことである．
これ自体は，$\mathcal{S}\cdot\mathcal{S}\to\mathcal{S}$型推論1ステップ，すなわち
$\{S_1, S_W\} \vdash S_C$
で達成できてしまう．
ただし，W文は W': Marijuana is similar to milk. ではないので，
字面通りに受け取れば，正統な三段論法は適用できないことに注意されたい（$S_W$の向きが逆）．この論理を許容するには，NATLの枠組みが必要である．（NATLに基づく議論サポートAIシステムが実現されれば，この議論を受け入れた上で，W'のように論述するほうが良いということを図式的に可視化した説明とともにユーザに指摘できるだろう）．}
{Coming back to the topic, the main purpose of this discussion is to derive $S_C$ from $C_D$, specifically $S_1$ in $C_D$. This itself can be achieved by $\mathcal{S}\cdot\mathcal{S}\to\mathcal{S}$ type inference one step, namely $\{S_1, S_W\} \vdash S_C$. Note, however, that the W sentence is not W': Marijuana is similar to milk. If taken literally, the orthodox syllogism cannot be applied ($S_W$ is in the opposite direction). A NATL framework is necessary to allow this logic. (If a discussion support AI system based on NATL is realized, it will be possible to accept this discussion and point out to the user that it is better to argue like W', along with a graphically visualized explanation.)
}
{Coming back to the topic, the main purpose of this argumentation is to derive $S_C$ from $C_D$, specifically $S_1$ in $C_D$. 
This itself can be achieved with one step of $\mathcal{S}\cdot\mathcal{S}\to\mathcal{S}$ type inference, namely, $\{S_1, S_W\} \vdash S_C$.
However, that the W statement is not W': ``Marijuana is similar to milk.''. If taken literally, the orthodox syllogism cannot be applied (the direction of $S_W$ is reversed).
A NATL framework is necessary to allow this logic. 
(If an argumentation supporting AI systems based on NATL is realized, it will be possible to accept this argumentation and then point out to the user with a diagrammatic visualization and explanation that it is better to argue as W').
}
{
Coming back to the topic, the main purpose of this argument is to derive $S_C$ from $C_D$, specifically $S_1$ in $C_D$. 
This itself can be achieved with one step of $\mathcal{S}\cdot\mathcal{S}\to\mathcal{S}$ type inference, namely, $\{S_1, S_W\} \vdash S_C$.
However, as sentence W is not W': ``Marijuana is similar to milk.'', if taken literally, the orthodox syllogism cannot be applied (the direction of $S_W$ is reversed).
A NATL framework is necessary to allow this logic. 
(If an argumentation supporting AI systems based on NATL is realized, it will be possible to accept this argumentation and then point out to the user with a diagrammatic visualization and explanation that it is better to argue as W').}

\xxparatran{ところで$\{S_1, S_W\} \vdash S_C$だけで結論が導けるのであれば，$C_1$の役割は何であろうか？
$C_1$の役割は$S_1$の主張を論理的に直接補強することではなく，主張者に高い推論能力があることを示すことで，
主張者の権威を高め，それにより間接的に議論の正当性を高めるという，社会的なものと考えられる．
つまりA even though Bという論述のパターンは，
Toulminが示した(2)社会的立場や(3)技術的立場としての論理に強く関係すると考えられる．
いわゆるレトリック\cite{Nouchi02}である．}
{By the way, if only $\{S_1, S_W\} \vdash S_C$ can lead to a conclusion, what is the role of $C_1$? The role of $C_1$ is not to directly reinforce the claim of $S_1$, but to increase the authority of the claimant by showing that the claimant has high reasoning ability, thereby indirectly supporting the argument. It is considered to be a social thing that enhances legitimacy. In other words, the argument pattern of A even though B is strongly related to Toulmin's (2) social standpoint and (3) logic as a technological standpoint. This is the so-called rhetoric \cite{Nouchi02}.}
{By the way, if $\{S_1, S_W\} \vdash S_C$ can draw the conclusion, what is the role of $C_1$?
The role of $C_1$ is considered to be a social thing, not to directly reinforce $S_1$'s argument logically, but to enhance the authority of the advocate by showing that the advocate has high reasoning ability, thereby indirectly enhancing the legitimacy of the argumentation.
In other words, the argumentation pattern of ``A even though B'' is strongly related to the logic of (2) the social position and (3) the technical position shown by Toulmin.
This is the so-called rhetorical\cite{Nouchi02}.
}
{
By the way, if $\{S_1, S_W\} \vdash S_C$ can draw the conclusion, what is the role of $C_1$?
The role of $C_1$ is considered to be a social thing, not to directly reinforce $S_1$'s argument logically, but to enhance the authority of the advocate by showing that the advocate has high reasoning ability, thereby indirectly enhancing the legitimacy of the argument.
In other words, the argumentation pattern of ``A even though B'' is strongly related to the logic at the sociological position and the technological position shown by Toulmin.
}

\xxparatran{社会的相互作用の場面において人が無意識のうちに自己の有能性を主張する発言を行うことが観察されており\cite{KomuroFunakoshi22}，
$C_1$はこのような相互行為実践の現れと考えられる．
このことは，議論的推論能力である人間の理性（reason）が第一に社会的能力であるとする\citet{Mercier2017}の主張とつながる．}
{It has been observed that people unconsciously make statements that assert their competence in social interaction situations.
$C_1$ is considered to be a manifestation of such interaction practice.
This leads to \citet{Mercier2017}'s claim that human reason, which is the ability to make argumentative reasoning, is primarily a social ability.}
{It has been observed that people unconsciously make statements that assert their competence in social interaction situations \cite{KomuroFunakoshi22}, and $C_1$ is considered to be a manifestation of such interaction practice. This leads to \citet{Mercier2017}'s claim that human reason, which is the ability to make argumentative reasoning, is primarily a social ability.
}
{
It has been observed that people unconsciously make statements that assert their competence in social interaction situations \cite{KomuroFunakoshi22}.
$C_1$ is considered to be a manifestation of such interaction practice.
This leads to \citet{Mercier2017}'s claim that human reason, which is the ability to make argumentative reasoning, is primarily a social ability.}

\xxparatran{$C_1$の役割を引き続きNATLを用いた分析を通じて考えてみる．
`even though most people drink it as children' という但し書き的な言及の背後には，
以下のような，中毒性という性質と入口薬物というカテゴリについての2つの常識的知識があると想定できる．
つまり，$L_1$「中毒性があるものは人気がある」と$L_2$「中毒性があるものは入り口薬物になる」．}
{We continue to consider the role of $C_1$ through analysis using NATL.
Behind the provisory reference to `even though most people drink it as children', we can assume that there are two common sense knowledges about the nature of addiction and the category of entry drug: In other words, $L_1$ ``Those that are addictive are popular'' and $L_2$ ``Those that are addictive are the entry point drugs''.}
{Let us continue to consider the role of $C_1$ through analysis using NATL.
Behind the proviso reference ``even though most people drink it as children'', it can be assumed that there are two common-sense knowledge about the addictive nature and the category of gateway drugs.
That is, $L_1$ "Addictive ones are popular" and $L_2$ ``Addictive ones are gateway drugs''.
}
{
Let us continue to consider the role of $C_1$ through analysis using NATL.
Behind the provisory reference ``even though most people drink it as children'', it can be assumed that there are two common sense knowledge pieces about the addictive nature and the category of gateway drugs.
That is, ``Addictive ones are popular'' ($L_1$) and  ``Addictive ones are gateway drugs'' ($L_2$).
}

\begin{quote}
\begin{tabular}{ll}
$L_1$:& $(x \to \term{addictive}) \To (x \to \term{popular})$ \\
$L_2$:& $(x \to \term{addictive}) \To (x \to \term{gateway\mh drug}) $ \\
\end{tabular}
\end{quote}

\xxparatran{このとき，$C_1$の含意として，$C_1':$ $\term{milk} \to \term{popular}$ が了解されると，
以下のように，$S_4$という$S_1$と矛盾する結論が導かれる．
つまり$C_1\ (C_1')$は，$S_1$および$S_C$を支持するのではなく否定する．}
{Then, if $C_1':$ $\term{milk} \to \term{popular}$ is understood as an implication of $C_1$, $S_4$ contradicts $S_1$ as follows. A conclusion is drawn.
So $C_1\ (C_1')$ negates rather than supports $S_1$ and $S_C$.}
{Then, if $C_1':$ $\term{milk} \to \term{popular}$ is understood as an implication of $C_1$, $S_4$ contradicts $S_1$ as follows. A conclusion is drawn as follows. That is, $C_1\ (C_1')$ negates $S_1$ and $S_C$ rather than supporting them.}
{
Then, if $C_1':$ $\term{milk} \to \term{popular}$ is understood as an implication of $C_1$, $S_4$ contradicts $S_1$ as follows. A conclusion is drawn as follows.
That is, $C_1\ (C_1')$ negates $S_1$ and $S_C$ rather than supporting them.}

\begin{quote}
\begin{tabular}{ll}
$\{C_1', L_1\} \vdash$ & $S_3: \term{milk} \to \term{addictive}$\\
 $\{S_3, L_2\} \vdash$ & $S_4: \term{milk} \to \term{gateway\mh drug}$\\
\end{tabular}
\end{quote}

\xxparatran{しかしなぜ自らの主張を否定する情報を提示するのか？
その答えが，前述の社会的相互行為実践である．
自らの論理の弱点を予め自ら示唆することで，自らに注意深い推論の能力があることを示し，
それにより（自らの論理そのものの妥当性ではなく）論者の権威を高めようとする行為である．もちろん以上の議論は，$L_1, L_2$ が主張者の脳内に表示されていたと仮定した上でのもので憶測の域を出ないが，
NATLの形式理論の枠組の中で論述の構成要素の関係性を明晰に分析できることは示された．}
{But why present information that refutes their claims? The answer is the aforementioned social interaction practice. By suggesting the weaknesses of one's own logic in advance, it is an act of showing one's ability to make careful inferences, thereby increasing the author's authority (rather than the validity of one's own logic itself). Of course, the above discussion is based on the assumption that $L_1, L_2$ were displayed in the claimant's brain, and is speculative.
It was shown that the relationship between the constituent elements of discourse can be analyzed clearly within the framework of formal theory of NATL.
}
{But why present information that denies one's own claims?
The answer is the aforementioned social interaction practice.
By suggesting the weaknesses of one's logic in advance, it is an act of showing that one has the ability of careful reasoning, thereby increasing the authority of the theorist (rather than the validity of one's logic itself).
Of course, the above discussion is speculative based on the assumption that $L_1, L_2$ were displayed in the advocate's brain, but it has shown that the relationship between the components of the argumentation can be analyzed clearly within the framework of the formal theory of NATL.
}
{
But why does the arguer present the information that denies one's own claim?
The answer is the aforementioned social interaction practice.
By suggesting the weaknesses of one's logic in advance, it is an act of showing that one has the ability of careful reasoning, thereby increasing the authority of the arguer (rather than the validity of one's logic itself).
Of course, the above discussion is speculative based on the assumption that $L_1, L_2$ were represented in the arguer's brain, but it has shown that the relationship between the components of the argument can be analyzed clearly within the framework of the formal theory of NATL.}

\subsection{Example 3: Is Google a Harmful Monopoly?}\label{sec:ex_Google}
\xxparatran{
3つ目の例は以下の3文（\cite{niven-kao-2019-probing}，
図1）\footnote{自然さのため元の議論の主張を反転したものを用いている．}
からなる．
D: People cannot choose not to use Google.
C: Google is a harmful monopoly.
W: Other search engine redirect to Google.
これらは以下の$C_D, C_W, S_C$として表示できる．}{
The third example uses the following three sentences (\cite {niven-kao-2019-probing}, Fig. 1) \footnote {inverted from the original argument for naturalness.}.
D: People cannot choose not to use Google.
C: Google is a harmful monopoly.
W: Other search engine redirect to Google.
These can be displayed as the following $ C_D, C_W, S_C $.}
{
The third example uses the following three sentences (\cite {niven-kao-2019-probing}, Fig. 1) \footnote {For the sake of naturalness, we have inverted the claims of the original argument.}.
D: People cannot choose not to use Google.
C: Google is a harmful monopoly.
W: Other search engines redirect to Google.
These can be displayed as the following $ C_D, C_W, S_C $.
}
{
The third example uses the following three sentences (\cite {niven-kao-2019-probing}, Fig.\,1). \footnote {For the sake of naturalness, we have inverted the claims of the original argument.}
D: People cannot choose not to use Google.
C: Google is a harmful monopoly.
W: Other search engines redirect to Google.
These can be expressed as $ C_D, C_W, S_C$:
}
\begin{quote}
\begin{tabular}{ll}
$C_D$:& $(\term{cannot\mh choose\mh not\mh to\mh use}, \term{people}, \term{Google})$ \\
$C_W$:& $(\term{redirect\mh to}, \term{other\mh search\mh engines}, \term{Google})$\\ 
$S_C$:& $\term{Google} \to \term{harmful\mh monopoly}$\\
\end{tabular}
\end{quote}

\xxparatran{この例では，WはもともとのTouleminモデルでいうところのwarrantではない．
これまで見た2例と異なり，この例でのWはDからCを導く推論経路の中には含まれない．
Wの役割は，Dからスタートする論証の確立ではなく，そもそもの出発点であるDの妥当性を担保することである．}
{In this example, W is not a warrant in the original Toulemin model. Unlike the two examples we have seen so far, W in this example is not included in the inference path leading from D to C. The role of W is not to establish an argument starting from D, but to ensure the validity of D, which is the starting point in the first place.}
{In this example, W is not the warrant in the original Toulemin model.
Unlike the two examples we have seen so far, W in this example is not included in the inference path leading from D to C.
The role of W is not to establish an argument starting from D, but to ensure the validity of D, which is the starting point in the first place.
}
{
In this example, W is not the warrant in the original Toulmin model.
Unlike the two examples we have seen so far, W in this example is not included in the reasoning path leading from D to C.
The role of W is not to establish an argument starting from D, but to ensure the validity of D, which is the starting point in the first place.
}

\xxparatran{まず先にDからCを導くのに必要な真のwarrantを考えると，それは$L_0$のような社会の中で身を守るための知識だろう．
すなわち$L_0$:「もし$x$が$y$の使用を避けられないなら，$y$は$x$にとって危険である」．}
{First of all, considering the true warrant required to derive C from D, it would be knowledge to protect ourselves in a society like $ L_0 $.
That is, $ L_0 $: "If $ x $ cannot avoid using $ y $, then $ y $ is dangerous for $ x $".}
{First of all, considering the true warrant required to derive C from D first, it would be the knowledge to protect oneself in a society like $L_0$.
That is, $L_0$: ``If $x$ cannot avoid using $y$, then $y$ is dangerous for $x$.''.
}
{
First of all, considering the true warrant required to derive C from D first, it would be the knowledge to protect oneself in a society like $L_0$.
That is, $L_0$: ``If $x$ cannot avoid using $y$, then $y$ is dangerous for $x$.''.
}
\begin{quote}
\begin{tabular}{ll}
$L_0$:& $(\term{cannot\mh choose\mh not\mh to\mh use}, x, y) \To (\term{is\mh harmful\mh for}, y, x)$
\end{tabular}
\end{quote}

\xxparatran{$x$が$y$を使わなければいけないとしても$y$が$x$にとって有害になる物理的な必然性はないため，無垢な人間はこのような状況に対して無頓着で危機感を覚えない．人間はすぐにカテゴリ錯誤を起こすので\cite{Nass00}，物理的な道具での経験を非物理的な道具（サービス）にも容易に過適用する．しかし，選択の余地のない側の立場が弱くなり，引いてはただの道具であったはずの強い側の利益の犠牲にされることは社会的によくある不幸で，$L_0$のような知識を教授することはしばしば行われることだろう．
この議論の推論本体は， $C_D$と$L_0$から$(\term{is\mh harmful\mh for}$, $\term{Google}, \term{people})$ を導出し，これが項$S_C$と意味内容的にほぼ同義であることを認識すれば完了する．このようにWはDからCへの推論に直接関与しない．}
{Since there is no physical necessity for $y$ to be detrimental to $x$ even though $x$ must use $y$, innocent human beings are indifferent and alarmed by such situations. Humans are quick to make the category fallacy, so it's easy to over-apply experiences with physical tools to non-physical tools (services). However, it is a common misfortune in society that the position of the side with no choice is weakened, and the benefits of the strong side, which should have been just a tool, are sacrificed. Teaching knowledge will often be done. The body of reasoning for this argument derives from $C_D$ and $L_0$ $(\term{is\mh harmful\mh for}$, $\term{Google}, \term{people})$, which is the term If you recognize that it is semantically equivalent to $S_C$, you are done. Thus W does not directly participate in the inference from D to C.}
{Since there is no physical necessity for $y$ to be detrimental to $x$ even though $x$ must use $y$, innocent human beings are indifferent and alarmed by such situations.
Because humans easily make category errors, they may over-apply their experience with physical tools to non-physical tools (services).
However, it is a common social misfortune that the side with no choice is weaker, and in turn, is sacrificed to the interests of the stronger side which should have been just a tool. So, teaching knowledge such as $L_0$ should be done more often.
The main body of reasoning in this argumentation is completed by deriving $(\term{is\mh harmful\mh for}$, $\term{Google}, \term{people})$ from $C_D$ and $L_0$ while recognizing that it is almost synonymous with the term $S_C$ in semantic content.
}
{
Since there is no physical necessity for $y$ to be detrimental to $x$ even though $x$ must use $y$, innocent human beings are indifferent and not alarmed by such situations.
Because humans easily make category errors, they may over-apply their experience with physical tools to non-physical tools (services).
However, it is a common social misfortune that the side with no choice is weaker, and in turn, is sacrificed to the interests of the stronger side which should have been just a tool. So, teaching knowledge such as $L_0$ will often be done.
The body of reasoning for this argument is completed by deriving $(\term{is\mh harmful\mh for}$, $\term{Google}, \term{people})$ from $C_D$ and $L_0$ while recognizing that it is almost synonymous with the term $S_C$ in semantic content.
}

\xxparatran{引き続き，WとDの関係を見る．
この関係を理解するには以下の3つの認識・知識が必要と思われる．}
{We continue to look at the relationship between W and D. In order to understand this relationship, the following three recognitions and knowledge are necessary.
}
{Next, let's look at the relationship between W and D.
To understand this relationship, the following three perceptions or knowledge may be necessary.
}
{
Next, let's look at the relationship between W and D.
To understand this relationship, the following three knowledge pieces may be necessary.
}

\begin{quote}
\begin{tabular}{ll}
$L_1$:& $(\term{and}, (\term{use}, x, y), (\term{use}, y, z)) \To (\term{use}, x, z)$ \\
$S_1$:& $(\term{redirect\mh to}, x, y) \to (\term{use}, x, y)$\\
$C_1$:& $(\term{want}, \term{people}, (\term{use}, \term{people}, \term{other\mh search\mh engines\mh than\mh Google}))$ 
\end{tabular}
\end{quote}

\xxparatran{$L_1$は，useという行為について推移率が成立しうることを表す（推移律は動詞一般に成立するものではないので個別に知っていなければならない）．
$S_1$は，redirectという行為が，ある種のuseであることを表す（これはコンピュータネットワークに関する専門知識であろう）．
$C_1$は，Dの前提として意図認識されるべきことで，「人々がGoogle以外の検索エンジンを使おうとしている」ということである．}
{$L_1$ indicates that the transition rate can hold for the act of use (the transit law does not hold for verbs in general, so it must be known individually). $S_1$ indicates that the act of redirecting is a kind of use (this would be an expert knowledge of computer networks). $C_1$ is a premise of D that should be recognized as an intention, saying that ``people are trying to use search engines other than Google.''.
}
{$L_1$ indicates that the transitive rate can be established for the act of ``use'' (the transitive law is not established for verbs in general and must be known individually).
$S_1$ indicates that the act of ``redirect'' is some kind of ``use'' (this would be an expert knowledge of computer networks).
$C_1$ should be recognized as a premise of D, which means that ``People are trying to use search engines other than Google.''.
}
{
$L_1$ indicates that the transitive law can be established for the act of ``use'' (the transitive law is not established for verbs in general and must be known individually).
$S_1$ indicates that the act of ``redirect'' is some kind of ``use'' (this would be an expert knowledge of computer networks).
$C_1$ should be recognized as a premise of D, which means that ``People are trying to use search engines other than Google.''.
}

\xxparatran{我々はしばしばコミュニケーションの場において，「AということをしたいけどAをできない」ということを伝えたいときに，
「Aをできない」とだけ伝える．例えばあるドアを開けたいときに「このドア重たくて開かないんだよ」とだけいうような場合である．
一般的な聞き手はこの発言から発話者がドアを開けたいと思っていることを推察する．
これは一般に意図認識とよばれる問題の一種であるが，
「〇〇できない」と伝える行為は，
先程もふれた相互行為実践（「行為理解のための仕掛け」\cite{KomuroFunakoshi22}）
として捉えるのが適切だろう．
このような意図認識が人間において記号推論的に行われているのか，それとも記号処理能力以前から別の形で備わっている能力によって
記号推論と並列的に処理されているのかはわからない（多くの場合で意図認識は，周囲の状況情報に広く依存するマルチモーダルな問題なので，意図認識自体は記号的推論とは別の仕組みである可能性が高そうではある）が，
ここではひとまず$C_D$から$C_1$が認識されたとする．
ここで，$S_1$と$C_W$から，$C_2: (\term{use}, \term{other\mh search\mh engine}, \term{Google})$が了解される．
すると，$C_1$と$C_2$が$L_1$の左項と（ソフトに）単一化されることで，
$L_1$より$C_3: (\term{use}, \term{people}, \term{Google})$が帰結される．
$C_W$を用いた推論の帰結$C_3$が当初の主張である$C_D$と整合するため，WはD文の蓋然性を高める効果を持つことになる．}
{When we want to say, ``I want to do A, but I can't do A,'' we often say, ``I can't do A.'' For example, when you want to open a certain door, you just say, "This door is too heavy to open." A typical listener will infer from this utterance that the speaker wants to open the door. This is a kind of problem generally called intention recognition.
It would be appropriate to think of the act of telling someone that ``I can't do XX'' as the practice of mutual action that I mentioned earlier (``a mechanism for understanding behavior''\cite{KomuroFunakoshi22}). It is not known whether such intent recognition is performed in humans by means of symbolic reasoning, or whether it is processed in parallel with symbolic reasoning by another form of ability that has existed before the ability to process symbols (in many cases, Intention recognition is a multimodal problem that widely depends on surrounding situational information, so it seems likely that intention recognition itself is a different mechanism from symbolic reasoning), but here we will start from $C_D$ Suppose $C_1$ is recognized. Here $C_2: (\term{use}, \term{other\mh search\mh engine}, \term{Google})$ is understood from $S_1$ and $C_W$. Then, by (softly) unifying $C_1$ and $C_2$ with the left term of $L_1$, $C_3: (\term{use}, \term{people}, \ term{Google})$ results. Since the result $C_3$ of the inference using $C_W$ is consistent with the original assertion $C_D$, W has the effect of increasing the probability of the D sentence.}
{Often in communication, when we want to say, ``I want to do A, but I can't do A.'', we simply say, ``I can't do A.''. For example, when we want to open a door but we can't open it, we simply say, ``This door is too heavy to open.''.
The typical listener infers from this statement that the speaker wants to open the door.
This is a kind of problem generally referred to as intention recognition, and the act of telling the speaker ``I can't do it'' can be appropriately viewed as an interaction practice(``Mechanism for understanding actions'' \cite {KomuroFunakoshi22}), as mentioned earlier.
It is not clear whether such intention recognition is performed by symbolic inference in humans, or whether it is processed in parallel with symbolic inference by the ability that has been provided in another form before the symbol processing ability(In many cases, intention recognition is a multi-modal problem that relies extensively on information about the surrounding situation, so it seems likely that intention recognition itself is a distinct mechanism from symbolic inference.), but let us assume that $C_1$ is recognized from $C_D$.
Here, $C_2: (\term{use}, \term{other\mh search\mh engine}, \term{Google})$ is understood from $S_1$ and $C_W$.
Then, $C_1$ and $C_2$ are (softly) unified with the left term of $L_1$, resulting in $C_3: (\term{use}, \term{people}, \term{Google})$ from $L_1$.
Since the result $C_3$ of the inference using $C_W$ is consistent with the original assertion $C_D$, W has the effect of increasing the probability of the D statement.
}
{
Often in communication, when we want to say, ``I want to do A, but I can't do A.'', we simply say, ``I can't do A.''. For example, when we want to open a door but we can't open it, we simply say, ``This door is too heavy to open.''
The typical listener infers from this statement that the speaker wants to open the door.
This is a kind of problem generally referred to as intention recognition, and the act of telling the speaker ``I can't do it'' can be appropriately viewed as an interaction practice (``mechanism for understanding others' actions'' \cite{KomuroFunakoshi22}), as mentioned earlier.
It is not clear whether, in humans, such intention recognition is performed by symbolic reasoning, or whether it is processed in parallel with symbolic reasoning by the ability that has been provided in another way (In many cases, intention recognition is a multi-modal problem that relies extensively on information about the surrounding situation, so it seems likely that intention recognition itself is a distinct mechanism from symbolic reasoning.), but let us assume that $C_1$ is recognized from $C_D$ anyway.
Here, $C_2: (\term{use}, \term{other\mh search\mh engines}, \term{Google})$ is understood from $S_1$ and $C_W$.
Then, $C_1$ and $C_2$ are (softly) unified with the left term of $L_1$, resulting in $C_3: (\term{use}, \term{people}, \term{Google})$ from $L_1$.
Since the result $C_3$ of the inference using $C_W$ is consistent with the original assertion $C_D$, W comes to have the effect of increasing the probability of statement D.
}

\section{Other Applications}\label{sec:applications}

\xxparatran
{\S\ref{sec:argumentation}では，NATLが議論の背景にある推論の構造を明らかにできることを見た．
この過程を計算機実装できれば，議論を支援するAIの実現などに貢献できる．
本章では応用・発展として考えられる方向性を3つ示す．}
{In \S\ref{sec:argumentation}, we saw that NATL can reveal the structure of reasoning behind arguments.
If this process can be implemented on a computer, it will contribute to the realization of AI that supports discussions.This chapter presents three possible directions for application and development.}
{In \S\ref{sec:argumentation}, we saw that NATL can reveal the structure of inference behind arguments.
If this process can be implemented on a computer, it will contribute to the realization of artificial intelligence that supports discussions.
This chapter presents three possible directions for application and development.
}
{
In \S\ref{sec:argumentation}, we saw that NATL can reveal the structure of reasoning behind arguments.
If this process can be implemented on a computer, it will contribute to the realization of artificial intelligence that supports discussions.
This chapter presents three possible directions for other applications.}

\subsection{Solving math problems heuristically}\label{sec:math_problem}

\xxparatran
{算数の問題を解く過程は，
（少なくとも同型の問題を反復することで習慣化・自動化されるまでは）
意識的な記号推論に依存する部分が大きいと思われる．
もちろん幾何的な問題を解くには空間認識能力なども不可欠であるが，
それらの能力とNATLによる表示・推論能力の組み合わせで，
人間と同じように算数の問題を解く（そして人間と同じ誤り方を再現する）
AIを実現できる可能性がある．
ここでは初歩的な算数の問題の解答を得るまでの過程の説明を試みる}{
The process of solving arithmetic problems (At least until it becomes habitualized/automated by repeating the same type of problem) seems to depend largely on conscious symbolic reasoning.
Of course, spatial recognition ability is also essential for solving geometric problems, but combining these abilities with the display and reasoning capabilities of NATL may lead to the creation of an AI that can solve arithmetic problems in the same way as humans do (And reproduce the same mistakes as humans).
Here, I will try to explain the process of getting answers to elementary math problems.}
{The process of solving arithmetic problems (At least until it becomes habitualized by repeating the same type of problem) seems to depend largely on conscious symbolic reasoning. 
Of course, spatial recognition ability is also essential for solving geometric problems, but combining these abilities with the display and reasoning capabilities of NATL may lead to the creation of an AI that can solve arithmetic problems in the same way as humans do (and reproduce the same mistakes as humans).
Here, we will try to explain the process of getting answers to elementary math problems.
}
{
The process of solving math problems (at least until it becomes habitualized and automated by repeating the same type of problem) seems to depend largely on conscious symbolic reasoning. 
Of course, spatial recognition ability is also essential for solving geometric problems, but combining these abilities with the representing and reasoning capabilities of NATL may lead to the creation of an AI that can solve arithmetic problems in the same way as humans do (and reproduce the same mistakes as humans).
Here, we will try to explain the process of getting answers to elementary math problems.}

\begin{figure}[t]
    \centering
    \includegraphics[width=.8\linewidth]{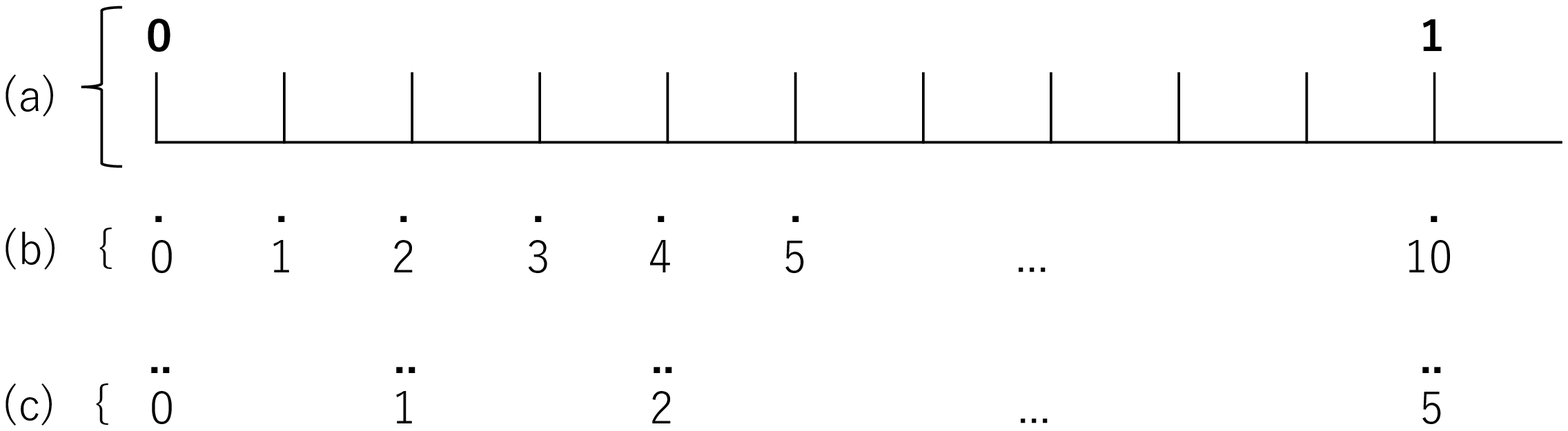}
    \caption{Number line and terms}
    \label{fig:number-line}
\end{figure}

\xxparatran
{小学生向けの算数の実例として，「図の数直線を使って，$\square$に当てはまる数を考えよ」という問題を取り上げる．
「図の数直線」は，\ref{fig:number-line}の最上部(a)に示したものであり，
$\square$とは次の等式 $\dfrac{\mathit{4}}{\mathit{10}}=\dfrac{\square}{\mathit{5}}$ 中のものである．
ここで小学生に求められていることは，数直線を手がかりに２つの数列の間の対応を
発見的に見出すことで分子と分母の関係について幾何的な理解を深めることであって，
予め教えられた代数的な記号操作を機械的に適用して，つまり$\square = 5\times4\div10$という算術演算を行って解を得ることではない%
\footnote{代数的な記号操作による演算もNATLで扱えるべきであるし，
汎用性を指向する深層学習モデルが不得手とする重要な問題\cite{Fujisawa21}であるが，
これについての検討は別稿に譲る．
}．ここではこの問題を与えられた想像上の小学生が答えを導くまでの過程を，NATLを用いて追う．}
{As an actual example of arithmetic for elementary school students, we take up the problem "Using the number line in the figure, think of a number that fits $\square$."
The "figure number line" is shown at the top of \ref{fig:number-line}, (a), where $\square$ is defined by the following equation $\dfrac{\mathit{4}} {\mathit{10}}=\dfrac{\square}{\mathit{5}}$.
Elementary school students are expected to deepen their geometric understanding of the relationship between the numerator and the denominator by finding the correspondence between two number sequences heuristically using the number line as a clue. It is not to mechanically apply some algebraic symbolic manipulations, i.e. perform the arithmetic operation $\square = 5\times4\div10$ to get the solution%
\footnote{NATL should be able to handle operations using algebraic symbolic manipulations, and this is an important problem \cite{Fujisawa21} that general-purpose deep learning models are not good at, but I will leave the discussion of this to another article.}
Here, we use NATL to follow the process of an imaginary elementary school student given this problem until he or she derives the answer.}
{As an actual example of arithmetic for elementary school students, we take up the problem "Using the number line in the figure, think of a number that fits $\square$." 
The number axis is shown at the top of \ref{fig:number-line}, (a), where $\square$ is defined by the following equation $\dfrac{\mathit{4}} {\mathit{10}}=\dfrac{\square}{\mathit{5}}$.
Elementary school students are expected to deepen their geometric understanding of the relationship between the numerator and the denominator by finding the correspondence between two number sequences heuristically using the number line as a clue.
It is not to mechanically apply the algebraic symbol operation teaching in advance, that is, to perform $\square = 5\times4\div10$ arithmetic operation to obtain the solution. %
\footnote{NATL should be able to handle operations using algebraic symbolic manipulations, and this is an important problem \cite{Fujisawa21} that general-purpose deep learning models are not good at, but we will leave the discussion of this to another article.} 
Here, we use NATL to follow the process of an imaginary elementary school student given this problem until he or she derives the answer.
}
{
As an actual example of basic math for elementary school students, 
we take up the problem: ``Using the number line in the figure, think of a number that fits $\square$.'' 
The number axis is shown at the top of Fig.~\ref{fig:number-line}, (a), 
and $\square$ is defined by the following equation $\dfrac{\mathit{4}}{\mathit{10}}=\dfrac{\square}{\mathit{5}}$.
Elementary school students are expected to deepen their geometric understanding of the relationship between the numerator and the denominator by finding the correspondence between two number sequences heuristically using the number line as a clue.
It is not to mechanically apply the algebraic symbol operation taught in advance, that is, to perform the arithmetic operation of $\square = 5\times4\div10$ to obtain the solution.%
\footnote{
NATL should be able to handle operations using algebraic symbolic manipulations, and this is an important problem \cite{Fujisawa21} that general-purpose deep learning models are not good at, but we will leave the discussion of this to another article.
} 
Here, we use NATL to follow the process of an imaginary elementary school student given this problem until he or she derives the answer.
}

\xxparatran
{まず，小学生は問題の指示により\ref{fig:number-line}~(a)の数直線に注目・観察し，0と添え書きされた
棒から1と添え書きされた棒までの間に，それらを含めて11の棒（つまり目盛り）があることに気づく．
（この計数行為は，幼少からの遊びを通じて身についた習慣的行為として，条件反射的に駆動されるとここでは考える．）
ここで，
小学生は既に定規の使い方を知っており，その経験とのアナロジーと，問題の式の左辺に$10$という数があることから，
数直線の下に$0$から$10$の数を並べてみることを発想する．
この配置の結果が，\ref{fig:number-line}の(b)であり，これらはNATLが扱う基本項となる．
配置された項はそれぞれ対応する数概念とともに固有の空間的位置を伴うトークン記号であるので，
$\dot{0}, \ldots, \dot{10}$のように付点することでそのことを表す．}{
First, elementary school students pay attention to and observe the number line \ref{fig:number-line}~(a) according to the instructions of the problem, and move them between the bar annotated with 0 and the bar annotated with 1. Notice that there are 11 bars (or tick marks) inclusive.
(Here, we think that this counting act is driven by a conditional reflex as a habitual act acquired through play from childhood.)
Elementary school students already know how to use a ruler, and by analogy with their experience and the fact that the number $10$ is on the left side of the equation in question, it is recommended to arrange the numbers from $0$ to $10$ under the number line. come up with an idea
The result of this arrangement is (b) in \ref{fig:number-line}, which are the elementary terms handled by NATL.
Since each placed term is a token symbol with a unique spatial position along with the corresponding number concept, we can indicate that by noting $\dot{0}, \ldots, \dot{10}$. represents.}
{
First, elementary school students pay attention to and observe the number line \ref{fig:number-line}~(a) according to the instructions of the problem, and move them between the bar annotated with 0 and the bar annotated with 1. Notice that there are 11 bars (or tick marks) inclusive.
(It can be considered that this counting behavior is driven by a conditional reflex as a habitual act acquired through play from childhood.) 
Elementary school students already know how to use a ruler, and by analogy with their experience and the fact that the number $10$ is on the left side of the equation in question, it is recommended to arrange the numbers from $0$ to $10$ under the number line.
The result of the arrangement is the \ref{fig:number-line} (b),  which are the elementary terms handled by NATL. Since each placed term is a token symbol with a unique spatial position along with the corresponding number concept, we can indicate that by noting $\dot{0}, \ldots, \dot{10}$.
}
{
First, the pupil pays attention to and observe the number line Fig.~\ref{fig:number-line}~(a) according to the instructions of the problem, and 
notices that there are 11 bars (or tick marks) between the bar annotated with 0 and the bar annotated with 1. (It can be considered that this counting behavior is driven by a conditional reflex as a habitual act acquired through play from childhood.) 
The pupil already knows how to use a ruler, and by analogy with their experience and the fact that the number $10$ is on the left side of the equation in question, the pupil comes up with an idea to arrange the numbers from $0$ to $10$ under the number line.
The arrangement is shown as Fig. \ref{fig:number-line} (b), which are the basic terms handled by NATL. Since each placed term is a token symbol with a unique spatial position along with the corresponding number concept, we indicate that by noting $\dot{0}, \ldots, \dot{10}$.}

\xxparatran
{一旦等式に戻り，等式左辺でグループを作っている数$\mathit{4}$と$\mathit{10}$に注目して
（この問題では，等号や分数という概念自体は既に教わっていることが前提），先程並べた基本項と結びつける．
すなわち，$S_1: \mathit{4} \leftrightarrow \dot{4}$, $S_2: \mathit{10} \leftrightarrow \dot{10}$．
（繋辞$\leftrightarrow$は，トークンとしては異なるが，タイプとしては同一であるという認識を表す．）}
{Returning to the equation, focus on the numbers $\mathit{4}$ and $\mathit{10}$ (this question assumes that the concepts of equals and fractions have already been taught) that form a group on the left side of the equation, and connect them with the elementary terms listed earlier.
That is, $S_1: \mathit{4} \leftrightarrow \dot{4}$, $S_2: \mathit{10} \leftrightarrow \dot{10}$. 
(The copula $\leftrightarrow$ denotes the recognition that the tokens are different but the types are the same.)}
{Returning to the equation, focus on the numbers $\mathit{4}$ and $\mathit{10}$ that form a group on the left side of the equation (for this problem, the concepts of equal sign and fraction have been taught, which is the premise) and connect them with the elementary terms listed earlier.
That is, $S_1: \mathit{4} \leftrightarrow \dot{4}$, $S_2: \mathit{10} \leftrightarrow \dot{10}$ (The copula $\leftrightarrow$ denotes the recognition that the tokens are different but the types are the same).
}
{
Returning to the equation, the pupil focus on the numbers $\mathit{4}$ and $\mathit{10}$ that form a group on the left side of the equation (for this problem, the concepts of equal sign and fraction have been taught, which is the premise) and connects them with the basic terms listed earlier.
That is, $S_1: \mathit{4} \leftrightarrow \dot{4}$, $S_2: \mathit{10} \leftrightarrow \dot{10}$ (The copula $\leftrightarrow$ denotes the recognition that the tokens are different but the types are the same).}

\xxparatran
{また等式の空間的配置としての対称性や分数における分子／分母の役割の対称性から，以下の陳述項を認識する．
$S_3: \mathit{4} \leftrightsquigarrow \square$, $S_4: \mathit{10} \leftrightsquigarrow \mathit{5}$．
（繋辞$\leftrightsquigarrow$は，同一ではないが一定の対応関係を持つことをここでは表す．）
すると $\{S_2, S_4\} \vdash S_5: \dot{10} \leftrightsquigarrow \mathit{5}$ と推論できる．
（前提となる2つの陳述項が異なる繋辞を持つときに，推論結果がどのような繋辞を持つのかについては，今後の検討が必要である．）}
{Also, from the symmetry of the spatial arrangement of equations and the symmetry of the role of the numerator/denominator in fractions, we recognize the following statements. $S_3: \mathit{4} \leftrightsquigarrow \square$, $S_4: \mathit{10} \leftrightsquigarrow \mathit{5}$. (The suffix $\leftrightsquigarrow$ is used here to denote a certain but not identical correspondence.) Then $\{S_2, S_4\} \vdash S_5: \dot{10} \leftrightsquigarrow \mathit{5}$ can be inferred as. (It is necessary to examine what kind of affixes the inference results have when the two premises have different affixes.)}
{
Also, from the symmetry of the role of the numerator/denominator in fractions and symmetry as the spatial arrangement of equations, we recognize the following statement terms. $S_3: \mathit{4} \leftrightsquigarrow \square$, $S_4: \mathit{10} \leftrightsquigarrow \mathit{5}$．(Copula, $\leftrightsquigarrow$, expresses that is not identical but has a certain correspondence here.) Then, we can infer $\{S_2, S_4\} \vdash S_5: \dot{10} \leftrightsquigarrow \mathit{5}$ (When two statement terms as premises have a different copula, it is necessary to examine what kind of copulas the inference result has in the future).}
{
Also, from the symmetry of the role of the numerator/denominator in fractions and symmetry as the spatial arrangement of equations, the pupil recognizes the following statement terms: $S_3: \mathit{4} \leftrightsquigarrow \square$, $S_4: \mathit{10} \leftrightsquigarrow \mathit{5}$. (Copula, $\leftrightsquigarrow$, expresses that is not identical but has a certain correspondence here.) Then, $\{S_2, S_4\} \vdash S_5: \dot{10} \leftrightsquigarrow \mathit{5}$ is inferred. (It is a future work to examine what kind of copulas the inference result has when two statement terms as premises have different copulas).}

\xxparatran
{$S_5$の認識に触発され，小学生は$\dot{10}$の下に，異なる数列を並べてみることを発想する．
これが(c)の列$\ddot{0},\dots,\ddot{5}$である．
（(c)の列の最初を$1$ではなく$0$にすること，$0$から続く数を空間的に等間隔で並べることを発想するのは，
審美的感覚やゲシュタルト知覚などの，論理とはまた別の要因%
\footnote{\citet{Schelling60}は，彼が考案した調整ゲーム（待ち合わせなどを行うゲーム）の実験を通じて，
ゲームの成功には「論理よりも審美的感覚などの想像力」が重要と指摘している．}
によって誘導されているとここでは考える．）}
{Inspired by the recognition of $S_5$, elementary school students come up with the idea of arranging different sequences under $\dot{10}$.
This is the sequence $\ddot{0},\dots,\ddot{5}$ in (c). (The idea of starting the column in (c) with $0$ instead of $1$ and arranging the numbers starting with $0$ at equal intervals in space is due to aesthetic sense and Gestalt Factors other than logic, such as perception, %
\footnote{\citet{Schelling60}, found that through experiments with coordination games (games involving waiting, etc.) that he devised, the success of the game was more important than logic. He points out that ``imagination such as aesthetic sense'' is important.} is induced here.}
{Inspired by the recognition of $S_5$, pupils come up with the idea of arranging different sequences under $\dot{10}$. This is the sequence $\ddot{0},\dots,\ddot{5}$ of (c). (The starting of (c) is $0$ rather than $1$, and the idea of arranging numbers starting from 0 at equal intervals in space is induced based on other factors%
\footnote{\citet{Schelling60} pointed out that ``imagination such as aesthetic sense rather than logic'' is important for the success of the game through the experiment with designed coordination games (a game in which a meeting is held).}
rather than logic, such as aesthetic sense and Gestalt perception.
}
{
Inspired by the recognition of $S_5$, the pupil comes up with the idea of arranging different sequences under $\dot{10}$. This is the sequence $\ddot{0},\dots,\ddot{5}$ of (c). (The starting of (c) is $0$ rather than $1$, and the idea of arranging numbers starting from 0 at equal intervals in space is induced based on other factors
rather than logic, such as aesthetic sense and Gestalt perception.\footnote{\citet{Schelling60} pointed out that ``imagination such as aesthetic sense rather than logic'' is important for the success in his designed coordination games (e.g., a game to coincide with each other in a map without communication) through his experiments.})}

\xxparatran
{小学生は(b)列と(c)列の間の知覚的群化\cite{Thorisson94}により，$S_6: \dot{4} \leftrightsquigarrow \ddot{2}$ という認識を得る．
すると， 
$\{S_1, S_6\} \vdash S_7: \mathit{4} \leftrightsquigarrow \ddot{2}$，
$\{S_3, S_7\} \vdash \ddot{2} \leftrightsquigarrow \square$
を推論し，「$\square$に当てはまる数は2」という解を得ることができる．
}
{Elementary school students perceive $S_6: \dot{4} \leftrightsquigarrow \ddot{2}$ by perceptual grouping \cite{Thorisson94} between columns (b) and (c). $\{S_1, S_6\} \vdash S_7: \mathit{4} \leftrightsquigarrow \ddot{2}$, $\{S_3, S_7\} \vdash \ddot{2} \leftrightsquigarrow \square$ and you can get the A solution that ``$\square$ is 2''.}
{Pupils perceive $S_6: \dot{4} \leftrightsquigarrow \ddot{2}$ by the perceptual grouping between columns (b) and (c). Then, $\{S_1, S_6\} \vdash S_7: \mathit{4} \leftrightsquigarrow \ddot{2}$, $\{S_3, S_7\} \vdash \ddot{2} \leftrightsquigarrow \square$ is inferred and the answer “The number in $\square$ is 2.” is obtained.
}
{
The pupil perceives $S_6: \dot{4} \leftrightsquigarrow \ddot{2}$ by the perceptual grouping between rows (b) and (c). Then, $\{S_1, S_6\} \vdash S_7: \mathit{4} \leftrightsquigarrow \ddot{2}$, $\{S_3, S_7\} \vdash \ddot{2} \leftrightsquigarrow \square$ is inferred and the answer “The number in $\square$ is 2.” is obtained.}

\xxparatran
{NATLは項のソフトなマッチングを許すので，可能な推論の幅は枠組み的にはほとんど無制限である．
先の例で言えば，$\{S_4, S_6\} \vdash S_8: \mathit{10} \leftrightarrow \ddot{2}$のような，
何の役にも立たない無駄な推論もできてしまう．
このような無駄を排し，有用な結論を得られるように推論を適切に制御するには，
いつなにをするべきか，を暗黙知的に判断できる推論器が必要である．
また，上述の推論過程で出てきた，これも暗黙知的な，空間的「発想」ができなければ，
推論の部品となる適切な陳述項を得ることができない．
このように，本稿で示したNATLの理論だけで，人間が行う多様な認知的思考能力を全て説明できるわけではない．
しかし，NATLはそれらの多様な認知能力を統合して記号的推論・思考の過程の構築と説明を可能にする「骨格」となることができる．}
{Since NATL allows soft matching of terms, the range of possible inferences is framework-wise almost unlimited. In the previous example, we can make useless reasoning like $\{S_4, S_6\} \vdash S_8: \mathit{10} \leftrightarrow \ddot{2}$. It will end. In order to eliminate such waste and appropriately control reasoning so that useful conclusions can be obtained, we need a reasoner that can implicitly and intelligently decide what to do. Also, without the spatial ``idea'', which is also tacit knowledge, which came out in the reasoning process described above, it is impossible to obtain appropriate statements that serve as the components of reasoning. In this way, the theory of NATL presented in this paper alone cannot explain all of the diverse cognitive thinking abilities that humans perform. However, NATL can be a ``skeleton'' that integrates these diverse cognitive abilities and enables the construction and explanation of symbolic reasoning and thinking processes.}
{Since NATL allows soft matching of terms, the range of possible inferences is almost unlimited in the framework. In the previous example, the useless, unprofitable inference is made such as $\{S_4, S_6\} \vdash S_8: \mathit{10} \leftrightarrow \ddot{2}$. To eliminate this useless inference, the inference should be controlled appropriately in order to obtain a useful conclusion. A reasoner that can judge what should do at what time implicitly, and intelligently is needed. Also, without the implicit, intelligent, and spatial ``idea'' that is made by above mentioned inference process, it is impossible to obtain appropriate statements that serve as the components of inference. It can be seen that the NATL theory shown in this paper cannot fully explain the various cognitive thinking abilities of human beings. However, NATL can become a ``skeleton'' that can integrate these various cognitive abilities and enables the construction and explanation of symbolic reasoning and thinking processes.}
{
Since NATL allows soft matching of terms, the range of possible inferences is almost unlimited in the framework. In the previous example, the useless inference is made such as $\{S_4, S_6\} \vdash S_8: \mathit{10} \leftrightarrow \ddot{2}$. To eliminate this useless inference, the reasoning should be controlled appropriately in order to obtain a useful conclusion. A reasoner that can judge what should do at what time implicitly, and intelligently is needed. Also, without the implicit and spatial ideas used in the above-mentioned thinking process, it is impossible to obtain appropriate statements that serve as the components of reasoning. The NATL theory shown in this paper alone cannot fully explain the various cognitive thinking abilities of human beings. However, NATL can become a ``skeleton'' that can integrate these various cognitive abilities and enables the construction and explanation of symbolic reasoning and thinking processes.}

\subsection{Internal Representation Language of Multimodal AI Systems}

\xxparatran
{数学の言語が自然言語よりもその対象を明確・簡潔に表現できるように，
\S\ref{sec:trl}で導入した項表示言語は，主体（AIシステム）の中の認識と知識の構造を，自然言語を用いるよりも明確・簡潔に表現することを可能にする．
また，項表示言語の項が表示するものは，自然言語で表現できるものに限定されない．
従って，項表示言語はそれ自体で，知能ロボットのようなマルチモーダルAIシステムを構成する各種の認知モジュール間のインタフェース言語として有用性がある．
項論理を指向しているため，述語記号と存在記号を区別する必要がなく，一方で複合項の導入により，述語論理のように任意の数の項の間の関係を記述できる．}
{The term representation language introduced in \S\ref{sec:trl} is used to express the subject (AI system)'s cognition and knowledge, so that the language of mathematics can express its object more clearly and concisely than natural language. It makes it possible to express the structure of a language more clearly and concisely than using natural language. Also, the items displayed by the terms of the term display language are not limited to those that can be expressed in natural language. Therefore, the term display language itself is useful as an interface language between various cognitive modules that constitute a multimodal AI system such as an intelligent robot. Since it is oriented toward term logic, there is no need to distinguish between predicate symbols and existential symbols.On the other hand, by introducing compound terms, it is possible to describe the relationship between any number of terms as in predicate logic.}
{Like mathematical language can express objects more clearly and concisely than natural language, the term expression language introduced by \S\ref{sec:trl} might express the structure of cognition and knowledge in the subject (AI system) more clearly and concisely than natural language. In addition, those expressed by terms of term expression language are not limited to those that can be expressed by natural language. Therefore, the term expression language itself is useful as an interface language between various cognitive modules that constitute a multi-modal AI system such as an intelligent robot. Because of the pointing to term logic, there is no need to distinguish between predicate symbols and existential symbols. On the other hand, the introduction of compound terms allows us to describe the relationships between any number of terms as in predicate logic.
}
{
Like mathematical language can express objects more clearly and concisely than natural language, TRL introduced in \S\ref{sec:trl} might express the structure of cognition and knowledge in the subject (AI system) more clearly and concisely than natural language. 
In addition, those expressed by terms are not limited to those that can be expressed by natural language. Therefore, TRL itself is useful as an interface language between various cognitive modules that constitute a multi-modal AI system such as an intelligent robot. 
Since it is oriented toward term logic, there is no need to distinguish between predicate symbols and existential symbols. On the other hand, the introduction of compound terms allows us to describe the relationships between any number of terms as in predicate logic.}

\xxparatran
{\begin{figure}[t]
\footnotesize
    \centering
    \begin{tabular}{ll}
    1. & 両手鍋で油を熱する。\\
       & セロリと青ねぎとニンニクを加え、\\
       & 1分ほど炒める。\\
    2. & ブイヨンと水とマカロニと胡椒を加えて、\\
       & パスタが柔らかくなるまで煮る。\\
    3. & 刻んだセージをまぶす。
    \end{tabular}
    \caption{3ステップからなるレシピの例\cite{Maeta17}}
    \label{fig:recipe}
\end{figure}}
{\begin{figure}[t]
\footnotesize
     \centering
     \begin{tabular}{ll}
     1. & Heat the oil in a two-handled pan. \\
        & Add celery, green onion, and garlic, \\
        & fry for 1 minute. \\
     2. & add broth, water, macaroni, and pepper, \\
        & simmer until the pasta is tender. \\
     3. & sprinkle with chopped sage.
     \end{tabular}
     \caption{3-step recipe example\cite{Maeta17}}
     \label{fig:recipe}
\end{figure}}
{\begin{figure}[t]
\footnotesize
    \centering
    \begin{tabular}{ll}
    1. & Heat the oil in the saucepot. \\
       & Add celery, green onions, and garlic, \\
       & Stir fry for about 1 minute. \\
    2. & Add broth, water, macaroni and pepper, \\
       & Boil until the pasta is soft. \\
    3. & Sprinkle chopped sage.
    \end{tabular}
    \caption{A example of 3-step recipe \cite{Maeta17}}
    \label{fig:recipe}
\end{figure}}
{\begin{figure}[t]
\normalsize
     \centering
     \begin{tabular}{ll}
     1. & Heat the oil in a two-handled pan. \\
        & Add celery, green onion, and garlic, and fry for 1 minute. \\
     2. & Add broth, water, macaroni, and pepper, and simmer until the pasta is tender. \\
     3. & Sprinkle with chopped sage.
     \end{tabular}
     \caption{3-step recipe example \cite{Maeta17}}
     \label{fig:recipe}
\end{figure}}

\xxparatran
{自然言語処理が主体となるタスクであっても，明示的に言語化されない要素が重要となるものは多々ある．
手順文書の理解はその例の1つである．
\citet{Maeta17}はレシピ調理手順から有向非循環グラフとして
意味構造（レシピフローグラフ）を抽出する手法を提案している．
レシピフローグラフでは，調理に使用する食材や道具，調理行為がグラフの頂点
となり，それらの間の関係が辺のラベルとして表現される}
{Even in tasks that mainly involve natural language processing, there are many things where elements that are not explicitly verbalized are important. Understanding procedural documents is one example. \citet{Maeta17} proposes a method to extract the semantic structure (recipe flow graph) from recipe cooking procedures as a directed acyclic graph. In the recipe flow graph, the ingredients, tools, and cooking actions used for cooking are the vertices of the graph, and the relationships between them are expressed as edge labels.}
{Even in the tasks with natural language processing as the main part, there are many factors that are not explicitly linguistically become important. Understanding procedural documents is an example. \citet{Maeta17} proposed a method to extract semantic structure (recipe flow graph) from recipe cooking steps as a directed acyclic graph. The ingredients, tools, and cooking behaviors used in cooking become the vertices of the recipe flow graph, and the relationship between them is expressed by the labels on the edges.
}
{
Even in tasks that mainly involve natural language processing, there are many cases where elements that are not explicitly verbalized are important. Understanding procedural documents is one example. \citet{Maeta17} propose a method to extract the semantic structure (recipe flow graph) from a recipe cooking procedure as a directed acyclic graph. In the recipe flow graph, the ingredients, tools, and cooking actions used for cooking are the vertices of the graph, and the relationships between them are expressed as edge labels.}

\xxparatran
{レシピフローグラフでは，ある行為により状態変化が起きた行為対象や生成物が，
その後の別の行為をなすための対象や場所になることを利用して，
手順の順序関係（依存関係）を表現する．
このとき行為の頂点はそれによって生成物や状態変化が起きた対象を代表する．
例えば，\ref{fig:recipe}の第1ステップの「熱する」という行為は，その結果としての
「高温の油を擁する高温の両手鍋」を代表し，
その後の「加える」という行為の「方向 (Dest)」として関係付けられる
（グラフ表現上では頂点「熱す」から頂点「加え」にDestというラベルの付いた辺が張られる）．
このように，手順文書の解釈では，言語化されない中間生成物や状態変化の適切な取り扱いが鍵となるが，
言語化された要素だけでその内容を表現することには限界があるだろう\footnote{%
例えば「すいかをくり抜く」という行為の後にはくり抜かれた実と皮が残るが，それぞれ食材と容器としてその後の工程で別々に使用・参照されることがしばしばある．フローグラフでは異なる2つの対象を「くり抜く」という行為の結果として表される1つの状態で代表しなければいけない．}．
項表示言語を用いれば，言語化された対象と言語化されていない対象を一様・簡潔に表示できるので，
より精度の高い意味解析・状況理解の実現を期待できる．}
{A recipe flow graph expresses the order (dependency) of a procedure by using the fact that an action target or a product that causes a state change due to a certain action becomes an object or place for performing another action after that. do. In this case, the vertex of the action represents the product or the object whose state is changed by it. For example, the act of ``heating'' in the first step of \ref{fig:recipe} represents the resulting ``hot two-handled pan with hot oil'', followed by the act of ``adding''. is related as the ``direction (Dest)'' of . In this way, the key to interpreting procedure documents is the appropriate handling of intermediate products and state changes that are not verbalized, but there is a limit to expressing the content only with verbalized elements. \footnote{%
'For example, after the act of ``cutting out a watermelon,'' the cut fruit and skin remain, but they are often used and referred to separately in later processes as ingredients and containers, respectively. In the flow graph, two different objects must be represented by one state that is expressed as a result of the act of "hollowing out". }.
By using a term display language, it is possible to uniformly and concisely display both verbalized and non-verbalized objects, so we can expect to achieve more accurate semantic analysis and situation understanding.}
{In the recipe flow graph, the behavior object or product whose state changes due to a certain behavior represents the order (dependency) of ordinal relation by taking advantage of being an object or place that becomes the other behaviors later. At that time, the vertices of the behaviors represent the product or the object whose state changes based on that. For example, the ``heating'' behavior of step 3 represents the result ``high temperature saucepot has high temperature oil'' associated as the ``Direction (Dest)'' of the subsequent act of ``adding'' (In the graph representation, the edge labeled by ``Dest'' extends from the vertex ``hot'' to the vertex “add”). Like this, in the interpretation of procedure documents, the key is to properly process intermediate products and state changes that are not verbalized. However, there is a limit to expressing that content only in terms of verbalized elements.\footnote{%
For example, after the act of 
``digging watermelon'', the excavated fruit and peel will be left, and they are often used as food materials and containers respectively, which will be used and referred to in subsequent processes. In the flow graph, the different two objects must be represented by one state that presents as the result of behavior like ``digging''.} If the term representation language is used, the objects that have been verbalized and the objects that have not been verbalized can be presented uniformly, and concisely. Therefore, it can be expected to achieve more accurate semantic analysis and situation understanding.
}
{
A recipe flow graph expresses the order (dependency) of a procedure by using the fact that an action target or a product that causes a state change due to a certain action becomes an object or place for performing another action after that. In this case, the vertex of an action represents the product or the object whose state is changed by it. For example, the act of ``heating'' in the first step of Fig. \ref{fig:recipe} represents the resulting ``hot two-handled pan with hot oil'', and it is related as the ``direction (Dest)'' of the following act of ``adding''. In this way, the key to interpreting procedure documents is the appropriate handling of intermediate products and state changes that are not verbalized, but there is a limit to expressing the content only with verbalized elements. \footnote{%
'For example, after the act of ``cutting out a watermelon,'' the cut fruit and skin remain, but they are often used and referred to separately in later processes as ingredients and containers, respectively. In the flow graph, two different objects must be represented by one state that is expressed as a result of the act of "hollowing out".}
By using TRL, it is possible to uniformly and concisely express both verbalized and non-verbalized objects, so we can expect to achieve more accurate semantic analysis and situation understanding.}

\subsection{Understanding and generation of analogy and metaphor} \label{sec:analogy_metaphor}

\xxparatran
{アナロジーは，ある領域の概念集合を別の領域の概念集合に写像することで，
新しい状況での問題解決を容易にする\cite{Holyoak95}．
同様にして，比喩は抽象的な対象や新しい状況の理解とその言語化の助けになる\cite{Lakoff80}が，
それだけでなく，
会話者が新規な対応関係を積み重ねることで会話参与者間の（感情的な）繋がりを深めたり\cite{Jang17}，
社会的相互作用の観点からも重要な現象である．
いずれも人間の創造性の源泉の一部と認識されているが，少なくともアナロジーは
必ずしも生得的な能力ではなく，
発達の過程で段階的に獲得される部分も多いことがわかっている\cite{Holyoak95}}
{An analogy maps a set of concepts in one domain to a set of concepts in another domain to facilitate problem solving in new situations \cite{Holyoak95}.
Similarly, metaphors help the understanding of abstract objects and new situations and their verbalization \cite{Lakoff80}. It is also an important phenomenon from the perspective of deepening (emotional) ties and social interaction. Both are recognized as part of the source of human creativity, but at least analogy is not necessarily an innate ability, and it is known that many parts are acquired step by step in the process of development. \cite{ Holyoak95}}
{The analogy is to map the concept set of one domain to the concept set of another domain, making it easy to solve problems under new situations \cite{Holyoak95}. Similarly, metaphor is helpful to understand abstract objects, and new situations and their verbalization \cite{Lakoff80}, but not only that, interlocutors deepen the (emotional) connection between dialogue participants by accumulating new corresponding relationships \cite{Jang17}, which is also an important phenomenon from the perspective of social interaction. Although all of them are considered as part of the source of human creativity, at least analogies are not always an innate ability, and many of them are acquired in the different stages during the growth process \cite{Holyoak95}.
}
{
Analogy maps a set of concepts in one domain to a set of concepts in another domain to facilitate problem solving in new situations \cite{Holyoak95}.
Similarly, metaphor helps the understanding of abstract objects and new situations, and their verbalization \cite{Lakoff80}. 
Metaphor is also an important phenomenon from the perspective of deepening (emotional) ties and social interaction \cite{Jang17}.
Both are recognized as the sources of human creativity, but at least analogy is not necessarily an innate ability, and it is known that many parts are acquired step by step in the process of development \cite{Holyoak95}.}

\xxparatran
{NATLの複合項は集合表現として用いることができ，
その集合が何を表しているかという集合自体についてのメタな情報を，複合項の関係項$R$の意味ベクトル
にエンコードすることができる}
{A compound term in NATL can be used as a set representation, and meta-information about the set itself, what the set represents, can be encoded in the semantic vector of the compound term's relational term $R$.}
{The compound term of NATL can be presented as a set, and meta-information about the set itself, what the set represents, can be encoded in the semantic vector of linkage term $R$ of the compound term.
}
{
A compound term in NATL can be used as a set representation, and meta-information about the set itself, what the set represents, can be encoded in the semantic vector of the compound term's relational term $R$.}

\xxparatran
{項表示言語により，アナロジーや比喩における写像を，概念と概念の連結として，陳述項 $B_S \to B_T$ の形で素直に表示することができる．そして個々の写像を束ねたものを，
1つの複合項 $(R, B_S^1\to B_T^1, B_S^2\to B_T^2, \ldots, B_S^n \to B_T^n)$ として表示できる\footnote{%
例えば，\cite{Jang17}で挙げられている``He is the pointing gun, we are the bullets of his desire.''という比喩からは，$(\term{METAPHOR}$, $\term{he} \to \term{gun}$, $\term{we} \to \term{bullet})$ というような写像構造の表示を得られるだろう．
仮に``He is the pointing gun.''とだけ聞けば，一般的には銃に関するフレーム知識\cite{Fillmore76}から，
$(\term{METAPHOR}$, $\term{he} \to \term{gun}$, $\term{we} \to \term{target})$というような
写像構造として比喩を解釈するのが普通であろう．
どちらの写像構造を文脈として持つかで，``He will not be stingy with his bullets.''の解釈やその後の比喩の使用が変わる．
関係項$\term{METAPHOR}$に付与するベクトル意味表現には，比喩であることをマークするベクトルと，そこで用いられている意味フレームを表すベクトルを足し合わせることで，推論器が適切に動作するための情報をエンコードすることが考えられる．
}．
この表現方法をもとにNATLに基づく推論器の上で，アナロジーや比喩を駆使するシステムを実現できる可能性がある．
人間においてもアナロジー・比喩の使用が発達的に獲得されることは，
推論器に対して相応の訓練が必要であることと相似する．}
{With the term display language, we can simply display the mapping in analogy and metaphor in the form of a statement term $B_S \to B_T$ as a connection between concepts. Then, the individual mappings are bundled into one compound term $(R, B_S^1\to B_T^1, B_S^2\to B_T^2, \ldots, B_S^n \to B_T^n)$ can be represented as 
\footnote{%
For example, $(\term{METAPHOR}$, $\term{he} \ to \term{gun}$, $\term{we} \to \term{bullet})$ will give an indication of the mapping structure. 
If you just ask ``He is the pointing gun.'', you will generally get $(\term{METAPHOR}$, $\term{he} \to \ It is common to interpret metaphors as mapping constructs such as term{gun}$, $\term{we} \to \term{target})$. The interpretation of ``He will not be stingy with his bullets.'' and the subsequent use of metaphors will change depending on which mapping structure is used as the context. For the vector semantic representation attached to the relational term $\term{METAPHOR}$, the vector that marks the metaphor and the vector that represents the semantic frame used in the metaphor are added together so that the reasoner can operate properly. It is conceivable to encode information for }.
Based on this expression method, it is possible to realize a system that makes full use of analogies and metaphors on a NATL-based reasoner. The developmental acquisition of the use of analogies and metaphors in humans is similar to the need for appropriate training for reasoners.}
{
According to the term representation language, the mapping in analogy and metaphor, as the connection of concepts and concepts, can be presented frankly in the form of the statement term $B_S \to B_T$. Then, what bundles each mapping can be presented by one composite term $(R, B_S^1\to B_T^1, B_S^2\to B_T^2, \ldots, B_S^n \to B_T^n)$. \footnote{%
For instance, the metaphor example ``He is the pointing gun, we are the bullets of his desire.'' is given by the present mapping structure like \cite{Jang17},$(\term{METAPHOR}$, $\term{he} \to \term{gun}$, $\term{we} \to \term{bullet})$ might be obtained. If you only listen that ``he is the pointing gun.'', because of the frame knowledge about guns \cite{Fillmore76}, $(\term{METAPHOR}$, $\term{he} \to \term{gun}$, $\term{we} \to \term{bullet})$ such mapping structure is usually used to explain metaphors. Depending on which mapping structure is used in the context, the interpretation of ``He will not be stingy with his bullets.'' and the use of subsequent metaphors will also change. The vector semantic representation attached to the relational term $\term{METAPHOR}$ encodes information for proper operation of the inference machine by adding the vector that marks the metaphor and the vector that represents the semantic frame used here.} Based on this representation method, it is possible to realize a system using analogy and metaphor on the reasoner based on NATL. As for human beings, they can acquire the usage of analogy and metaphor in the stage of development, which is similar to the necessary corresponding training of the reasoner.
}
{
With TRL, we can simply represent the mapping in analogy and metaphor in the form of a statement term $B_S \to B_T$ as a connection between concepts. Then, the individual mappings can be bundled into one compound term $(R, B_S^1\to B_T^1, B_S^2\to B_T^2, \ldots, B_S^n \to B_T^n)$.\footnote{%
For example, an example from \cite{Jang17}, ``He is the pointing gun. We are the bullets  of his desire.", will give an mapping structure like $(\term{METAPHOR}$, $\term{he} \ to \term{gun}$, $\term{we} \to \term{bullet})$. 
If it is simply ``He is the pointing gun.'', you will generally get $(\term{METAPHOR}$, $\term{he} \to \term{gun}$, $\term{we} \to \term{target})$. The interpretation of ``He will not be stingy with his bullets.'' and the subsequent use of metaphors will change depending on which mapping structure is used as the context. For the vector semantic representation attached to the relational term $\term{METAPHOR}$, it is conceivable to encode information that marks metaphor and that represents the semantic frame used in the metaphor.}
Based on this expression method, it is possible to realize a system that makes full use of analogies and metaphors on a NATL-based reasoner. The developmental acquisition of the use of analogies and metaphors in humans is similar to the need for appropriate training for reasoners.}

\xxparatran
{\citet{Hofstadter2013}は，さらに踏み込み，アナロジーは人間の認知・思考の根源であると
主張している．Hofstadterらの主張では，記憶の想起もアナロジーと同じ仕組みによる．
そして，分類（categorization）という根源的な認知能力がそもそもアナロジーを作る能力と同一であると説く．
NATLの出発点である項論理の別名は分類論理（categorical logic）である\cite{Wang13}．
Hofstadterらの主張が正しければ，NATLはまさに思考の形式表現の根幹となれるはずである．}
{\citet{Hofstadter2013} goes further and claims that analogy is the root of human cognition and thinking. According to Hofstadter et al., recall of memory is based on the same mechanism as analogy. He then argues that the fundamental cognitive ability of categorization is the same as the ability to create analogies. Another name for term logic, the starting point of NATL, is categorical logic. If Hofstadter et al.'s assertion is correct, NATL should become the basis of the formal expression of thought}
{\citet{Hofstadter2013} further proposed that analogy is the root of human cognition and thinking. Hofstadter et al. believe that the recall structure of memory is the same as that of analogy. He also said that the fundamental cognitive ability of categorization is the same as the ability to establish analogical thinking. The start point of NATL is the term logic and the alias of it is called ``categorical logic'' \cite{Wang13}. If Hofstadter et al. is right, NATL is the backbone of the formal expression of thinking.
}
{
\citet{Hofstadter2013} goes further and claims that analogy is the root of human cognition and thinking. According to Hofstadter et al., recall of memory is based on the same mechanism as analogy. He then argues that the fundamental cognitive ability of categorization is the same as the ability to create analogies. Another name for term logic, the starting point of NATL, is categorical logic. If Hofstadter et al.'s assertion is correct, NATL should become the basis of the formal expression of thought.}

\xxparatran
{\citet{Fauconnier02}は，比喩・アナロジーを含む，人間の創造的・発見的思考の普遍的なモデルとして，概念ブレンディングという理論を提示している．
概念ブレンディングの理論は，2つの入力領域（メンタルスペース）の間を様々な関係で写像的に接続し1つの新規な解釈領域として融合するという機構に基づいて，物理法則・因果律に拘束されない人間の柔軟な思考を説明する．
ブレンディングを計算論的に扱おうとする研究は多くないが\footnote{
\cite{Fauconnier02}の共著者のM. Turnerによって
まとめられた論文一覧には2010年から2019年までで12件しかない．
\url{https://markturner.org/blending.html}}
，項表示言語によりメンタルスペースと写像関係を表示することで，概念ブレンディングと非公理的項論理を統合した思考の計算理論を実現することも目指したい．}
{\citet{Fauconnier02} presents the theory of concept blending as a universal model of human creative and heuristic thinking, including metaphors and analogies. The theory of concept blending is based on the mechanism of connecting two input areas (mental space) mapographically in various relationships and fusing them as one new interpretation area. Explain flexible thinking. Although there are not many studies that attempt to treat blending computationally,
\footnote{\cite{Fauconnier02} has only 12 publications from 2010 to 2019 in the list compiled by co-author M. Turner.
\url{https://markturner.org/blending.html}},  I also want to realize a computational theory of thinking that integrates concept blending and non-axiomatic term logic by displaying mental space and mapping relations in a term display language.}
{\citet{Fauconnier02} proposed the conceptual blending theory as a universal mode of human creativity and discovery thinking including metaphor and analogy. The theory of conceptual blending based on the structure that the two input areas (mental spaces) are connected by various relationship mappings and merged into a new interpretation area explains the flexible thinking of human beings which is not constrained by the laws of physics and causality. Although there are not many studies that attempt to process blending computationally \footnote{
There are only 12 papers in the list collated by M.Turner, the co-author of \cite{Fauconnier02}, from 2010 to 2019. \url{https://markturner.org/blending.html}}, our another goal is to realize the computational theory of thinking that integrates conceptual blending and NATL by representing mental spaces and mapping relationships based on TRL.
}
{
\citet{Fauconnier02} proposed the conceptual blending theory as a universal mode of human creativity and discovery thinking including metaphor and analogy. The theory of conceptual blending is based on the structure where the two input areas (mental spaces) are connected by various relationship mappings and merged into a new interpretation area and can explain the flexible thinking of human beings which is not constrained by the laws of physics and causality. 
Although there are not many studies that attempt to process blending computationally \footnote{There are only 12 papers in the list collated by M.Turner, the co-author of \cite{Fauconnier02}, from 2010 to 2019. \url{https://markturner.org/blending.html}}, our another goal is to realize the computational theory of thinking that integrates conceptual blending and NATL by representing mental spaces and mapping relationships based on TRL.
}

\section{Prospects for Computer Implementation and Future Challenges}\label{sec:implementation_and_issues}

\xxparatran{\S\ref{sec:argumentation}および\S\ref{sec:applications}で
例示した非公理的項論理に基づく推論は，
意味的類似性に基づく柔軟な単一化の能力と，
適切な単一化だけを選択する能力の存在を前提としていた．
これらの能力の実現は，今後の研究課題である．
本章では，これらの能力の実装について見通しと課題を述べる．}
{Reasoning based on non-axiomatic argument logics, exemplified in \S\ref{sec:argumentation} and \S\ref{sec:applications}, is based on the ability of flexible unification based on semantic It was premised on the existence of the ability to select only unity. Realization of these abilities is a future research topic. This chapter describes the outlook and challenges for implementing these capabilities.
}
{The reasoning based on the non-rational term logic exemplified by \S\ref{sec:argumentation} and \S\ref{sec:applications} was premised on the existence of the flexible unification ability based on semantic similarity and the ability to select only appropriate unification in advance.
The realization of these abilities is a topic for future research.
This chapter describes the outlook and challenges for implementing these capabilities.
}
{
The reasoning based on NATL exemplified in S\ref{sec:argumentation} and \S\ref{sec:applications} is premised on the ability of the flexible unification based on semantic similarity and the ability to select only appropriate unification.
The realization of these abilities is a future research topic.
This section describes the outlook and challenges for implementing these capabilities.}

\subsection{Semantic vector representations}\label{sec:vector_semantics}

\xxparatran{\S\ref{sec:latent_repr}で述べたように，多次元ベクトルの形式で，
潜在空間内の点として意味表象を項に付与することで，項が指し示す事物や概念の情報を保持する．
\S\ref{sec:cont_copula}で述べたように，繋辞にも項同様の意味表象を与える．
項$T$に付与された意味表象（意味ベクトル）を$\mu(T)$で表すと，
基本項$B$，複合項$C$，陳述項$S$，連関項$L$，繋辞$c$の意味ベクトルを
与える関数を以下のように定義できる（$R$，$T_i$は\S\ref{sec:trl}の定義に従う．$T_{l/r}$は左項/右項を表す．）．}
{As described in , by giving a semantic representation to a term as a point in the latent space in the form of a multidimensional vector, it holds the information of the thing or concept pointed to by the term. As mentioned in \S\ref{sec:cont_copula}, suffixes have the same semantic representation as terms. If the semantic representation (semantic vector) assigned to the term “$T$” is represented by $\mu(T)$, the basic term $B$, the compound term $C$, and the declarative term $S$, the associated term $L$, and the prefix $c$ can be defined as follows (where $R$ and $T_i$ are \S\ref{sec:trl} $T_{l/r}$ represents the left term/right term).}
{As described in \S\ref{sec:latent_repr}, by adding a semantic representations to a term as a point in the latent space in the form of multidimensional vector, information about the things and concepts to which the terms refer is retained.
As mentioned in \S\ref{sec:cont_copula}, the copula is given the same semantic representation as the term.
We can define the function that gives the semantic vector of the basic term $B$, the compound term $C$, the statement term $S$, the linkage term $L$, and the copula $c$ as follows ($R$ and $T_i$ follow the definition of \S\ref{sec:trl}. $T_{l/r}$ denotes the left/right terms).
}
{
As described in \S\ref{sec:latent_repr}, by adding a semantic representations to a term as a point in the latent space in the form of multidimensional vector, information about the things and concepts to which the terms refer is retained.
As mentioned in \S\ref{sec:cont_copula}, copulas are given semantic representations as well as terms.
We can define the function that gives the semantic vector of basic term $B$, compound term $C$, statement term $S$, linkage term $L$, and copula $c$ as follows ($R$ and $T_i$ follow the definition of \S\ref{sec:trl}. $T_{l/r}$ denotes the left/right terms).
Functions $F_\cdot()$ can be constructed as neural networks.
}

\begin{align*}
    \mu(B) &= F_B(B)\\
    \mu(C) &= F_C(\mu(R), \mu(T_1), ..., \mu(T_n)) \\
    \mu(S) &= F_S(\mu(c), \mu(T_l), \mu(T_r)) \\
    \mu(L) &= F_L(\mu(c), \mu(T_l), \mu(T_r)) \\
    \mu(c) &= F_c(c)
\end{align*}

\xxparatran{関数$F_\cdot()$はいずれもニューラルネットとして構築するのがよいと思われる．}{
The function $F_ \cdot()$ should be constructed as a neural network.
}
{
It is considered that all functions $F_\cdot()$ should be constructed as neural networks.
}
{
}

\xxparatran{$F_B(B)$は，基本項$B$と結びついている種々の情報を，潜在空間表象としてエンコードする．
例えば，項$B$が自然言語により提示された記号，つまり特定の単語列に対して
認識された項であるならば，BERT~\cite{BERT}などの事前学習済み言語モデルを
用いて$F_B$を構成することが考えられる．
同様にして，その項が視覚的・聴覚的に知覚されている特定の対象と結びついているならば，
その感覚刺激をエンコードする関数として$F_B$を構成できるだろう．
おそらくオートエンコーダとして構成することが基本の実装方法になるだろう．}
{$F_B(B)$ encodes various information associated with the base term $B$ as a latent space representation. For example, if the term $B$ is a symbol presented by natural language, that is, a term recognized for a particular word string, then a pretrained language model such as BERT~\cite{BERT} is used to $F_B$ can be considered. Similarly, if the term is associated with a particular object being visually or audibly perceived, then $F_B$ could be constructed as a function encoding that sensory stimulus.
Perhaps configuring it as an autoencoder would be the basic implementation.}
{$F_B(B)$ encodes various information associated with the basic term $B$ as latent space representations.
For example, if the term $B$ is a symbol presented by natural languages, such as a term recognized for a particular word sequence, then a pre-trained language model such as BERT~\cite{BERT} could be used to construct $F_B$.
In the same way, if the term is associated with a particular object that is perceived visually or audibly, $F_B$ could be constructed as a function that encodes that sensory stimulus.
Perhaps the basic implementation method is to configure it as an autoencoder.
}
{
$F_B(B)$ encodes various information associated with basic term $B$ as a latent space representation.
For example, if term $B$ is a symbol presented by a natural language, such as a term recognized for a particular word sequence, then a pre-trained language model such as BERT~\cite{BERT} could be used to construct $F_B$.
In the same way, if the term is associated with a particular object that is perceived visually or audibly, $F_B$ could be constructed as a function that encodes that sensory stimulus.
Perhaps the basic implementation approach is to configure it as an autoencoder.
}

\xxparatran{$F_c$はいわゆる埋め込み層（誤差逆伝搬によって重みを学習し，
明示化された入力空間の疎ベクトルから潜在空間の密ベクトルに線形変換を行うルックアップテーブルとしての行列表現）
として対応するベクトルを直接学習するのが
出発点としてはよいだろう．
ただしこの場合は，\S\ref{sec:cont_copula}で考察したように繋辞を派生的・開放的に捉えるのではなく，
予め定めた有限個の繋辞で閉じたシステムを構築することになる．
従って，本研究構想においては，あくまで出発点としてのアプローチである．
$F_c$を何かしらの副課題で事前に構築するのか，
後述する推論器の学習の中で乱数初期値から同時に獲得するのかは，
今後の検討課題である．}
{$F_c$ is a so-called embedding layer (a matrix representation as a lookup table that learns weights by error back-propagation and linearly transforms sparse vectors in the input space to dense vectors in the latent space). Direct learning is a good starting point. However, in this case, instead of treating the suffixes in a derivative and open manner as discussed in \S\ref{sec:cont_copula}, we construct a closed system with a predetermined finite number of suffixes. Become. Therefore, in this research concept, it is an approach as a starting point. Whether $F_c$ is constructed in advance by some sub-task, or whether it is simultaneously obtained from the random number initial value during training of the inference machine, which will be described later, is a subject for future study.}
{$F_c$ as a so-called embedded layer (Matrix representation as a lookup table that learns weights by error back propagation and performs a linear transformation from the sparse vector of the explicit input space to the dense vector of the latent space) learn the corresponding vector directly, which is a good starting point.
However, in this case, instead of viewing the copula as derivative and open, as discussed in \S\ref{sec:cont_copula}, we decide to construct a closed system with a predefined finite number of copulas.
Therefore, this approach is just an approach as a starting point.
Whether $F_c$ is constructed in advance through some kind of sub-tasks or or to acquire it from the initial random values at the same time in the training of reasoner described later is a topic for future study.
}
{
$F_c$ as a so-called embedded layer (Matrix representation as a lookup table that learns weights by error back propagation and performs a linear transformation from the sparse vector of the explicit input space to the dense vector of the latent space) learns the corresponding vector representation directly, which is a good starting point.
However, in this case, instead of viewing copulas as derivative and open as discussed in \S\ref{sec:cont_copula}, a closed system with a predefined finite number of copulas is constructed.
Therefore, this approach just serves a starting point.
Whether $F_c$ is constructed in advance through some kind of sub-tasks or or to acquire it from the initial random values at the same time in the training of reasoner described later is a subject for future study.}

\xxparatran{繋辞の意味を表すベクトル表現をどのように構成するべきかは重要な問題であり，工学的な力技に任せるだけでなく，定性的な考察に基づく科学的な理解を深めることも不可欠だろう．以下では，まず言語の特徴として重要な「否定」に焦点を当てて，定性的な考察を進める．}
{How to construct a vector representation of the meaning of affixes is an important issue, and it is essential to deepen scientific understanding based on qualitative considerations, rather than leaving it to engineering brute force. Let's go. In the following, we first focus on ``negative'', which is an important feature of language, and proceed with a qualitative discussion.}
{How to construct a vector representation that expresses the meaning of a copula is an important issue, and it will be essential to deepen scientific understanding based on qualitative considerations rather than just leaving it to engineering abilities.
In the following, we will first focus on ``negation'', which is an important feature of the language, and proceed with qualitative considerations.
}
{
How to construct a vector representation that expresses the meaning of a copula is an important issue, and it will be essential to deepen scientific understanding of copulas based on qualitative considerations rather than just leaving it to engineering brute force.
In the following, we will focus on ``negation'', which is an important feature of natural language, and proceed with qualitative considerations.
}

\xxparatran{否定は人間が「思考実験」を行うことを可能にする知能にとって重要な要素であるが\cite{Todayama14}，
大規模言語モデルにおいてもどこまで扱えているのかまだ明らかでない問題の1つである\cite{Tashiro22}．
また，大規模知識グラフを埋め込みによって高速に検索する技術の研究が盛んになされており，
その中で，否定を含む述語論理式として表現された検索クエリを扱う研究もなされている（例えば\cite{ConE_Neurips21}）．}
{Negation is an important component of intelligence that allows humans to conduct ``thought experiments'' \cite{Todayama14}, but it is one of the problems that is not yet clear to what extent even large-scale language models can handle it \cite{Tashiro22}. 
In addition, there is a lot of research on technology to search large-scale knowledge graphs at high speed by embedding.
Among them, there is also research that deals with search queries expressed in predicate logic that include negation (for example, \cite{ConE_Neurips21}).}
{Negation is an important factor for intelligence that enables humans to perform ``thought experiments'' \cite{Todayama14}, but it is one of the problems that it is unclear to what degree it can be handled even in large-scale language models\cite{Tashiro22}.
In addition, research on techniques for searching large-scale knowledge graphs at high speed by embedding is being actively conducted, and research on handling search queries expressed as predicate logic expressions including negation is also being conducted (for example, \cite{ConE_Neurips21}).
}
{
Negation is an important factor for intelligence that enables humans to perform ``thought experiments'' \cite{Todayama14}, but it is one of the problems that it is unclear to what degree it can be handled even in large-scale language models \cite{Tashiro22}.
In addition, there is a lot of research on technology to search large-scale knowledge graphs at high speed by embedding.
Among them, there is also research that deals with search queries expressed in predicate logic that include negation (for example, \cite{ConE_Neurips21}).}

\xxparatran{前者においては，否定はnotやneverという語として，純粋に他の一般的な語と同等に，自己教師あり学習の元で「語の用法」として学習されている．従って，対義語に関する推論など用法的な要素が強いものについては，言語モデルが獲得する意味表現を用いるだけである程度対処できると予想されるし，一方で\S\ref{sec:separation_of_logic_and_language}で取り上げた\cite{Nye_Dual-System_Neurips21}の議論からは，それだけではうまく機能しない場面が多々あることも示唆される．現在得られている知見\cite{Tashiro22}からも，そのような2面的な傾向が確認されているように見える．}
{In the former, negation is learned purely as the words not and never as ``uses of words'' under self-supervised learning in the same way as other general words. Therefore, it is expected that the semantic expressions acquired by the language model can be used to some extent to deal with things that have a strong usage element, such as reasoning about antonyms. The discussion of \cite{Nye_Dual-System_Neurips21} also suggests that there are many situations where this alone does not work well. The current findings \cite{Tashiro22} seem to confirm such a two-sided tendency.}
{In the former, negation as the words ``not'' and ``never'', just like other common words, are learned as ``word usage'' under self-supervised learning.
Therefore, it is expected that the semantic representation obtained by the language model can be used to deal with to some extent with the strong pragmatic elements such as inference about antonyms, while the discussion in \S\ref{sec:separation_of_logic_and_language} suggests \cite{Nye_Dual-System_Neurips21} that there are many situations where it does not work well on its own.
It seems that such a two-sided tendency is confirmed from the currently obtained knowledge \cite{Tashiro22}.
}
{
In the former, negation as the words \textit{not} and \textit{never} are learned as ``use of words'' under self-supervised learning in the same way as other common words. Therefore, it is expected that the semantic expressions acquired by a language model can be used to some extent to deal with things that have a strong usage element, such as reasoning about antonyms. The discussion of \cite{Nye_Dual-System_Neurips21} also suggests that there are many situations where this alone does not work well. The current findings \cite{Tashiro22} seem to confirm such a two-sided tendency.}

\xxparatran{後者においては，「エジソン」や「電球」といった事物をノードとし，「発明した」というような関係をグラフのエッジとして表現した上で，ノードに潜在空間の位置としてのベクトル表現を与え，エッジに同空間内の方位としてのベクトル表現を対応させることで，検索を空間内の近接性に基づく演算（近傍探索）として行えるようにしている．ここでは，論理和・論理積・否定といった論理的要素自体は，埋め込み表現を与えられるのではなく，特別な操作（例えば分布の反転）として処理系に固定的に組み込まれている．}
{In the latter, things such as "Edison" and "light bulb" are used as nodes, and the relationship such as "invented" is expressed as the edge of the graph, and then the node is given a vector expression as the position of the latent space, and the edge is given a vector expression. By associating the vector representation as the orientation in the same space, the search can be performed as an operation (neighborhood search) based on the proximity in the space.
Here, the logical elements such as OR, AND, and Negative are not given embedded expressions, but are fixedly incorporated into the processing system as special operations (for example, inversion of distribution).}
{In the latter case, things such as ``Edison'' and ``light bulb'' are used as nodes, and the relationship such as ``invented'' is expressed as the edge of the graph, the node is represented as a vector representation of the location in the latent space and the edge is represented as a vector representation of the orientation in the same space, allowing the search to be performed as an operation based on proximity in space (neighborhood search).
Here, logical elements such as OR, AND, and negation themselves are not given embedded representations, but are fixedly incorporated into the processing system as special operations (for example, distribution inversion).
}
{
In the latter case, things such as ``Edison'' and ``light bulb'' are used as nodes, and the relationship such as ``invented'' is expressed as the edge of the graph, the node is represented as a vector representation of the location in the latent space and the edge is represented as a vector representation of the orientation in the same space, allowing the search to be performed as an operation based on proximity in space (neighborhood search).
Here, logical elements such as OR, AND, and negation themselves are not given embedded representations, but are fixedly incorporated into the processing system as special operations (for example, distribution inversion).}

\xxparatran{提案したNATLの枠組みにおいては，否定関係を含む繋辞は，「発明した」のような一般の関係表現とは異なる特別な位置づけを与えられながらも，大規模言語モデルにおける否定語のように，それ自体にベクトル表現が与えられる必要がある．
最も単純には，先に述べたように「埋め込み層」によって，タスクに最適なベクトル表現を，繋辞の集合に対して離散的に獲得させることが考えられる．
しかし，「AはBである」と「AはBでない」のような相補的な関係を持つ繋辞については，空間内で逆方向を向くような制約をかけることで，より推論器が有効に働くベクトル表現を得られるかもしれない．}
{In the proposed NATL framework, suffixes containing negative relations are given a special status different from general relational expressions such as "invented". It must itself be given a vector representation. The simplest way is to discretely acquire the optimal vector representation for the task for a set of affixes by means of an ``embedding layer'' as described above. However, for suffixes with complementary relationships such as ``A is B'' and ``A is not B'', the reasoner can be more effective by constraining them to face in opposite directions in space. might get a vector representation that works for}
{In the proposed NATL framework, copulas containing negative relations are given a special position different from general relational expressions such as ``invented'', but like negative words in large-scale language models, it needs to be given a vector representation in itself.
In the simplest case, as mentioned above, it is conceivable to obtain the optimum vector representation for the task discretely with respect to the set of copulas by the ``embedding layer''.
However, for copulas with complementary relations, such as ``A is B'' and ``A is not B'', the reasoner is more effective by applying a constraint that points in the opposite direction in space. 
}
{
In the proposed NATL framework, copulas containing negative relations are given a special position different from general relational expressions such as ``invented'', but like negative words in large-scale language models, it needs to be given a vector representation.
In the simplest case, as mentioned above, it is conceivable to obtain the optimum vector representation for the task discretely with respect to the set of copulas by the ``embedding layer''.
However, for copulas with complementary relations, such as ``A is B'' and ``A is not B'', the reasoner would work more effectively by applying a constraint that makes those copulas point to the opposite directions in the latent space. }

\xxparatran{さらに否定から離れて繋辞一般について考えると，繋辞の本質は，全体的な類似・相違を超えて，特定の視点から2つの要素を結合する（あるいは分離する）点にあるといえるだろう．これの最も端的な例は，\S\ref{sec:analogy_metaphor}で議論した，比喩・アナロジーにおいて異質なものを動的に対応付ける写像関係である．
一方で述語論理ベースの知識グラフにおける埋め込みは，類似したものが類似した位置に，同じ関係を持つものが同じ方向に存在するように埋め込むことで機能する静的な性質のものであり，これまでまったく異質なものとして遠く認識されていたものを次の瞬間には間近に結びつけて認識するという認知的処理に馴染まない．
この繋辞についての考察は，宇宙物理学において理論的な存在として議論されるワームホールを連想させる．つまり空間のワープである．
このことから，まだ空想的な着想の段階に過ぎないが，微分幾何学によって空間の歪みを記述する数学的枠組みを用いて，繋辞が持つベクトル表現に応じた潜在空間の動的な変形をモデル化するという試みの可能性が示唆される．}
{
Moving further away from negation and considering affixes in general, we can say that the essence of affixes is to go beyond their overall similarity/difference to join (or separate) two elements from a particular point of view. The most obvious example of this is the mapping relationship that dynamically associates heterogeneous metaphors and analogies, discussed in \S\ref{sec:analogy_metaphor}. On the other hand, embeddings in predicate logic-based knowledge graphs are static in nature, functioning by embedding similar objects in similar positions and objects with the same relationship in the same direction. We are not accustomed to the cognitive processing of recognizing something that was previously perceived as something completely different from a distance, but which in the next moment is closely connected. Consideration of this prefix is reminiscent of wormholes, which are discussed as theoretical entities in astrophysics. In other words, it is a warp of space. For this reason, although it is still only at the stage of an imaginative conception, using a mathematical framework that describes the distortion of space using differential geometry, we can dynamically transform the latent space according to the vector representation of the affix. The possibility of an attempt to model is suggested.}
{If we move away from negation and consider copulas in general, we can say that the essence of copulas is to combine (or separate) two elements from a specific point of view, beyond the overall similarities or differences.
The most straightforward example of this is the mapping relationship that dynamically associates different things in metaphors and analogies, as discussed in \S\ref{sec:analogy_metaphor}.
On the other hand, embedding in predicate logic-based knowledge graphs is static in nature, functioning by embedding similar things in similar positions and having the same relationship in the same direction. It does not fit into the cognitive processing of recognizing what were previously perceived as completely dissimilar and distant, and then in the next instant, recognizing them in close proximity.
This discussion of copulas is reminiscent of wormholes, which are discussed as a theoretical entity in astrophysics. In other words, it is a warp of space.
From this, although it is still only at the stage of fanciful conception, this suggests the possibility of attempting to model the dynamic deformation of a latent space according to the vector representation of the concatenation, using a mathematical framework that describes the distortion of the space through differential geometry.
}
{
Moving further away from negation and considering copulas in general, we can say that the essence of copulas is to combine (or separate) two elements from a specific point of view, beyond the overall similarities or differences.
The most straightforward example of this is the mapping relationship that dynamically associates different things in metaphors and analogies, as discussed in \S\ref{sec:analogy_metaphor}.
On the other hand, embedding in predicate logic-based knowledge graphs is static in nature, functioning by embedding similar things in similar positions and having the same relationship in the same direction. It does not fit into the cognitive processing of recognizing what were previously perceived as completely dissimilar and distant, and then in the next instant, recognizing them in close proximity.
This discussion of copulas is reminiscent of wormholes, which are discussed as a theoretical entity in astrophysics. In other words, it is a warp of space.
From this, although it is still only at the stage of fanciful conception, this suggests the possibility of attempting to model the dynamic deformation of a latent space according to the vector representation of the copula, using a mathematical framework that describes the distortion of the space by means of differential geometry.}

\subsection{Unification}\label{sec:unification}

\xxparatran{基本項$B_i, B_j$間の単一化は，前節で検討した意味ベクトルに基づいて計算される
類似度が一定以上である場合に，新たな陳述項$B_i \leftrightarrow B_j$
を明示的に認識することで実現できる．}
{Unification between elementary terms $B_i and B_j$ explicitly creates a new statement term $B_i \leftrightarrow B_j$ when the similarity calculated based on the semantic vector discussed in the previous section is above a certain level. This can be achieved by recognizing.}
{The Unification between the basic terms $B_i, B_j$ can be achieved by explicitly recognizing a new statement term $B_i \leftrightarrow B_j$ when the similarity calculated based on the semantic vectors examined in the previous section exceeds a certain level.
}
{
Unification between basic terms $B_i, B_j$ can be achieved by explicitly recognizing a new statement term $B_i \leftrightarrow B_j$ when the similarity calculated based on the semantic vectors examined in the previous section exceeds a certain level.}

\xxparatran{構成項の間の単一化については，
単一化しようとする項の間で要素数と順序が一致していることを前提条件とすれば，
比較的単純な手続き的アルゴリズムで，単一化可能かどうかの判定と，変数の束縛の処理を実現できると思われる．
例えば\cite{Arabshahi21}のsoft unificationもこの前提をおいている．}
{The unification between the constituent terms is a relatively simple procedural algorithm, provided that the number of elements and the order match between the terms to be unified. It seems that it is possible to judge whether it is possible and to process variable binding.
For example, the soft unification of \cite{Arabshahi21} also makes this premise.
}
{For unification among the composed terms, if the number and order of elements are consistent among the terms to be unified, a relatively simple procedural algorithm can be used to determine whether unification is possible or not and to bind the variables.
For example, the soft unification of \cite{Arabshahi21} is based on this assumption.
}
{
For unification among the composed terms, if the number and order of elements are consistent among the terms to be unified, a relatively simple procedural algorithm can be used to determine whether unification is possible or not and to bind the variables. For example, the soft unification of \cite{Arabshahi21} is based on this assumption.}

\xxparatran{しかしながら，より一般的かつ大規模なドメインを対象としようとすると，このような前提を置くことがすぐに足かせとなる．
例えば，`X gives Y Z'と`U gives V to W'という単純な2文から認識された複合項
$(\term{give}, X, Y, Z)$と$(\term{give\mh to}, U, V, W)$を単一化することを考えるだけも，
この前提は成り立たない．$Y$と対応すべきものは$V$ではなく$W$である．
また\S\ref{sec:ex_Google}では，推論の過程で，
$(\term{use}, x, y)$と
$(\term{want}, \term{people}$, $(\term{use}, \term{people}$, $\term{other\mh search\mh}$ $\term{engine}))$
の間の単一化を仮定した．
このような単一化を一般的に実現する方法としては，
seq2seqの形で単一化の際の対応関係を生成的に求めるアプローチが考えられる．}
{However, when we attempt to target a more general and large-scale domain, such assumptions quickly become a hindrance.
For example, the compound terms $(\term{give}, X, Y, Z)$ and $(\term{give\ mh to}, U, V, W)$, this assumption does not hold. $Y$ should correspond to $W$, not $V$. In \S\ref{sec:ex_Google}, $(\term{use}, x, y)$ and $(\term{want}, \term{people}$, we assumed unification between $(\term{use}, \term{people}$, $\term{other\mh search\mh}$ $\term{engine}))$. As a general method for realizing such unification, an approach to generatively find the correspondence in unification in the form of seq2seq is conceivable.}
{However, when trying to target more general and large domains, making such assumptions quickly becomes a stumbling block.
For example, this assumption does not hold even if we only consider unifying the compound terms $(\term{give}, X, Y, Z)$ and $(\term{give\mh to}, U, V, W)$ recognized from two simple sentences, `X gives Y Z' and `U gives V to W'.
What should correspond to $Y$ is not $V$ but $W$.
In addition, in\S\ref{sec:ex_Google}, we assumed unification between $(\term{use}, x, y)$ and $(\term{want}, \term{people}$, $(\term{use}, \term{people}$, $\term{other\mh search\mh}$ $\term{engine}))$ in the inference process.
As a general method for realizing such unification, we can consider an approach to generatively find the correspondence in unification in the form of seq2seq.
}
{
However, when trying to target more general and large domains, making such assumptions quickly becomes a stumbling block.
For example, this assumption does not hold even if we only consider unifying the compound terms $(\term{gives}, X, Y, Z)$ and $(\term{gives\mh to}, U, V, W)$ recognized from two simple sentences, `X gives Y Z' and `U gives V to W'.
What should correspond to $Y$ is not $V$ but $W$.
In addition, in\S\ref{sec:ex_Google}, we assumed unification between $(\term{use}, x, y)$ and $(\term{want}, \term{people}$, $(\term{use}, \term{people}$, $\term{other\mh search\mh engines}))$ in the inference process.
As a general method for realizing such unification, we can consider an approach to generatively find the correspondence in unification in the form of seq2seq modeling.
}

\subsection{Reasoner}\label{sec:reasoner}

\xxparatran{NATLが扱うべき推論には，少なくとも2種類がある．
すなわち所与の前提知識と問いに対し，問いの答えとして新たな認識（結論）を導く問題解決の推論（\S\ref{sec:math_problem}）と，
所与の前提知識と結論に対し，その間を繋ぐ経路を導く説明の推論（\S\ref{sec:argumentation}）である．
いずれにしても，NATLの推論器は，
前提知識としての項集合を$\mathbf{E}$とし，問いまたは結論を表す項$Q$を入力としたときに，
推論に使用した2項$T_1, T_2 \in \mathbf{E}$と新たな帰結の項の3つ組$(T_1, T_2, T_c)$を出力する関数$\mathcal{R}$として定式化できる．
すなわち，$(T_1, T_2, T_c, t) = \mathcal{R}(\mathbf{E},Q)$．
ここで$t$は推論結果についての確信度で，NALにおける真理値に相当する．
$[0,1]$間の実数値で主観確率として表現するのがもっとも自然な方法と思われるが，それに限る必要はない．}
{There are at least two kinds of reasoning that NATL should handle. In other words, for given premise knowledge and questions, problem-solving reasoning (\S\ref{sec:math_problem}) that leads to a new recognition (conclusion) as an answer to the question, and for given premise knowledge and conclusions, It is an explanation reasoning (\S\ref{sec:argumentation}) that guides the path connecting between them.
In any case, the NATL reasoner uses $\mathbf{E}$ as the term set as the premise knowledge, and the term $Q$ representing the question or the conclusion as the input. It can be formulated as a function $\mathcal{R}$ that outputs $T_1, T_2 \in \mathbf{E}$ and the triple $(T_1, T_2, T_c)$ of the new consequence term. That is, $(T_1, T_2, T_c, t) = \mathcal{R}(\mathbf{E},Q)$.
where $t$ is the degree of confidence about the inference result, which corresponds to the truth value in NAL. It seems to be the most natural way to express subjective probability with real values between $[0,1]$, but it is not necessary to limit it to that.}
{There are at least two types of inference that NATL should handle.
That is, problem-solving reasoning (\S\ref{sec:math_problem}), which leads to a new recognition (conclusion) as an answer to a given prerequisite knowledge and question, and explanatory reasoning (\S\ref{sec:argumentation}), which leads to a path connecting a given premise and conclusion.
In any case, the NATL reasoner can be formulated as a function $\mathcal{R}$ that, given a term set $\mathbf{E}$ as the assumed knowledge and a term $Q$ representing the question or conclusion as input, outputs a triplet $(T_1, T_2, T_c)$ consisting of the two terms $T_1, T_2 \in \mathbf{E}$ is used in the inference and a new consequent term.
Namely, $(T_1, T_2, T_c, t) = \mathcal{R}(\mathbf{E},Q)$.
Here, $t$ is the confidence of the inference result and corresponds to the truth value in NAL.
It seems to be the most natural way to express it as a subjective probability with a real-value between $ [0,1] $, but it is not necessary to limit it.
}
{
There are at least two kinds of reasoning that NATL should handle.
That is, problem-solving reasoning (\S\ref{sec:math_problem}), which leads to a new recognition (conclusion) as an answer to a given prerequisite knowledge and question, 
and explanatory reasoning (\S\ref{sec:argumentation}), which leads to a path connecting a given premise and conclusion.
In any case, the NATL reasoner can be formulated as a function $\mathcal{R}$ that, given a term set $\mathbf{E}$ as the assumed knowledge and a term $Q$ representing the question or conclusion as input, outputs a triplet $(T_1, T_2, T_c)$ consisting of the two terms $T_1, T_2 \in \mathbf{E}$ used in the inference and a new consequent term.
Namely, $(T_1, T_2, T_c, t) = \mathcal{R}(\mathbf{E},Q)$.
Here, $t$ is the confidence of the inference result and corresponds to the truth value in NAL.
It seems to be the most natural way to express it as a subjective probability with a real-value between 
$[0,1]$, but it is not necessary to limit it to that.}

\xxparatran{推論が完了したかどうかは，出力された$T_c$が問いの答えとなっている，あるいは前提知識と結論の間の推論経路が完成しているかどうかで判断される．ここで推論が完了していなければ，$T_c$は推論途中の中間結果であることや，その確信度といったメタ情報を伴う形で，$\mathbf{E}$の中に追加され，次のステップの推論（$\mathcal{R}(\mathbf{E},Q)$の計算）が繰り返される．}
{Whether or not the reasoning is completed is judged by whether the output $T_c$ is the answer to the question or whether the reasoning path between the premise knowledge and the conclusion is completed. If the inference is not completed here, $T_c$ is added to $\mathbf{E}$ with meta-information such as the fact that it is an intermediate result in the middle of inference and its confidence, and the next Step inference (computation of $\mathcal{R}(\mathbf{E},Q)$) is repeated.}
{Whether or not the inference is completed is judged by whether or not the output $ T_c $ is the answer to the question, or whether or not the inference path between the prerequisite knowledge and the conclusion is completed.
If the inference is not completed here, $T_c$ is added to $\mathbf{E}$ with meta-information such as the the intermediate result in the inference and its confidence level, and then the inference (Calculation of $\mathcal{R}(\mathbf{E},Q)$) of the next step is repeated.
}
{
Whether or not the inference is completed is judged by whether or not the output $ T_c $ is the answer to the question, or whether or not the inference path between the prerequisite knowledge and the conclusion is completed.
If the inference is not completed here, $T_c$ is added to $\mathbf{E}$ with meta-information such as the information that it is an intermediate result in the inference and its confidence level, and then the inference (Calculation of $\mathcal{R}(\mathbf{E},Q)$) of the next step is repeated.}

\xxparatran{推論器がその内部で行うべき主要な処理は，まず代入処理を適用する$T_1$と$T_2$を選択し，
次に必要に応じて代入処理の方向と出力項の繋辞を選択することである．
これらを推論状態$(\mathbf{E},Q)$に応じて適切に行う必要がある．
この処理を実現するには，\S\ref{sec:reinforcement_on_reasoner}で検討したように，
関数$\mathcal{R}$を教師あり学習または強化学習によって訓練するアプローチが有望と思われる．
訓練用データセットをどのように用意するかが，大きな課題となる．
以下，この点について見通しを議論する．}
{The main processing that the reasoner should do internally is to first select $T_1$ and $T_2$ to which the substitution process is applied, and then select the direction of the substitution process and the suffix of the output term if necessary. is.
It is necessary to perform these appropriately according to the inference state $(\mathbf{E},Q)$.
In order to realize this processing, as discussed in \S\ref{sec:reinforcement_on_reasoner}, a promising approach is to train the function $\mathcal{R}$ by supervised learning or reinforcement learning. How to prepare the training dataset is a big issue. Below, we discuss the prospects for this point.}
{The main process that the reasoner must perform internally is to first select $T_1$ and $T_2$ to which the assignment process is applied, and then select the direction of the assignment process and the copula of the output terms if necessary.
To realize this process, an approach in which the function $\mathcal{R}$ is trained by supervised learning or reinforcement learning, as discussed in \S\ref{sec:reinforcement_on_reasoner}, seems promising.
A major issue is how to prepare the training data set. The outlook for this point will be discussed below.
}
{
The main process that the reasoner performs internally is to first select $T_1$ and $T_2$ to which the substitution operation is applied, and then select the direction of the substitution and the copula of the output term if necessary.
To realize this process, an approach in which the function $\mathcal{R}$ is trained by supervised learning or reinforcement learning, as discussed in \S\ref{sec:reinforcement_on_reasoner}, seems promising.
A major issue is how to prepare the training data set. The outlook for this point will be discussed below.}

\xxparatran{NATLの運用には論理項に関する知識ベース（KB）の存在が前提となる．
（人間の子供の発達のように知識と思考能力の獲得を同時に進めることも考えられるが，
それは更に将来的な課題と位置づける．）
機能する推論器の実現にむけ
特定の推論タスクで意味のある検証と評価を行うためには，
そのKBを運用することで，与えられた問題が解けることがはっきりしている必要がある．
陳述項についてはFreeBase~\cite{Freebase}やYago~\cite{Yago}のような既存の知識グラフやオントロジなどから，
連関項については$\textsc{Atomic}^{20}_{20}$~\cite{Hwang2021COMETATOMIC2O}などの因果的知識も収集した
常識KBから数的には相当な量を集められると思うが，
特定の推論タスクと組み合わせたときに，必要な知識がどの程度カバーされているのかはっきりしない．}
{The operation of NATL assumes the existence of a knowledge base (KB) for logical terms. (Although it is conceivable to promote the acquisition of knowledge and thinking ability at the same time, as in the development of human children, this will be positioned as a future task.)
In order to conduct meaningful verification and evaluation on a specific inference task toward the realization of a functioning reasoner, it must be clear that the given problem can be solved by operating the KB. Predicate terms are derived from existing knowledge graphs and ontologies such as FreeBase~\cite{Freebase} and Yago~\cite{Yago}, and association terms are derived from $\textsc{Atomic}^{20}_{20}$ ~\cite{Hwang2021COMETATOMIC2O}, which also collects causal knowledge, I think you can gather a considerable amount numerically from the common sense KB, but how much of the necessary knowledge is covered when combined with a specific reasoning task? It's not clear if there are any.}
{The operation of NATL is  premised on the existence of a knowledge base (KB) of logical terms (It is also possible to simultaneously acquire knowledge and thinking skills, as in the development of a human child, but this is an issue for further study in the future).
In order to perform meaningful verification and evaluation of a particular inference task toward the realization of a functioning reasoner, it is necessary to clarify that the given problem can be solved by operating the KB.
I think we can collect a considerable amount of knowledge from existing knowledge graphs and ontologies such as FreeBase~\cite{Freebase} and Yago~\cite{Yago} for statement terms, and from common sense KB collecting causal knowledge such as $\textsc{Atomic}^{20}_{20}$~\cite{Hwang2021COMETATOMIC2O} for linkage terms, but how much the knowledge is required when combined with specific reasoning tasks is not clear.
}
{
The operation of NATL is premised on the existence of a knowledge base (KB) of logic terms (It is also possible to simultaneously acquire knowledge and thinking skills, as in the development of a human child, but this is an issue for further study in the future).
In order to perform meaningful verification and evaluation of a particular inference task toward the realization of a functioning reasoner, it is necessary to clarify that the given problem can be solved by operating the KB.
We can collect a considerable amount of knowledge from existing knowledge graphs and ontologies such as FreeBase~\cite{Freebase} and Yago~\cite{Yago} for statement terms, and from common sense KB collecting causal knowledge such as $\textsc{Atomic}^{20}_{20}$~\cite{Hwang2021COMETATOMIC2O} for linkage terms, but how much the knowledge is required when combined with specific reasoning tasks is not clear.}

\xxparatran{この問題に対して取りうるアプローチは3つ考えられる．
1つ目は，少量の論理項集合を用意すれば推論ができることが明らかなタスクを用いることである．
例えば\S\ref{sec:rules}で触れたbAbiデータセット\cite{bAbi}は極小さな世界モデルから自然言語形式の入力と正解ラベルのセットが
アルゴリズム的に生成されたもので，個別の推論に必要な知識を連間項の形で列挙することは可能と思われる．
ただし，bAbiはあくまでトイプロブレムであるので，明示的な規則知識を与えた上でNATLでbAbiタスクを解いても，
それだけでは有用性・汎用性の主張は難しく，あくまでNATLが期待したように動くことの実証にとどまる．
2つ目は，例えば$\textsc{Atomic}^{20}_{20}$を前提として，
そこにある知識の組み合わせで解くことができる問題を人手で大量に作ることである．
しかしこのアプローチは，存在する知識の全体像を把握した上で創造性を働かせる必要があり，
よほど巧妙な問題作成ワークフローを準備できないと，
クラウドソーシング等を用いた多数の作業者に依頼しての問題作成は困難と思われる．
多数の作業者に依頼できなければ，大量に作ることは難しい．}
{There are three possible approaches to this problem. The first is to use tasks that are clearly inferenceable with a small set of logical terms. For example, the bAbi dataset \cite{bAbi}, mentioned in \S\ref{sec:rules}, is an algorithmically generated set of natural-language input and correct labels from a tiny world model, and individual inference It seems possible to enumerate the knowledge necessary for However, since bAbi is just a toy problem, even if NATL solves the bAbi task with explicit rule knowledge, it is difficult to assert its usefulness and versatility, and NATL works as expected. It is only a demonstration of the fact. The second is to manually create a large number of problems that can be solved by combining existing knowledge, for example, given $\textsc{Atomic}^{20}_{20}$. However, with this approach, it is necessary to use creativity after grasping the overall picture of existing knowledge. It seems difficult to create. If you can't ask a lot of workers, it's difficult to make a lot.}
{
There are three possible approaches to this problem.
The first is to use a task in which it is clear that inference can be performed by preparing a small set of logical terms.
For example, the bAbi dataset \cite{bAbi} mentioned in \S\ref{sec:rules} is an algorithmically generated set of inputs and correct labels in natural language form from a tiny world model. model, and it seems possible to enumerate the knowledge required for individual inferences in the form of linkage terms.
However, since bAbi is only a toy problem, even if you solve the bAbi task with NATL after giving explicit rule knowledge, it is difficult to claim usefulness and versatility by itself; it is only a demonstration that NATL works as expected.
The second is to manually create a large number of problems that can be solved by combining the knowledge available there, assuming, for example, $\textsc{Atomic}^{20}_{20}$.
However, this approach requires creativity to be exercised after grasping the whole picture of existing knowledge, and unless a very sophisticated problem creation workflow is in place, it will be difficult to create problems by crowdsourcing which requires a large number of workers or other methods.
Unless a large number of workers can be commissioned, it will be difficult to create a large number of problems.
}
{
There are three possible approaches to this issue.
The first is to use a task in which it is clear that inference can be performed by preparing a small set of logic terms.
For example, the bAbi dataset \cite{bAbi} mentioned in \S\ref{sec:rules} is an algorithmically generated set of inputs and correct labels in natural language form from a tiny world model. It seems possible to enumerate the knowledge required for individual inferences in the form of linkage terms.
However, since bAbi tasks are only toy problems, even if the tasks are solved with NATL after giving explicit rule knowledge, it is difficult to claim the usefulness and versatility of NATL by itself; it is only a demonstration that NATL works as expected.
The second is to manually create a large number of problems that can be solved by combining the knowledge available there, assuming, for example, $\textsc{Atomic}^{20}_{20}$.
However, this approach requires creativity to be exercised after grasping the whole picture of existing knowledge, and unless a very sophisticated problem creation workflow is in place, it will be difficult to create problems by crowdsourcing or other methods.
Unless a large number of workers can be commissioned, it will be difficult to create a large number of problems.
}

\xxparatran{3つ目は，推論タスクのデータセットから必要なKBを構築することである．
これについては，\S\ref{sec:argumentation}で検討したARCTデータセット\cite{habernal-etal-2018-ARCT}などに対して，
DatumからClaimを導くのに必要なWarrantを明文化させることでKBを構築することが考えられる．
つまり「雨が降っているから傘を持っていくべきだ」という基本となる主張から，それを論証するための，
「雨が降ると濡れる」「濡れるのは嫌なこと」「人は嫌なことを避ける」「傘を持てば濡れることを避けられる」
といった背景に暗黙的に存在している様々な知識を，作業者に「それはなぜですか？」と問うことで抽出していく．
この作業も，テキストに対する単純なラベル付与などにくらべると，
作業自体も成果物の粒度や質の水準の統制も難しくなると予想されるが，
\S\ref{sec:natl}で提示したNATLの形式に従うように制約した作業環境・手順を整備することで，
2つ目の問題作成のアプローチより実施可能性を高められると考える．
またこの作業を通じて，NATLの記述力の検証も行える．
今後は，1つ目と3つ目のアプローチを並行して進めて，動作原理の実証と有用性の実証を行いたい．}
{The third is to construct the necessary KB from the dataset of the inference task. Regarding this, for the ARCT dataset \ cite {habernal-etal-2018-ARCT} examined in \ S \ ref {sec: argument}, clarify the Warrant required to derive Claim from Datum. It is conceivable to construct a KB with.
In other words, from the basic claim that "it's raining, you should bring an umbrella", to prove it, "get wet when it rains", "don't want to get wet", "people don't like it" Various knowledge that implicitly exists in the background such as "avoid things" and "you can avoid getting wet with an umbrella" is extracted by asking the worker "why?".
Compared to simple labeling of texts, it is expected that this work itself will be more difficult to control the particle size and quality level of deliverables, but the NATL presented in \ S \ ref {sec: natt} It is considered that the feasibility can be improved from the second problem creation approach by preparing a work environment / procedure restricted to follow the format.
In addition, through this work, the descriptive power of NATL can be verified.
In the future, we would like to proceed with the first and third approaches in parallel to demonstrate the operating principle and usefulness.}
{The third is to construct the necessary KB from the dataset of the inference task. For this, KB could be constructed by making explicit the Warrant necessary to derive the Claim from the Datum, such as for the ARCT dataset \cite{habernal-etal-2018-ARCT} discussed in \S\ref{sec:argumentation}.
In other words, from the basic assertion that ``it is raining and you should bring an umbrella,'' to the arguments for it, which exist implicitly in the background, such as ``when it rains, you get wet'', ``getting wet is unpleasant'', ``people avoid unpleasant things'', and ``if you carry an umbrella, you can avoid getting wet.''. The various kinds of knowledge are extracted by asking ``Why?'' by the workers.
This work is also expected to be more difficult than simple labeling of text, both in terms of the work itself and in controlling the level of granularity and quality of the output, but we believe that by developing a work environment and procedures that are constrained to follow the NATL format presented in \S\ref{sec:natl}, we can improve the feasibility from the second problem creation approach. 
In addition, through this work, the descriptive power of NATL can be verified.
In the future, we would like to proceed with the first and third approaches in parallel to demonstrate the operating principle and usefulness.
}
{
The third is to construct the necessary KB from the dataset of realistic inference tasks. A KB could be constructed by making explicit the Warrant necessary to derive the Claim from the Datum, such as for the ARCT dataset \cite{habernal-etal-2018-ARCT} discussed in \S\ref{sec:argumentation}.
In other words, from the basic assertion such as ``it is raining and you should bring an umbrella,'' we will be able to extract various kinds of knowledge which exists implicitly in the background of the arguments, such as ``when it rains, you get wet'', ``getting wet is unpleasant'', ``people avoid unpleasant things'', and ``if you carry an umbrella, you can avoid getting wet.'' by asking ``Why?'' to the workers.
This annotation work is also expected to be more difficult than simple labeling of text, both in terms of the work itself and in controlling the level of granularity and quality of the output, but we believe that by developing a work environment and procedures that are constrained to follow the NATL format presented in \S\ref{sec:natl}, we can have better feasibility than the second approach of creating problems. 
In addition, through this annotation work, the descriptive power of NATL can be verified. In the future, we would like to proceed with the first and third approaches in parallel to demonstrate the theoretical validity and practical usefulness of NATL.}

\xxparatran{上記の実証を行う上で解決が必要な他の主要な要検討事項として，
(1) \S\ref{sec:unification}で述べた基本項の単一化としての陳述項$B_i \leftrightarrow B_j$の生成や，
\S\ref{sec:ex_Umbrella}で導入した2つの推論結果の連言化を，
推論器$\mathcal{R}$が推論の1ステップとして行うのか，それとも推論器と並置された別のモジュールによって並列的に行うのかを定めることと，
(2) 確信度$t$の計算をどのように行うのかを定めること，の2点がある．}
{Other major considerations that need to be resolved in carrying out the above demonstration are:
(1) Generation of the statement term $B_i \leftrightarrow B_j$ as a unification of the elementary term described in \S\ref{sec:unification} and the two inferences introduced in \S\ref{sec:ex_Umbrella} (2) whether the concatenation of results is performed by the reasoner $\mathcal{R}$ as one step of inference or in parallel by a separate module co-located with the reasoner; There are two points: defining how the computation of degrees $t$ is done.}
{Other major considerations that need to be resolved in order to demonstrate the above proofs are (1) Generation of the statement term $B_i \leftrightarrow B_j$ as a unification of the basic terms described in \S\ref{sec:unification} and the conjunction of the two inference results introduced in \S\ref{sec:ex_Umbrella} are done by the reasoner $\mathcal{R}$ as one step of inference or in parallel by another module juxtaposed to the reasoner, and (2) to specify how to determine the confidence $t$.
}
{
Other major considerations that need to be resolved in order to demonstrate the above proofs are (1) to determine the generation procedure of the statement term $B_i \leftrightarrow B_j$ as a unification of the basic terms described in \S\ref{sec:unification} and the design choice that the conjunction of the two inference results introduced in \S\ref{sec:ex_Umbrella} are done by the reasoner $\mathcal{R}$ as one step of inference or in parallel by another module juxtaposed to the reasoner, and (2) to specify how to estimate confidence $t$.}

\subsection{Translation from natural language to term representation language}\label{sec:nl2natl}

\xxparatran{最後に，自然言語で記述された知識を項表示言語へ変換する方法について見通しを述べる．
対象とするドメインにおける言語データの複雑さに応じて様々なアプローチが考えられるが，
まず必要なことは，基本項に対応するまとまり（単純には文中の区間としての単語列）を抽出することである．
それらを抽出できれば，\ref{sec:vector_semantics}で述べたように，事前学習済み言語モデルを用いて，ベクトル意味表現を得ることができる．}
{Finally, we present a perspective on how to convert knowledge written in natural language into a term representation language. 
Various approaches can be considered depending on the complexity of the linguistic data in the target domain.
First of all, it is necessary to extract a group corresponding to a basic term (simply a word string as an interval in a sentence).
If we can extract them, we can obtain a vector semantic representation using a pretrained language model, as described in \S\ref{sec:vector_semantics}.}
{Finally, we give a perspective on how to convert knowledge written in natural language into a term display language.
Various approaches can be considered depending on the complexity of the linguistic data in the target domain, but the first thing is to extract the group corresponding to the basic term (simply the word string as an interval in the sentence). be.
If they can be extracted, vector semantic expressions can be obtained using the pre-learned language model as described in \S\ref{sec:vector_semantics}.
}
{
Finally, we give a perspective on how to convert knowledge described in natural language into TRL.
Various approaches can be considered depending on the complexity of the linguistic data in the target domain.
First of all, it is necessary to extract the chunks corresponding to basic terms (in the simplest cases, a word sequence as a span in a sentence). If they can be extracted, vector semantic expressions can be obtained using a pretrained language model as described in \S\ref{sec:vector_semantics}.}

\xxparatran{このような抽出には，オープン情報抽出技術（Open IE）\cite{Angeli-CoreNLP-OpenIE,stanovsky-etal-2018-supervised}を利用できる．
\S\ref{sec:reasoner}でも触れたbAbiデータセットに現れるような単純な言語表現（例えば，``Mary went to the bathroom.''）であれば，
例えばAllenNLP\cite{Gardner2017AllenNLP}の一部として公開されている
学習済みモデル\footnote{https://demo.allennlp.org/open-information-extraction}
などとアドホックな修正規則を組み合わせることで一定の用をなすと思われる．
sheなどの代名詞の照応先（指示先）の同定には，
同様に共参照解析\cite{Lee2018HigherorderCR}\footnote{
https://demo.allennlp.org/coreference-resolution}
が使える．
共参照解析によって代名詞をその指示先（先行詞）に置換してから，
オープン情報抽出を行えばよい．}
{For such extraction, Open Information Extraction Technology (Open IE) \cite{Angeli-CoreNLP-OpenIE,stanovsky-etal-2018-supervised} can be used.
For simple linguistic expressions that appear in the bAbi dataset mentioned in \S\ref{sec:reasoner} (eg ``Mary went to the bathroom.''),
For example, by combining ad-hoc correction rules with trained models published as part of AllenNLP\cite{Gardner2017AllenNLP}\footnote{https://demo.allennlp.org/open-information-extractio} Similarly, coreference analysis \cite{Lee2018HigherorderCR}\footnote{https://demo.allennlp.org/coreference-resolution}
can be used. After replacing pronouns with their referents (antecedents) by coreference analysis, open information extraction can be performed.
}
{For such extraction, the Open Information Extraction technique (Open IE) \cite{Angeli-CoreNLP-OpenIE,stanovsky-etal-2018-supervised} can be used.
A simple linguistic expression(for example, ``Mary went to the bathroom.''), such as the one that appears in the bAbi dataset mentioned in \S\ref{sec:reasoner}, may be of some use by combining ad-hoc modification rules with, for example, a pre-trained model\footnote{https://demo.allennlp.org/open-information-extraction} published as part of AllenNLP\cite {Gardner2017AllenNLP}.
Similarly, coreference resolution\cite{Lee2018HigherorderCR}\footnote{
https://demo.allennlp.org/coreference-resolution} can be used to identify the referent (indication) of pronouns such as ``she''.
After replacing the pronoun with its indication (antecedent) by coreference resolution, open information extraction can be performed.
}
{
For such extraction, Open Information Extraction (Open IE) \cite{Angeli-CoreNLP-OpenIE,stanovsky-etal-2018-supervised} can be used.
A simple linguistic expression (for example, ``Mary went to the bathroom.''), such as the one that appears in the bAbi dataset mentioned in \S\ref{sec:reasoner}, may be of some use by combining ad-hoc modification rules with, for example, a pre-trained model
published as part of AllenNLP \cite{Gardner2017AllenNLP}.\footnote{https://demo.allennlp.org/open-information-extraction}
Similarly, coreference resolution \cite{Lee2018HigherorderCR}\footnote{
https://demo.allennlp.org/coreference-resolution} can be used to identify the referent (antecedent) of pronouns such as ``she''.
After replacing the pronoun with its referent found by coreference resolution, Open IE can be performed.}

\xxparatran{しかしながら，十分な修正規則を用意することは単純なドメインであってもそれほど簡単ではない．
例えば，脚注14で挙げた例文``She goes to the zoo everyday.''を，
前述のオープン情報抽出の学習済みモデルに与えると，
$\mbox{[ARG0: She] [V: goes] [ARG4: to the zoo everyday].}$
という出力を得る．
これを$(\term{goes\mhyp to}, \term{she}, \term{the\mhyp zoo}, \term{everyday})$
のように翻訳したければ，
前置詞toを$\mbox{[ARG4:]}$から$\mbox{[V:]}$の区間に含め直す処理と，
副詞everydayを切り分ける修正処理が必要になる．
規則を用いた修正で対処しきれなくなれば，
独自に学習データを用意してオープン情報抽出モデルを構築する必要がある．}
{However, providing enough modification rules is not so easy even for simple domains. For example, if the example sentence ``She goes to the zoo everyday.'' given in footnote 14 is given to the trained model for extracting open information, $\mbox{[ARG0: She] [V: goes] [ARG4 : to the zoo everyday].}$ output. If you want to translate this as $(\term{goes\mhyp to}, \term{she}, \term{the\mhyp zoo}, \term{everyday})$, replace the preposition to with $\mbox{ It is necessary to re-include it in the interval from [ARG4:]}$ to $\mbox{[V:]}$ and to modify it to separate the adverb ``everyday"".
If it becomes impossible to deal with the correction using rules, it is necessary to prepare training data independently and construct an open information extraction model.
}
{However, it is not so easy to prepare sufficient modification rules even for a simple domain.
For example, if the example sentence ``She goes to the zoo everyday.'' in footnote 14 is given to the learned model of open information extraction described above, we obtain the output $\mbox{[ARG0: She] [V: goes] [ARG4: to the zoo everyday]}$.
If we want to translate this into $(\term{goes\mhyp to}, \term{she}, \term{the\mhyp zoo}, \term{everyday})$, we need to reinclude the preposition ``to'' in the interval $\mbox{[ARG4:]}$ to $\mbox{[V:]}$ and modify the adverb ``everyday'' to separate it.
If the correction using the rules cannot be dealt with, it is necessary to prepare the learning data independently and build an open information extraction model.
}
{However, it is not so easy to prepare sufficient modification rules even for a simple domain.
For example, if the example sentence ``She goes to the zoo everyday.'' in footnote 14 is given to the learned model of Open IE described above, we obtain the output $\mbox{[ARG0: She] [V: goes] [ARG4: to the zoo everyday]}$.
If we want to translate this into $(\term{goes\mhyp to}, \term{she}, \term{the\mhyp zoo}, \term{everyday})$, we need to reinclude the preposition ``to'' in the interval $\mbox{[ARG4:]}$ to $\mbox{[V:]}$ and modify the adverb ``everyday'' to separate it.
If the correction using the rules is not tractable enough, it is necessary to prepare the training data from scratch to build an Open IE model.}

\xxparatran{また，基本項に対応するまとまりの認識だけでなく，文に対応するべき構成項の種別（複合項・陳述項・連関項）を正しく認識する必要がある．
これも，平易な表現しか現れない小規模なドメインを扱ううちは
キーワード・パターンマッチングに基づく規則処理で対応できると思われるが，
複雑・大規模になれば，教師あり学習による分類器が必要になるだろう．}
{In addition, it is necessary to correctly recognize the types of constituent terms (compound term, declarative term, and associated term) that should correspond to a sentence, in addition to recognizing groups corresponding to elementary terms. As long as we deal with small-scale domains in which only simple expressions appear, rule processing based on keyword/pattern matching seems to be sufficient. It will be.}
{In addition to recognizing the clusters corresponding to the basic term, it is necessary to correctly recognize the type of composed term (compound terms, statement terms, and conjunction terms) that should correspond to the sentence.
It seems that this can also be handled by rule processing based on keyword pattern matching while dealing with small domains where only plain expressions appear, but if it becomes complicated and large, a classifier by supervised learning is required.
}
{
In addition to recognizing the chunks corresponding to basic terms, it is necessary to correctly recognize the type of composed term (compound term, 
statement term, and conjunction term) that corresponds to the sentence.
This could also be handled by rule processing based on keyword pattern matching while dealing with small domains where only plain expressions appear.
If it becomes complicated and large scale, a classifier by supervised learning will be necessary.
}

\xxparatran{文中の基本項の認識，文に対応する構成項の種別の認識，
いずれに対しても学習データを用意して教師あり学習を行うのであれば，
両方のタスクを翻訳タスクとしてまとめてしまい，
GPT~\cite{GPT2,GPT3}やT5~\cite{ByT5}のような生成型大規模事前学習言語モデルをfine-tuningして，seq2seq形式で扱うのが良いと思われる．}
{Recognition of elementary terms in sentences, recognition of types of constituent terms corresponding to sentences,
If we prepare training data for both and perform supervised learning, we combine both tasks into a translation task, such as GPT~\cite{GPT2,GPT3} and T5~\cite{ByT5}. It seems to be better to fine-tune a generative large-scale pretrained language model and handle it in the seq2seq format.
}
{
For the following two tasks, recognition of the basic terms in a sentence and recognition of the types of composed terms corresponding to a sentence, if training data prepared for both tasks and supervised learning is performed, it would be better to combine both tasks as a translation task, fine-tune a generative large-scale pre-training language model such as GPT~\cite{GPT2,GPT3} or T5~\cite{ByT5}, and handle it in a seq2seq format.
}
{
For the following two tasks, recognition of basic terms in a sentence and recognition of the types of composed terms corresponding to a sentence, if training data prepared for both tasks and supervised learning is performed, it would be better to combine both tasks as a translation task, that is, fine-tuning a generative large-scale pre-trained language model such as GPT~\cite{GPT2,GPT3} or T5~\cite{ByT5} to handle the task in a seq2seq format.
}

\section{Conclusion}\label{sec:conclusion}

\xxparatran{本論文は，人間が行うような日常的推論，創造的な記号処理，記号を介した文化学習を行うAIシステムの
実現を目指す研究の第一歩として，項論理に着想を得た知識表現言語（項表示言語）と，
その言語を用いて推論を行うための非公理的項論理の理論を提出した．
その理論は，3つのクラスの知識表現上の5つの型の記号操作で，人間の推論・議論を記述し尽くせると主張するものである．
それは，アナロジーや比喩を含む，人間の多様な思考を，
(1) 時間・空間的な隔たりを持ちながらも一定の関係性をもつ複数の事物を纏めて1つの対象として再認識する，
(2) そのように纏められた2つの対象の間にさらに関係性を認め結びつける，
という2つの基本的な認知機構からなる形式的記号処理の基盤の上で説明することを目論む．}
{This paper is a knowledge expression language inspired by term logic as the first step of research aiming at the realization of an AI system that performs daily reasoning, creative symbol processing, and cultural learning through symbols as humans do. We submitted (term display language) and the theory of non-axiomatic term logic for making inferences using that language.
The theory argues that five types of symbolic manipulation on knowledge representation in three classes can describe human reasoning and argument.
It is a diverse human thinking, including analogies and metaphors.
(1) Re-recognize multiple things that have a certain relationship while having a time and space gap as one object.
(2) Recognize and connect further relationships between the two objects so grouped together.
I aim to explain on the basis of formal symbol processing consisting of two basic cognitive mechanisms.}
{This paper is a knowledge expression language (term representation language) inspired by term logic and the theory of non-axiomatic term logic as the first step of research aiming at the realization of an AI system that performs daily reasoning, creative symbolic processing, and cultural learning through symbols as humans do.
The theory claims that five types of symbolic manipulations on three classes of knowledge representation can fully describe human reasoning and argumentation.
It aims to explain the diversity of human thought, including analogy and metaphor, on the basis of formal symbolic processing, which consists of two basic cognitive mechanisms :
(1) To re-recognize multiple things as a single object, which have a certain relationship with each other while being separated temporally and spatially.
(2) Recognize and connect further relationships between the two objects so grouped together.
}
{
This paper has proposed Term Representation Language (TRL), a knowledge representation language inspired by term logic and Non-Axiomatic Term Logic (NATL), a computational theory of cognitive symbolic reasoning, as the first step of research aiming at the realization of an AI system that performs daily reasoning, creative symbol processing, and cultural learning through symbols as humans do.
The theory claims that five types of symbolic manipulation on knowledge representations in three classes can describe human reasoning and argumentation.
It aims to explain the divers human thinking, including analogy and metaphor, on the basis of formal symbolic processing, which consists of two basic cognitive mechanisms :
(1) To re-recognize multiple things as a single object, which have a certain relationship with each other while being separated temporally and spatially.
(2) Recognize and connect further relationships between the two objects so grouped together.}

\xxparatran{その理論的妥当性および有用性については，議論に対する例証的な分析を通じて，一定の定性的評価を与えられたと考えている．
さらなる事例分析等を通じた妥当性の検証と理論の修正，
定量的評価および計算機実装による実現可能性・応用的有用性の実証は，
今後の課題である．}
{We believe that the theoretical validity and usefulness were given a certain qualitative evaluation through an exemplary analysis of the argument.
Verification of validity through further case analysis, revision of theory, quantitative evaluation, and demonstration of feasibility and applied usefulness by computer implementation are future tasks.}
{We believe that the theoretical validity and usefulness were given a certain qualitative evaluation through an exemplary analysis of the argument.
Verification of theory through further case analysis, revision of theory, quantitative evaluation, and demonstration of feasibility and applicability through computer implementation are future tasks.
}
{
We believe that the theoretical validity and practical usefulness of NATL has been proven to a certain extent through the qualitative analysis of exemplary arguments.
Verification and revision of the theory through further case analysis,  quantitative evaluation, and demonstration of feasibility and applicability through computer implementation are future tasks.}

\xxparatran{本論文が提出した「非公理的項論理」の名は，\S\ref{sec:natl}の冒頭で触れたように，\citet{Wang94,Wang13}の非公理的論理（Non-Axiomatic Logic）に負っている．
しかしながら，本論がもう1つ大きく負うところのToulmin~\cite{Toulmin58}も，
同様のことを述べている（\textit{``Unfortunately an idealized logic, such as the mathematical model leads us to, cannot keep in serious contact with its practical application. \textbf{Rational demonstration is not a suitable subject for a timeless, axiomatic science};''}, p. 136）．
WangはToulminの著作を引いていない．
また両者の切り口はそれぞれ推論と議論で必ずしも同一ではない．
しかし両者の間の関連性は明らかであろう．
非公理的であることの重要性は，Wangだけが主張してきたものではない．おそらく，筆者が寡聞にして知らないだけで，同様の主張は多くの先人によって繰り返されてきたものと推察する\footnote{
例えばToulminはより最近の著書\cite{Toulmin01}にて，
同様の思想から近代西洋の合理主義を批判している（同書表題の\textit{Return to Reason}は，rationalityからreasonablenessへの回帰を意味する）．
同様の批判\cite{Nakano12}が日本思想史の観点からも展開されているが，中野はToulminを引いていない．
}．}{
As mentioned at the beginning of \ S \ ref {sec: natt}, the name of "non-axiomatic logic" submitted in this paper is the non-axiomatic logic of \ citet {Wang94, Wang13}. ).
However, Toulmin~\cite{Toulmin58}, which this paper owes to another major point, also states the same thing. Wang does not draw Toulmin's work.
In addition, both perspectives are not always the same in reasoning and discussion.
However, the relationship between the two will be clear.
The importance of being axiom is not the only claim of Wang. Perhaps the same claim has been repeated by many ancestors, just by the fact that the author does not know it\footnote{
For example, Toulmin wrote in his more recent book \cite {Toulmin01}.
He criticizes modern Western rationalism from a similar idea (\ textit {Return to Reason} in the title of the book means a return from rationality to reasonableness).
A similar criticism \ cite {Nakano12} has been developed from the perspective of the history of Japanese thought, but Nakano does not draw Toulmin.}.}
{The name ``Non-Axiomatic Term Logic'' submitted by this paper owes its name to the Non-Axiomatic Logic of \citet{Wang94,Wang13}, as mentioned at the beginning of \S\ref{sec:natl}.
However, Wang, states the same thing about Toulmin~\cite{Toulmin58} (\textit{``Unfortunately an idealized logic, such as the mathematical model leads us to, cannot keep in serious contact with its practical application. \textbf{A Rational demonstration is not a suitable subject for a timeless, axiomatic science}''}, p. 136), to whom this paper owes another major debt, does not cite Toulmin's writings.
The two approaches are not necessarily the same in terms of reasoning and argumentation, respectively.
However, the relevance between them is obvious.
Wang is not the only one who has insisted on the importance of being non-axiomatic. I assume that similar arguments have been repeated by many predecessors, perhaps only unknown to the author\footnote{
For example, Toulmin in his more recent book \cite{Toulmin01} criticizes modern Western rationalism from a similar idea (the title of the book,\textit{Return to Reason}, refers to a return from rationality to reasonableness).
A similar criticism \cite {Nakano12} has been developed from the perspective of the history of Japanese thought, but Nakano does not cite Toulmin.}.
}
{
The name of Non-Axiomatic Term Logic owes its name to Non-Axiomatic Logic of \citet{Wang94,Wang13}.
However, Toulmin also states the same thing (\textit{``Unfortunately an idealized logic, such as the mathematical model leads us to, cannot keep in serious contact with its practical application. \textbf{A Rational demonstration is not a suitable subject for a timeless, axiomatic science}''} \cite{Toulmin58}, p. 136), to whom this paper owes another major debt.
Wang does not cite Toulmin's work.
The approaches of the two are not necessarily the same, one is from reasoning and the other from argumentation.
However, the relevance between them is obvious.
The importance of being axiom is not the only claim of Wang. 
Perhaps the same claim has been repeated by many ancestors.\footnote{
For example, Toulmin, in his more recent book \cite{Toulmin01},
criticizes modern Western rationalism from a similar idea (\textit{Return to Reason} in the title of the book means a return from rationality to reasonableness).
A similar criticism has been developed from the perspective of the history of Japanese thought by Nakano \cite{Nakano12}, but he does not cite Toulmin.}
}

\xxparatran{述語論理を用いて照応現象を説明しようとした談話表示理論\cite{DRT}の主要な貢献の1つは，
\citet{Krahmer98}によれば，視覚的にアピールする形式的表示手段を提供した（\textit{``they are visually appealing.''}, p.35）ことであった．
\citet{Fauconnier02}は，適切な形式的表示手段の発明が数学の発展において重要であったことを指摘し，
さらに踏み込んで「（意味と不可分の）形式が想像力を支える」（\textit{``Like the warrior and the armor, meaning systems and formal systems are inseparable. \ldots\ forms prompt largely unconscious and unnoticed constructions of the imagination.''}, p.11)と訴えている\footnote{
Lakoffも数学の多くの重要な概念は比喩的なブレンディングであると主張している\cite{LakoffWMCF}．
}．
このような観点からいえば，非公理的項論理は，
人間の日常的推論・思考を計算機上に再現するための，意味と不可分な形式的表示手段の1案を提示したものと言える．}{
One of the major contributions of the discourse display theory \ cite {DRT}, which tried to explain the anaphoric phenomenon using predicate logic, provided a visually appealing formal display means, according to \citet{Krahmer98}. Was that.
\citet{Fauconnier02} points out that the invention of proper formal representation was important in the development of mathematics, and goes further and argues that "formality (meaning and inseparable) supports imagination."
From this point of view, it can be said that non-axiomatic term logic presents one idea of a formal display means that is inseparable from meaning in order to reproduce human daily reasoning and thinking on a computer.}
{One of the major contributions of discourse display theory \cite{DRT}, which attempted to explain the phenomenon of collocations using predicate logic, was, according to \citet{Krahmer98}, to provide a means of formal display that was visually appealing(\textit{``they are visually appealing.''}, p.35).
\Citet{Fauconnier02} points out that the invention of appropriate formal means of representation was important in the development of mathematics, and goes further to appeal that ``formality (inseparable from meaning) supports imagination'' (\textit{``Like the warrior and the armor, meaning systems and formal systems are inseparable. \ldots\ forms prompt largely unconscious and unnoticed constructions of the imagination.''}, p.11).\footnote{
Lakoff also argues that many important concepts in mathematics are figurative blending\cite{LakoffWMCF}． 
}
From this point of view, it can be said that non-axiomatic term logic presents one idea of a formal display means that is inseparable from meaning in order to reproduce human daily reasoning and thinking through a computer.
}
{
One of the major contributions of Discourse Representation Theory \cite{DRT}, which attempted to explain the phenomenon of anaphora using predicate logic, was, according to \citet{Krahmer98}, to provide a means of formal representation that was visually appealing (p.35).
\citet{Fauconnier02} point out that the invention of appropriate formal means of representation was important in the development of mathematics, and goes further to appeal that formality inseparable from meaning supports imagination. (\textit{``Like the warrior and the armor, meaning systems and formal systems are inseparable. \ldots\ forms prompt largely unconscious and unnoticed constructions of the imagination.''}, p.11).\footnote{
Lakoff also argues that many important concepts in mathematics are figurative blending \cite{LakoffWMCF}. 
}
From this point of view, it can be said that NATL presents one idea of a formal representation means that is inseparable from meaning in order to reproduce human daily reasoning and thinking through a computer.
}

\xxparatran{数理論理は，形式主義を推し進めて記号から意味（実在）を切り離すことで，形式だけによる厳密な議論を可能にした．
本研究はその流れに逆行するものであり，提出した非公理的項論理は形式だけではもはや何も語ることができない
（\S\ref{sec:argumentation}で示した分析は一定の形式に従っているが，その結果の妥当性の判断は記号の読み手の主観に依存している）が，
客観的で再現可能な意味の情報処理技術（埋め込み）と統合することで，
形式主義が攻撃した自然言語による記述の意味的曖昧性・属人性の問題を超越しつつ，
計算機を人と同じ意味の世界に共存させることを目論むものである．}{
Mathematical logic promoted formalism and separated meaning (existence) from signs, enabling rigorous discussions based on form alone.
This research goes against that trend, and the submitted non-irrational term logic can no longer be said by form alone, but it should be integrated with information processing technology (embedding) in an objective and reproducible sense. The aim is to make computers coexist in a world with the same meaning as humans, while transcending the problems of semantic ambiguity and personality of descriptions in natural language attacked by formalism.
}
{
Mathematical logic promoted formalism and separated meaning (existence) from signs, enabling rigorous discussions based on form alone.
This research goes against this trend, and although the submitted non-axiomatic term logic can no longer say anything only in form (The analysis presented in \S\ref{sec:argumentation} follows a certain form, but the judgment of the validity of the results depends on the subjectivity of the reader), by integrating it with information processing technology (embedding) of objective and reproducible meaning, it transcends the problem of semantic ambiguity and personality of descriptions in natural language that formalism attacked, and aims to make computers coexist in a world with the same meaning as humans.
}
{
Mathematical logic promoted formalism and separated meanings from symbols, enabling rigorous discussions based on form alone.
This research goes against this trend, and the submitted theory of NATL can no longer say anything only in form (The analysis presented in \S\ref{sec:argumentation} follows a certain form, but the judgment of the validity of the results depends on the subjectivity of the reader so far).
However, by integrating the form with objective and reproducible information processing technology of meaning (embedding), NATL tries to transcend the problem of semantic ambiguity and personality of descriptions in natural language that formalism attacked, and aims to make computers coexist in the same meaningful world as humans.
}

\xxparatran{本論文で提示した非公理的項論理による推論は，
連関項として表現される様々な規則・因果的知識に基づいて遂行される．
しかし，推論に必要な連関項が予め全て明示的に保持されていると考えることは無理があるように思われる．
例えば，脚注\ref{fn:choco}で示した因果規則的信念は，
「昨日」や「戸棚」の部分を変えてバリエーションを無限に作ることができる．
もちろんこれらの部分が抽象化・一般化された規則の存在を仮定することもできるが，
推論の必要に応じて連関項が過去の大量の経験の蓄積から再生的に生成される，
つまり説明のために作り出される，と考えることもできるだろう．
これは\cite{Hofstadter2013}の考えに近いように思われるし，
GPT~\cite{GPT2,GPT3}のような大規模生成言語モデルはこれを技術的に可能にするように思われる．
一方で，言語モデルによる生成は白昼夢的で，錨を失った船のように生成の海を漂う．
局所的な条件付けで言語モデルが生み出す部品をNATLに基づく推論器が大域的に組み立てることで，
人間のように語ることができるAIを生み出せるのではないかと期待する．
これはまさに\S\ref{sec:separation_of_logic_and_language}で議論した，推論能力（Mercierらによれば理性）
と言語運用能力の分離と連携である．}{
Inference by non-axiomatic term logic presented in this paper is performed based on various rules and causal knowledge expressed as associated terms. However, it seems unreasonable to think that all the associated terms necessary for inference are explicitly stored in advance. For example, the causal regular belief shown in the footnote \ref{fn:choco} can be infinitely varied by changing the parts of ``yesterday'' and ``cupboard''. Of course, it is possible to assume the existence of rules in which these parts are abstracted and generalized. You can think of it as being made for This seems close to the ideas of \cite{Hofstadter2013}, and large-scale generative language models such as GPT~\cite{GPT2,GPT3} seem to make this technically possible. On the other hand, language model generation is like a daydream, drifting in the sea of generation like a ship that has lost its anchor. We expect that a NATL-based reasoner can globally assemble the parts generated by the language model with local conditioning to create an AI that can speak like a human being. This is exactly the separation and linkage of reasoning ability (reason according to Mercier et al.) and linguistic ability discussed in \S\ref{sec:separation_of_logic_and_language}.
}
{Inference by non-axiomatic term logic presented in this paper is carried out based on various rules and causal knowledge expressed as linkage terms.
However, it seems unreasonable to assume that all the necessary linkage terms for inference are explicitly retained in advance.
For example, the causal rule beliefs shown in footnote \ref{fn:choco} can be created infinitely different by changing the ``yesterday'' and ``cupboard'' parts.
Of course, it is possible to assume the existence of rules in which these parts are abstracted and generalized, but as necessary for inference, linkage terms are regenerated from the accumulation of a large amount of past experience, that is, in the explanation.
This seems close to the idea of \cite{Hofstadter2013}, and large-scale generative language models such as GPT~\cite{GPT2,GPT3} seem to make this technically possible.
On the other hand, the generation by the language model is daydreaming, drifting through the ocean of generation like a ship that has lost its anchor.
We hope that the NATL-based inferencer will globally assemble the parts produced by the language model under local conditioning to create AI that can speak like a human.
This is exactly the separation and linkage of reasoning (According to Mercier et al.) and linguistic performance discussed in \S\ref{sec:separation_of_logic_and_language}.
}
{Reasoning by NATL presented in this paper is carried out based on various rules and causal knowledge expressed as linkage terms.
However, it seems unreasonable to assume that all the necessary linkage terms for reasoning are explicitly retained in advance.
For example, the causal rule beliefs shown in footnote \ref{fn:choco} can be created infinitely different by changing the ``yesterday'' and ``cupboard'' parts.
Of course, it is possible to assume the existence of rules in which these parts are abstracted and generalized. However, it is also reasonable to suppose that, as necessary for reasoning, linkage terms are regenerated from the accumulation of a large amount of past experience.
This seems close to the idea of \cite{Hofstadter2013}, and large-scale generative language models such as GPT~\cite{GPT2,GPT3} seem to make this technically possible.
On the other hand, the generation by those language models is daydreaming, drifting through the ocean of generation like a ship that has lost its anchor.
We hope that the NATL-based reasoner will globally assemble the parts produced by language models under local conditioning to create AI that can speak like a reasonable human.
This is exactly the separation and coordination of reasoning and linguistic competence discussed in \S\ref{sec:separation_of_logic_and_language}.
}


\end{document}